# *A Multilingual Sentiment Lexicon for Low-Resource Language Translation using Large Languages Models and Explainable AI*

Melusi Malinga, Isaac Lupanda, Mike Wa Nkongolo, and Phil van Deventer
Department of Informatics, University of Pretoria
Low-Resource Language Processing Lab (LRLPL)
Mike.wankongolo@up.ac.za, phil.vandeventer@up.ac.za

## Abstract

South Africa's and the Democratic Republic of Congo's (DRC) diverse linguistic landscape, with languages such as Zulu, Sepedi, Afrikaans, French, and Ciluba, poses significant challenges for AI-driven translation and sentiment analysis due to a lack of accurate, labeled data. This study addresses these challenges by proposing a lexicon originally designed for French and Tshiluba (Ciluba) to include translations in English, Afrikaans, Sepedi, and Zulu. Language-specific sentiment scores are integrated, enhancing cultural relevance in sentiment classification. A comprehensive testing corpus supports translation and sentiment analysis tasks, with various machine learning models, such as Random Forest, Support Vector Machine (SVM), Decision Trees, and Gaussian Naive Bayes, trained to predict sentiment across these languages. The Random Forest model demonstrated robust performance, effectively capturing sentiment polarity while managing language-specific nuances. Additionally, BERT, a Large Language Model (LLM), is applied to predict context-based sentiment with high accuracy, achieving 99% accuracy and 98% precision, outperforming other models. BERT's predictions were further explained using Explainable AI (XAI), improving transparency and fostering trust in sentiment classification. The findings reveal that the proposed lexicon and machine learning models significantly enhance translation and sentiment analysis for languages spoken in South Africa and the DRC. This study provides a foundation for future AI models supporting underrepresented languages, with practical implications for education, governance, and business in multilingual contexts.

# 1. INTRODUCTION

Multilingualism poses a significant challenge for machine learning and sentiment analysis in translation, especially in linguistically diverse regions like South Africa, with 11 official languages, and the Democratic Republic of Congo (DRC), with 4. Most lexicons and language translating machine learning models cater to global languages, leaving languages like Zulu, Sepedi, Afrikaans, and Ciluba underrepresented. This gap results in disadvantages for existing AI systems in generating accurate translation and sentiment analysis in the local languages. Furthermore, countries like South Africa and the DRC with many languages face challenges in developing comprehensive tools capable of transitioning between local languages while ensuring sentiment accuracy is maintained. Given the current popularity and need for AI systems supporting local languages, particularly in education, government applications, and business in South Africa and the DRC is crucial. Sentiment analysis and machine learning models are crucial in developing inclusive AI systems (Mazibuko et al., 2021, 66-68). The current gap is also due to the limitations of machine learning models and sentiment analysis which can seamlessly handle Low-Resource Languages (LRL) like Zulu, Afrikaans, Ciluba, and Sepedi. Most known lexicons cater to languages like English and French but have insufficient data for LRLs like Zulu, Sepedi, Ciluba, and Afrikaans. In addition, the lack of LRL translation APIs poses challenges in the development of sentiment analysis and machine learning models that can provide accurate data when using LRLs. Similarly, there is a huge gap in the integration of the lexicons with machine learning models to provide improved sentiment accuracy and quality language translation for LRLs. Most trained machine learning models use well know High Resource Languages (HRL) and this leaves a gap in the AI system performance when working with African languages (Mazibuko et. al., 2021, 80-83).

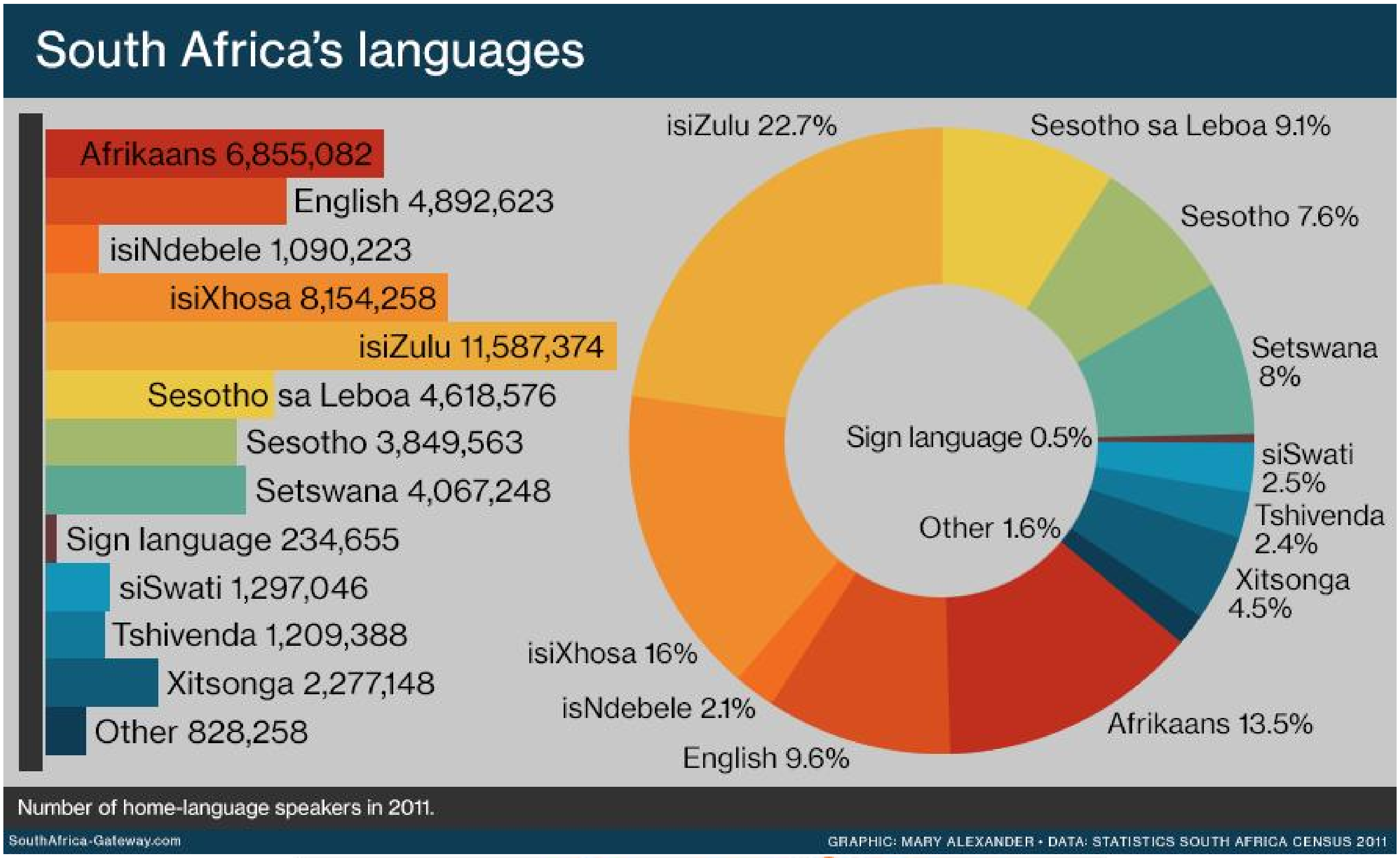

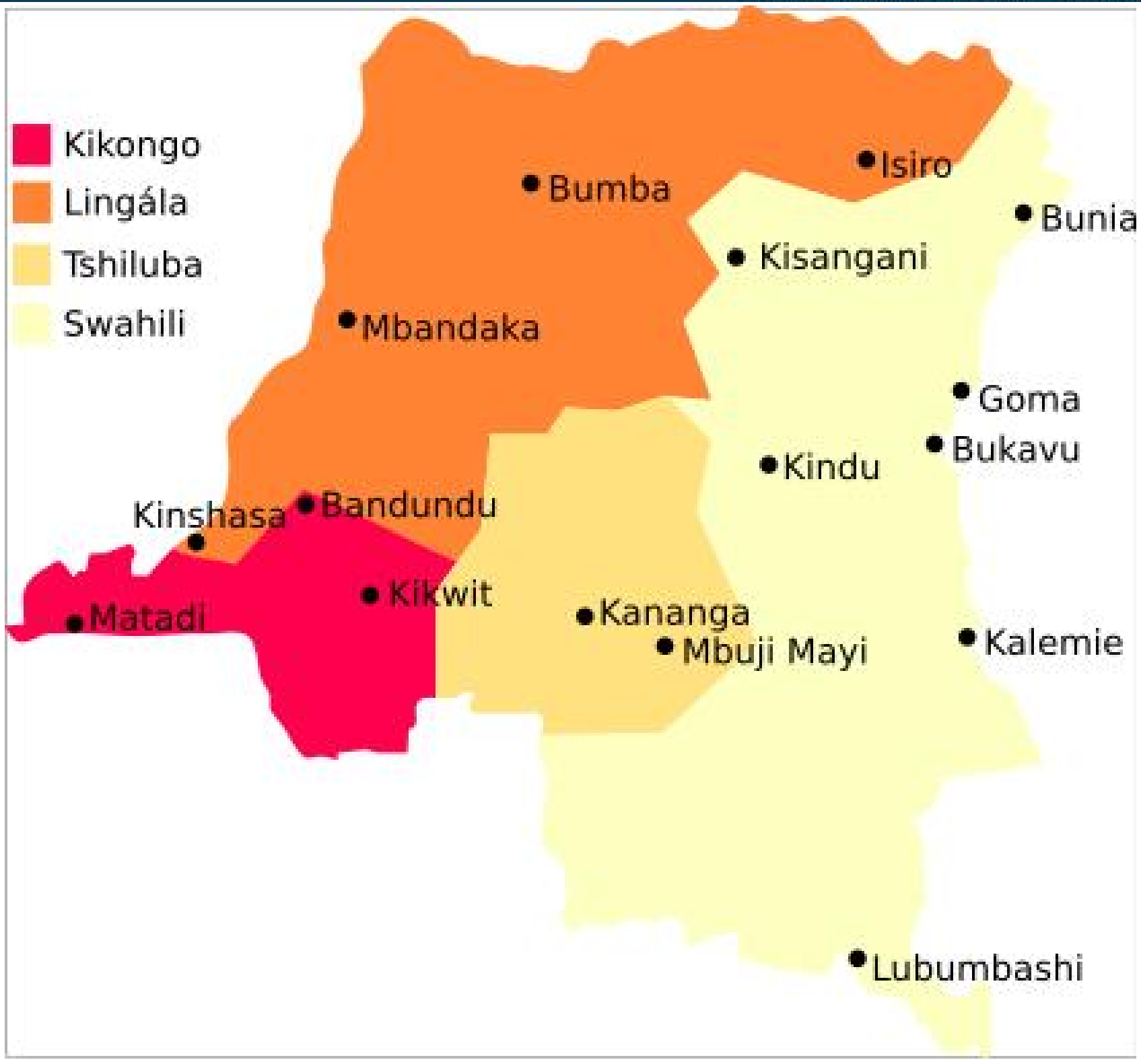

To address this gap, the research presents a multilingual lexicon designed for sentiment analysis and translation, initially constructed for French and Ciluba, and later expanded to include translations in English, Zulu, Afrikaans, and Sepedi. Each lexicon entry becomes annotated with sentiment polarity (tremendous, poor, or neutral) and intensity values starting from -9 to 9. To enhance translation accuracy and sentiment type, the experiments include Explainable AI (XAI) techniques and came up with a range of models, which includes Random Forest, SVM, Decision Trees, and Naive Bayes. These models were skilled and assessed using metrics along with recall, accuracy, precision, and the ROC curve. Cross-validation is employed to assess translation quality, where translation accuracy between languages was improved by verifying outputs across different translation systems. Our findings show that expanding the lexicon to include Zulu, Afrikaans, and Sepedi significantly improved sentiment analysis and translation accuracy, particularly for underrepresented languages like Ciluba. The machine learning models demonstrated varying levels of performance, with the Random Forest and Decision Trees models achieving the highest accuracy in sentiment classification. The use of XAI provided insights into model decision-making, enhancing transparency and trust in the models. Additionally, cross-validation between translation groups proved effective in improving translation accuracy, particularly when handling ambiguous terms in African languages. The combined use of lexicon expansion and machine learning models led to better performance in sentiment analysis for South African and the DRC languages, providing more nuanced and culturally appropriate translations. The implications of this study are very crucial for both African language translation and sentiment analysis for the development of AI systems in multilingual regions. Incorporating languages like Afrikaans, Zulu, Ciluba, and Sepedi into machine translation systems and sentiment analysis, will promote the inclusivity and accuracy of AI systems within South Africa and the DRC. Furthermore, using a combination of machine learning models such as Decision Trees, SVM, Random Forest, and Gaussian Naïve Bayes provides a robust framework for future sentiment analysis tasks in regions with diverse languages. The results suggest that expanding the lexicon and incorporating local languages can enhance the accuracy of AI systems in multilingual contexts. This project lays the groundwork for the development of lexicons, sentiment analysis techniques, and machine learning models, with potential applications in sectors such as education, digital communications, and government in Africa.

## 1.1 Background

In the new decade of globalisation and digitization, linguistic diversity is both a challenge and an opportunity for artificial intelligence (AI) systems, particularly in multilingual sentiment analysis and in machine translation. These tasks are integral to different applications, including social media monitoring, customer feedback analysis, and automated translation services, which increasingly demands accuracy across a multitude of languages. In South Africa, the complexity of language diversity is particularly pronounced with the eleven official languages - Afrikaans, English, Zulu, Xhosa, Sepedi, Setswana, Sesotho, Xitsonga, SiSwati, Tshivenda and isiNdebele, spoken by a population with lots of rich cultural backgrounds and incredibly unique linguistic challenges (Alshabi et al., 2021). In the DRC, the linguistic landscape is also highly complex and diverse. The country recognises four official languages: French, Lingala, Kikongo, and Swahili (Fraiture, 2024). In addition to these, there are over 200 languages spoken across the country, such as Ciluba, Kinyarwanda, and other regional languages (Fraiture, 2024). The rich cultural diversity in the DRC contributes to significant linguistic challenges, particularly in terms of translation, communication, and integration of local languages into official and technological systems. AI applications within South Africa and the DRC address this challenge to ensure inclusivity, accuracy, and cultural relevance across different sectors such as government, education and business, where language accessibility is essential to effectively serve many diverse communities.

Current sentiment analysis and machine translation systems predominantly cater to well-resourced languages, such as English, French and Mandarin, which have had extensive lexicons, labelled datasets, and computational resources, and literature supporting their integration into machine learning models (Chun et al., 2023; Brodowicz, 2024). Furthermore, South African and the DRC national languages are generally not as included as English and French throughout the computational linguistic sector. This places limits upon the capacity of AI to accurately capture the various differences of sentiment as well as various meanings across each of the languages (Chun et al., 2023). Furthermore, these languages generally contain culturally specific terms such as: idioms, contextual cues and unique words that are enormously different when compared to the global languages such as English and French. This complicates the AI system's ability to maintain appropriate translation values as well as sentiment accuracy when comparing and transitioning between different languages. This discrepancy not only affects individual AI tasks such as sentiment analysis but also creates restrictions on comprehensive multilingual AI models from offering decent linguistic support in diverse contexts (Brodowicz, 2024). Our approach of studying the topic of multilingual AI capabilities to further include different African languages is a complex and difficult task which requires a unique approach. Firstly, lexicons must be enriched with different sentiment labels as well as various polarity scores that correspond to contextual details that are unique to each African language. This allows for sentiment models to be further interpreted and to generalise to the emotional tone of phrases of that particular language as well as to accurately classify the different sentiment scores throughout the entirety of the lexicon. Secondly, the machine learning models must be custom made to handle the structural and grammatical differences of the underrepresented languages that all have unique prefixes and complex words to their unique language. For instance, within the Zulu language as well as other Bantu languages, there is a large degree of agglutination, meaning that they combine different affixes to create very new and complex words which express intricate meanings and details in a particular sentence being said or spoken. Furthermore, the machine learning models such as Random Forest classifiers do offer some potentially sound frameworks for handling these unique linguistic differences, as they are able to learn the constant patterns across diverse data inputs throughout the lexicon being trained upon. This creates unique challenges for various AI models as they have to be able to be trained appropriately on non-agglutinative languages such as, English and French (Chun et al., 2023). The integration of Explainable AI (XAI) methodologies, such as feature importance scores in tree-based models (Random Forest) or attention mechanisms in deep learning provides transparency throughout the decision-making process, allowing various stakeholders to understand how models interpret sentiment and translation outputs (Alshabi et al, 2021). For example, XAI can be used to highlight unique words or anecdotes that are able to influence a model's sentiment classification performance, flexibility and reliability. This allows analysts to uncover any potential biases or other inaccuracies that are specific to the local language's unique expressions. This analysis technique is crucial across the linguistically diverse countries in Africa, where the ability of AI systems to handle culturally different languages allows for the preservation of languages for future generations as well as allowing different people from different backgrounds to build trust (Brodowicz, 2024).

Furthermore, the products of this investigation are able to have other further implications for language AI applications across all of Africa. Especially in regions with linguistic diversity such as South Africa and the DRC. The addition of LRLs into AI systems will further improves the accuracy, sentiment analysis and machine translations of underrepresented African languages. Therefore, this research lays a foundational framework for further research and development of AI systems that can support the documentation of linguistic diversity, which promotes a more inclusive AI and communication landscape that is able to meet the demands of various communities in Africa. Furthermore, the utilisation of XAI provides crucial insight into a model performance and biases as well as accuracy metrics, that are essential for the responsible deployment of AI in business, education, and relevant government sectors, where these systems can facilitate an environment of clear communication across different languages and nations across Africa (Brodowicz, 2024). This study attempts to fill the current LRL translation gap by further developing a lexicon incorporating French, English, Zulu, Ciluba, Afrikaans, and Sepedi. Therefore, for every lexicon entry there is a labelled sentiment polarity (positive, negative, or neutral) as well as intensity values that range from -9 to 9, which allows for an assessment of sentiment across the different languages of South Africa and the DRC. This also allows for different machine learning models, such as: Naive Bayes, SVM, and Random Forest to be employed, to test and validate the accuracy of sentiment classification and translation tasks. Afterwards, Cross-Validation techniques were applied to enrich translation quality across different language groups. Thus, ensuring that outputs are linguistically and culturally accurate and translated well. With this, our study seeks to create an inclusive AI tool that is capable of performing accurate sentiment analysis with multilingual translation tasks in South Africa and the DRC's linguistically diverse environment by creating a lexicon with the integration of machine learning based XAI (Chun et al., 2023).

## 1.2 Problem Statement

South Africa' and the DRCs Linguistic Diversity presents a challenge for AI translation and sentiment analysis systems. There is a need for accurate, labelled data for LRLs spoken in these countries such as Zulu, Ciluba, Sepedi, and Afrikaans. The lack of labelled data limits the application of AI systems to these languages and in turn could hinder their long-term preservation. This study aims to begin to fill this gap by creating labelled training data for African languages while performing sentiment analysis on the created data.

## 1.3 Objectives

In order to achieve the study's aims the following objectives will be addressed:

1. Create a lexicon that includes English, French, Afrikaans, Ciluba, Sepedi, and Zulu translations while ensuring accuracy of the assigned sentiment and translations.
2. Expand the lexicon by adding language specific sentiment scores to translated words to account for cultural context in sentiment.
3. Create a corpus testing dataset for translations and sentiment analysis.
4. Train machine learning models to predict sentiment and to translate between languages in the lexicon.
5. Implement existing LLM architecture to predict context-based sentiment of individual words, while using XAI to explain predictions.
6. Evaluate the study's findings while making recommendations for future work.

# 2. LITERATURE REVIEW

There is an increase in the usage of AI in natural language processing (NLP) for processes such as sentiment analysis, however, in countries such as South Africa and the DRC, there is still a shortage of NLP resources. Hence, there is an opportunity to develop AI tools that integrate sentiment analysis and translation to help preserve LRLs in Africa.

## 2.1 Language Preservation and Low-Resource Languages in NLP

It is of utmost importance that we preserve LRLs in diverse, culturally rich, and multilingual countries such as South Africa and the DRC. Developing NLP technologies will help preserve low-resource languages by offering economic development, preventing language extinction, and encouraging the growth of these languages beyond the African borders. It is therefore important that we preserve these African languages using artificial intelligence tools and advancing technological applications in these areas (Magueresse et al.,202). Issues that arise in in preserving low-resource languages include the lack of availability of multilingual datasets which creates a demand for the creation of multilingual lexicons and sentiment datasets, this will help in preserving low-resource languages and increase their use in AI-driven systems (Gambäck et al.,2017). In addition, being able to develop these lexical resources specifically for languages like Sepedi, Ciluba, Zulu, and Afrikaans can contribute to their digital presence. Joshi et al. (2020) argues that creating NLP tools and lexicons for low-resource languages can help preserve these languages while improving AI's ability to be able to process multilingual data; this will help ensure that sentiment analysis reflects the cultural and linguistic diversity of African countries.

## 2.2 Sentiment Analysis in NLP

Sentiment analysis is the process of collecting and gathering text about people's views and opinions on certain matters to later analyse how the users or people feel about a certain topic, experience or product (Wankhade et al., 2022; Nkongolo Wa Nkongolo, 2023). Sentiment analysis over the years has moved from utilising lexicon-based methods to more advanced approaches including machine learning, deep learning, and transfer learning methods (Nkongolo Wa Nkongolo, 2023; Birjali et al., 2021). The applications of sentiment analysis include analysing customer reviews to gain insights on how consumers feel about a certain product or service , the monitoring of people's comments and other interactions on social media to gain perspectives on how different users feel about a current trend or topic and to conduct market research (Birjali et al., 2021). Over the course of many years there have been improvements in the method of which sentiment analysis has been performed. However, there are still a few problems such as the interpretation of different meanings of words in different types of contexts such as the various use of homonyms and homophones (Nkongolo Wa Nkongolo, 2023). Furthermore, AI systems have further struggled to detect sarcasm and negation. This has made sentiment analysis incredibly difficult to classify as well as analyse various sentiments as positive, negative or neutral (Markovml, 2024; Nkongolo Wa Nkongolo, 2023). We propose a unique challenge for sentiment analysis within South Africa and the DRC in the context of different meanings that can belong to the same word that is dependent on the context in which the word has been used.

## 2.3 Machine Learning in NLP

Machine Learning in NLP consists of algorithms and models that can be used to understand human text in processes such as sentiment analysis. The models and algorithms used are trained to identify the trends and patterns found in text to understand the sentiment behind the texts. In NLP , machine learning methods include supervised learning, unsupervised learning, semi-supervised learning, and deep learning methods. Supervised learning includes linear classifiers and probabilistic classifiers, models such as Random Forest that make use of a linear classifier approach, while Naive Bayes makes use of a probabilistic classifier (Saravanakumar, 2023). A study by Rahat et al. (2019) used a review dataset to compare the performance of Naive Bayes and SVM by evaluating each model's ability to perform sentiment analysis on the reviews of an airline regarding the customer's experience. The reviews contained either positive or negative experiences from the customer on which sentiment analysis was carried out. The SVM classifier outperformed the Naive Bayes algorithm in correctly identifying the sentiment in the reviews. Another study seeking to compare the performance of SVM and Naive Bayes in identifying hate speech or non-hate speech showed high performance of the SVM classifier (Asogwa et al. 2022). It seems as though the SVM classifier performs well in sentiment analysis where the dataset is binary in nature and it is able to correctly classify sentiments as either positive or negative. However, it might not yield the same result when the nature of the dataset is more complex. To compare the performance of Random Forest to SVM, a study by Khan et al. (2024) found that SVM still outperformed Random Forest using a dataset to perform sentiment analysis. Although SVM seems to be performing really well in differentiating between positive and negative texts according to accuracy metrics, the best choice of a model to perform sentiment analysis will vary according to the nature of the dataset and its complexity.

## 2.4 Explainable Artificial Intelligence (XAI)

Explainable AI in NLP generally focuses on the development of methodologies to ensure that NLP models are more interpretable and transparent, thus, helping users to understand how the models make decisions or generate relevant outputs. According to Khun et al (2021) explainable natural language processor (XNLP) is an interactive browser-based system embodying a living survey of recent state of the art research in the field of XAI within the domain of NLP. Furthermore, these systems are incredibly important since the NLP models, especially deep learning complex ones, can function as black boxes so to speak, making it difficult to understand how they arrive at specific predictions or classifications. XAI can help us understand these so-called black-boxes. An example of this is that when AI system rejects a loan application, the applicant should be entitled to understand why that decision has been made to ensure that the compliance with laws and regulations are met. This helps the user to understand what needs to be achieved in terms of being compliant with regulatory regulations (Saeed & Omlin, 2023).

## 2.5 BERT in Contextual Language Modelling

BERT can be defined as a transformer-based language model that utilises a bidirectional approach to linguistic modeling, allowing it to be trained to understand text in both directions—forward and backward. This bidirectional capability distinguishes BERT from traditional unidirectional language models (Cesar et al., 2023). BERT also utilises a masked language modelling (MLM) approach that is able to randomly mask out certain words in a sentence that allows for a model to be trained and to predict the masked words which makes it unique from traditional language models (Cesar et al., 2023; Rokon, 2023). Masked Language Modeling (MLM) enables BERT to learn context in a non-linear manner, making it highly robust and reliable in understanding the intricate relationships between words and phrases. Additionally, BERT employs Next Sentence Prediction (NSP), which allows the model to determine logical sentence flow, helping it capture relationships at the sentence level. This capability is essential for BERT to handle complex language tasks effectively (Rokon, 2023)

# 3. METHODOLOGY

## 3.1 Lexicon Expansion

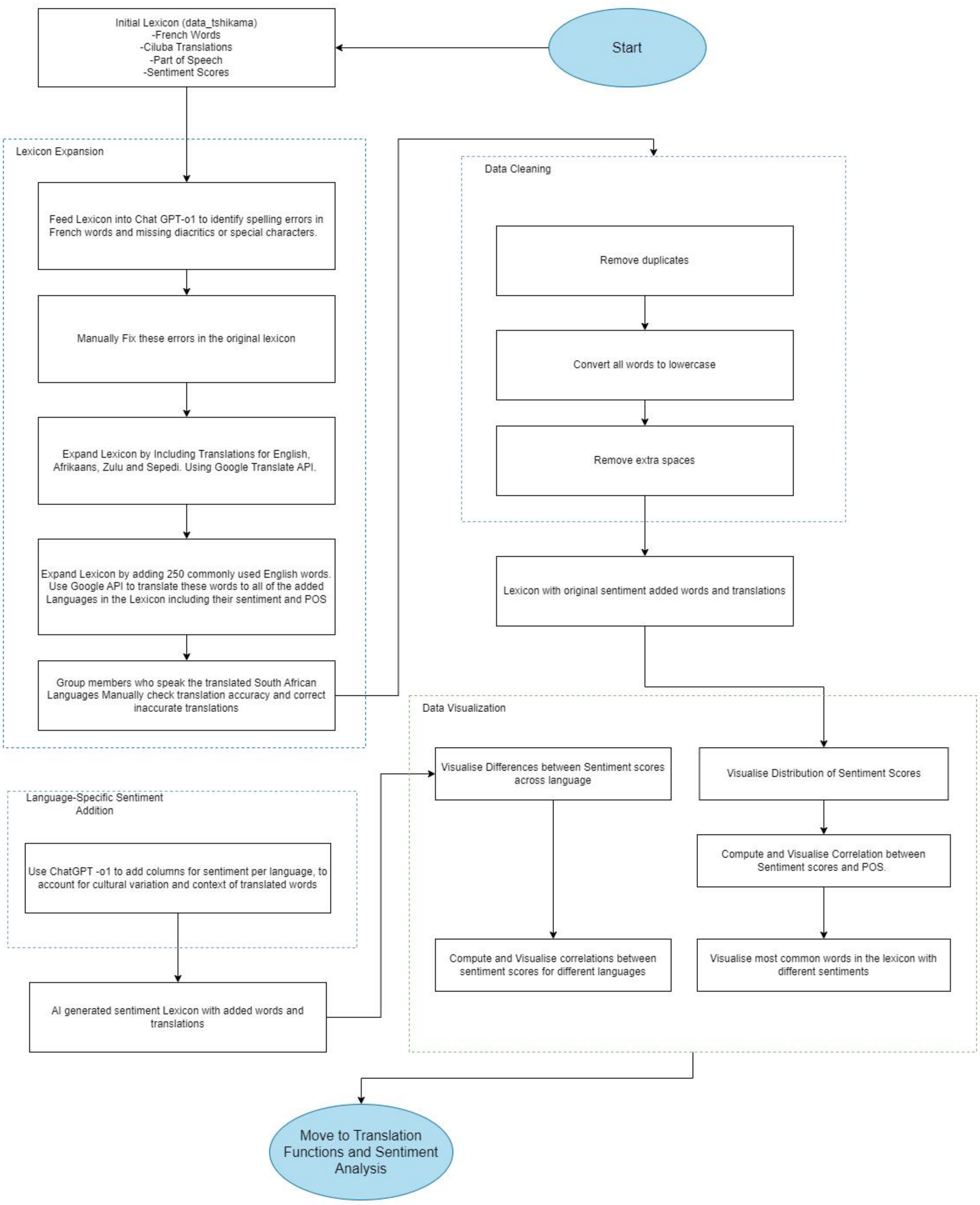

*Figure 1 Lexicon Expansion Flowchart.*

The Figure 1 outlines a process for expanding and refining a multilingual lexicon with sentiment scores for various languages, followed by data visualisation to analyse sentiment trends across languages.

### 3.1.1 Original Lexicon Cleaning

The initial lexicon contained just over 3000 French and Ciluba words with an assigned sentiment score, the part of speech (POS), and the words categorical sentiment (neutral, negative, and positive). With none of the authors being able to speak either of these languages it was difficult to identify any potential spelling errors or missing special characters. Therefore, the lexicon was fed to ChatGPT 0-1 preview, which has advanced reasoning capabilities and is able to deal with large amounts of text, to identify any errors in the French words before these errors are carried downstream in the translation process. The identified errors were corrected manually on the lexicon in Microsoft Excel. Exact duplicates, the same word with the same sentiment and part of speech, were removed from the data frame and any trailing or leading blank spaces were removed. This allowed for a clean lexicon to be expanded to South African and the DRC languages.

### 3.1.2 Lexicon Expansion

The Lexicon was expanded in both columns and rows. Firstly, Google Translates' API was used to translate the French words into English, which all authors can speak and could therefore identify errors. Thereafter 250 commonly used English words that were not present in the lexicon were added to create a more comprehensive set of words. The Google Translate API was then again used to translate the English words, including those added, into three South African languages namely, Afrikaans, Sepedi, and Zulu, extending the lexicon to not only have more languages but also more words. The authors who could speak the respective translated languages then manually went through the lexicon to identify and fix errors in translations from the google translate API, resulting in an accurately expanded *Lexicon*.

### 3.1.3 Language Specific Sentiment Addition

It was noticed through the process that there was only a single sentiment score across the Lexicon for all of the languages, which raised the question whether this would be an accurate representation of sentiment for each of the languages. In an attempt to capture sentiment for words that could differ across languages due to cultural significance, the expanded lexicon was given to ChatGPT 0-1 to generate sentiment scores for each word per language. The new version of the LLM has the ability to process these languages and it was therefore hypothesised that it would be able to add sentiment scores to the words. To evaluate the difference in sentiment across languages a new lexicon was created which contained sentiment scores per language, to serve as a comparison for the expanded lexicon with a single sentiment score across all languages.

### 3.1.4 Exploratory Data analysis (EDA)

EDA was performed on the lexicon with the language specific sentiments in order to determine if there was any effect on sentiment based on the language (Figure 2). The correlation between the original sentiment values and the language specific sentiment values were calculated and visualised. To gain an understanding into the distribution of sentiment in the expanded lexicon. And where there are potential gaps in the data, the distribution of sentiment scores and the parts of speech were visualised as well. With a cleaned expanded lexicon and an understanding of the data sentiment analysis, machine learning could be performed.

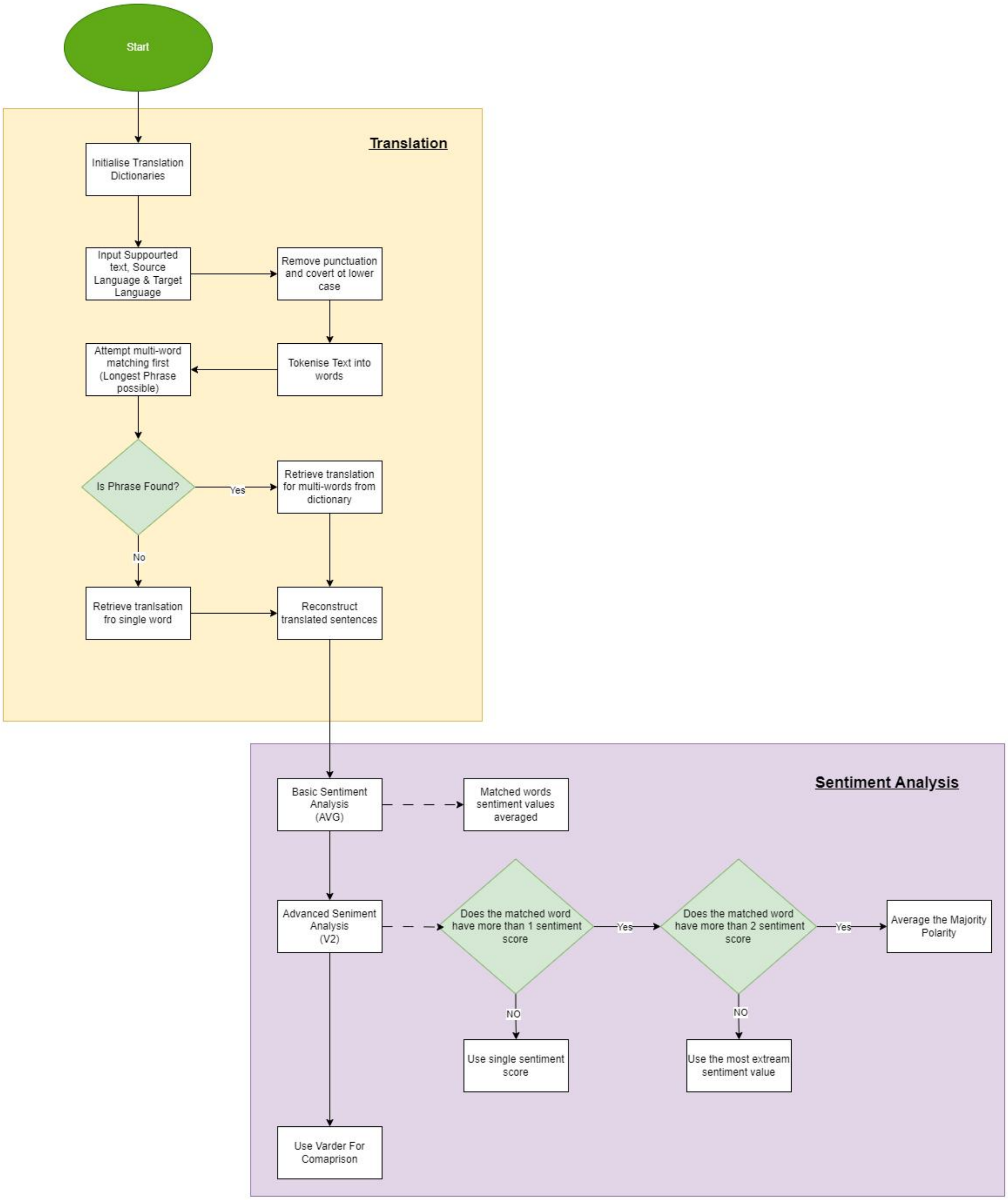

*Figure 2 Flowchart Showing Translation and Sentiment Analysis Phases.*

## 3.2 Sentiment Analysis with Machine Learning

After the cleaned data was formed from the EDA the research team proceeded to train a Random Forest, SVM, Gaussian Naive Bayes and a Decision tree classification model. We firstly started off by utilising the polarity values from -9 to 9 for each word row across our cleaned lexicon where each row had scores assigned to the Positive, Negative or Neutral sentiment of each word. Therefore, for each word there was a polarity value as well as an attached sentiment label. This allows each of our chosen models to be enriched with a standardised score as well as a standardised label. This allows for each model to generate predictions from a normalised dataset for different languages. This was generated to allow for words of similar sentiment scores to predict other words from another language with similar sentiment and scores. For example, trees from English being similar to boom in Afrikaans with a similar sentiment score as well as a similar sentiment label. Therefore, in our cleaned dataset each word and its relevant part of speech was able to be enriched with this standardised data to allow for the Random Forest, Support Vector, Gaussian Naive Bayes, and Decision tree models to have an accurate basis for predicting and translating words throughout the lexicon. This allowed us to split our datasets in terms of an 80 % training dataset and a 20% validation dataset. We also created a confusion matrix to identify how well the model classified true values from predicted values. Furthermore, each of the classes were based on the word's part of speech and was used for accurate depiction within the confusion matrix for each of the models. After this, various accuracy metrics were also calculated such as overall accuracy, precision, recall and F1-scores for each model created to appropriately inspect which algorithm performed the best. After we identified the best model, we further deployed it in the translations system.

## 3.3 Translations and Sentiment Analysis

In this section we look at translations performed with sentences created using the expanded lexicon. Once the sentences have been created, the translation is performed from the source language into the user selected target language. Once the translation is performed a sentiment analysis is conducted, to better understand the sentiment of each sentence. Each word in the lexicon has a given sentiment score that is used in the analysis. Two main methods of sentiment analysis are performed, namely averaging the sentiment scores and a more advanced stepwise logic. There are sometimes multiple repeat words in the lexicon with the same translations, only differing in sentiment score. This is due to words being used in different contexts having different sentiments (Yang and Chao, 2018). The simple sentiment analysis calculates sentence sentiment by adding all the sentiment scores of the first matched words/phrases sentiment scores and averaging them, regardless of the remaining words' sentiments. To account for the other sentiment values of words in different contexts, a more advanced method of analysing the sentiment was employed. The more advanced technique, V2, determines if there is one, two or more exact words with different sentiment scores (Elmurngi and Gherbi, 2017). If there is a single match, then the single score is used; if there are two scores, the more extreme of the values are used; if there are three or more matched words with differing sentiment scores, the majority polarity group is averaged for a generalised score in the dominant polarity (positive or negative). The Vader labelling logic tool is used using its built-in lexicon that considers intensity of words (Barik and Misra, 2024). However, the lack of punctuation reduces the effectiveness of the Vader tools.

## 3.4 Aspect-based Sentiment Analysis

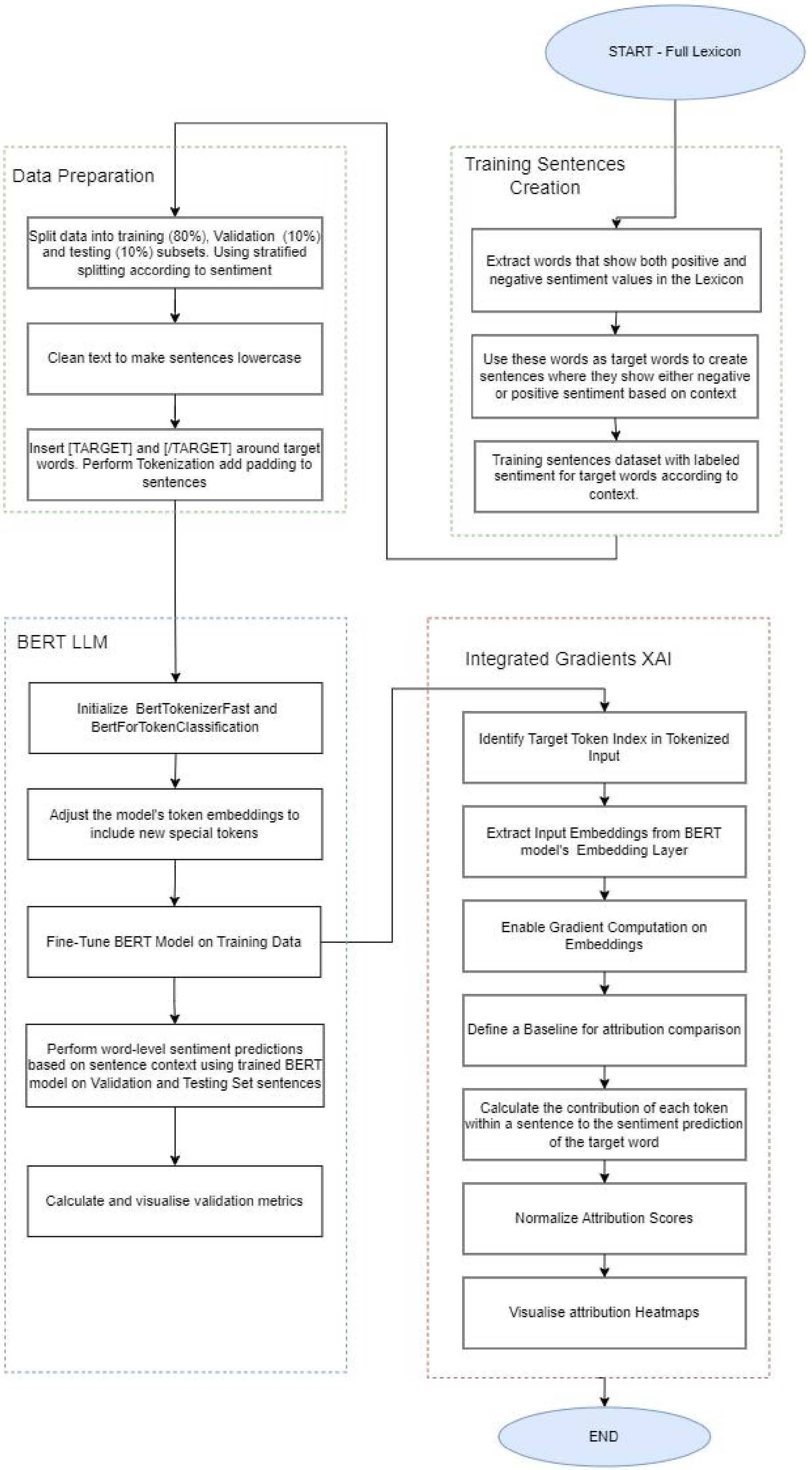

*Figure 3 Flowchart Showing Aspect-based Sentiment Analysis.*

Figure 3 details the methodology employed for sentiment analysis on individual words that could carry different sentiment depending on the context they are used in. For this task, words that could represent either positive or negative sentiments, as represented by repeating occurrences in the lexicon with different sentiments, were chosen and sentences were created with these words used in different sentiments. Existing large language models (LLM) architecture, namely BERT, was used to predict what sentiment should be assigned to a word based on the context used.

### 3.4.1 Testing Dataset Creation

Words in the lexicon that exhibited both positive and negative sentiments were identified as context-dependent, capable of conveying different sentiments based on their usage. These words were selected as target words, and sentences were constructed using these targets alongside other lexicon words. The sentences were crafted to reflect either a negative, neutral, or positive sentiment, depending on the context in which the target words were used. A total of 1,000 sentences were generated, with each target word labeled according to its sentiment classification in the lexicon.

### 3.4.2. Testing Set Preparation

To allow models to focus on the specific target word within the sentence, each target word for a sentence was marked with a token [Target] and closed with [/Target] tokens. For example, the sentence '*Earth is the third planet from the sun*' if earth is the target word it would be marked with the tokens [Target] *Earth* [/Target]. Tokenization was then applied to the test sentences using BERT's tokenizer (BertTokenizerfast), which separated sentences into tokens, breaking up the sentences into words with the words around the target word defining the sentiment that the target word is assigned (Vayadande et al., 2024). Each of the words within each sentence were assigned a sentiment score through the BERT tokenizer. The testing set was split into 70:20:10 for training, validation and testing sets respectively.

### 3.4.3 Model Implementation

The BERT model, which is an LLM, was called using a HuggingFace API (Sindane and Marivate, 2024). This was chosen for this task due to its proven strong performance and wide use in NLP. BERT processes text in both directions looking at the words before and after the target word in the sentence to determine the sentiment that is assigned to the word (Sindane and Marivate, 2024; Vayadande et al., 2024). The model also leverages a pre-trained architecture, enabling it to apply the insights gained from its training data while being fine-tuned for sentiment prediction tasks.

### 3.4.4 Model Training

A trainer class from Hugging Face's Transformers library was employed to manage the training loop, and the BERT model was trained to predict the sentiment of a target word based on the context in which it appeared within a sentence (Sindane and Marivate, 2024). Upon initial training it was noticed that there was a lack of neutral testing sentences as a class imbalance. To address this, the class weights were adjusted to ensure that the neutral class was not being dominated by the majority negative and positive classes, assigning greater importance to neutral sentences.

### 3.4.5 Evaluation

To understand the performance of the BERT-based sentiment analysis model, evaluation metrics were calculated and visualised. The accuracy, precision, recall and F-1 Score were calculated and visualised in a table for each of the sentiment classes (Rainio, Teuho, and Klén, 2024). A confusion matrix was presented showing the actual classes of the validation set and the predicted classes. ROC graphs were plotted to visualise how the model performs in discriminating between sentiment classes at different discrimination thresholds, giving a line on the graph for each class vs the other classes (Rainio, Teuho, and Klén, 2024).

### 3.4.6 XAI Evaluation

To better understand the model's decision-making process in assigning sentiment to a word based on context, Integrated Gradients, a post-hoc XAI technique, is implemented using the Captum library (Sithakoul, Meftah, and Feutry, 2024). As a post-hoc method, it was applied after predictions were made to explain the reasoning behind those predictions. Integrated Gradients were used as it is a model agnostic, meaning there was no need to change the architecture of the BERT model for explanations to be made using the XAI technique. For each test sentence, the target word was marked using the same [Target] tokens as described in section 3.4.2, which placed focus on the target word for attribution calculations within the sentence. The input embeddings corresponding to the tokenized sentence were extracted from BERT's embedding layer and gradient computation allowed for the attributions of each token in the sentence with regard to its influence in sentiment prediction of the target word. The attributions were visualised using heat maps to show the attribution scores of each word in a sentence, tokens with higher attribution scores would indicate that they place a large role in determining the sentiment of the target word.

# 4. RESULTS

## 4.1 Exploratory Data Analysis

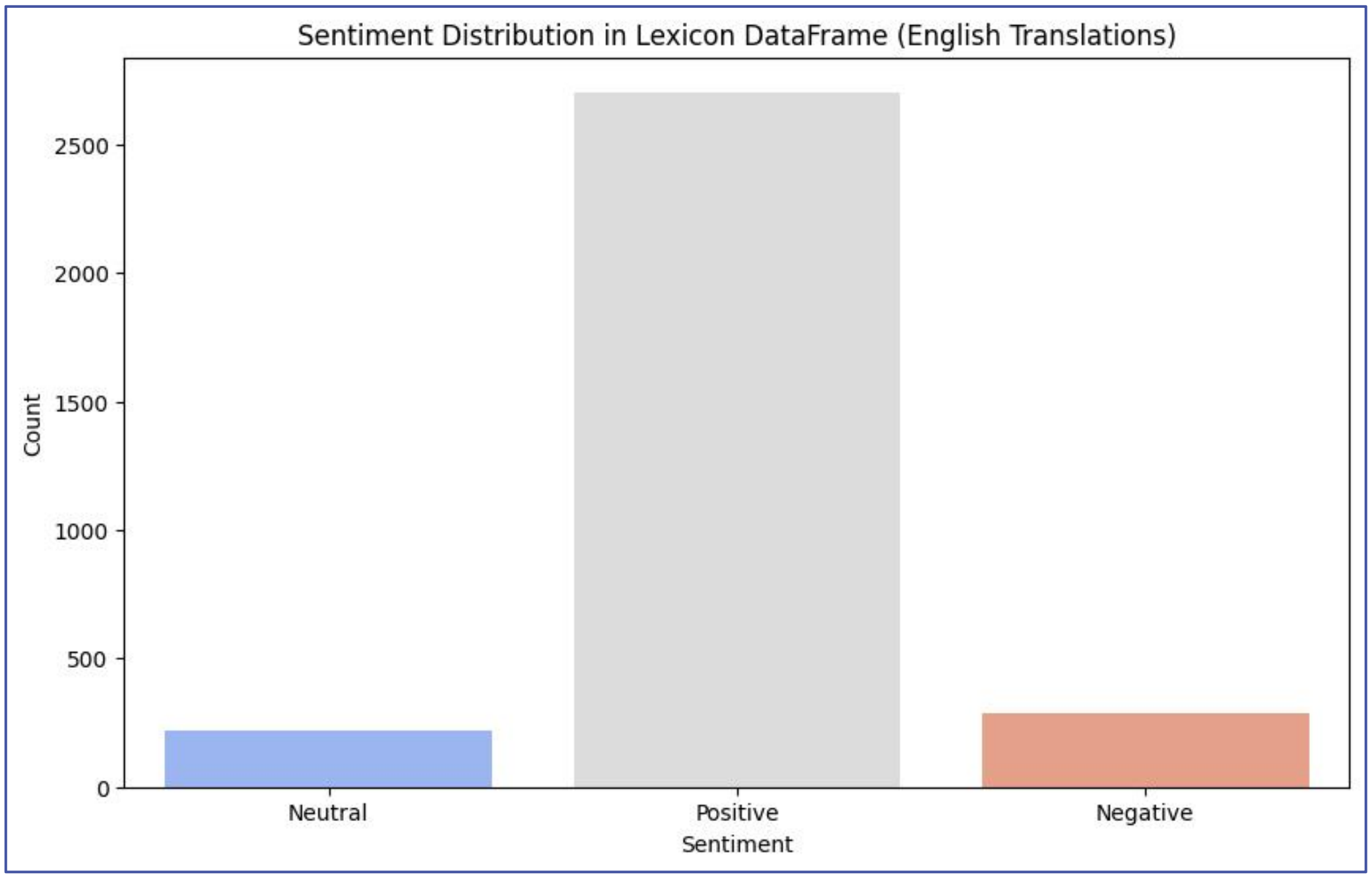

*Figure 4 Sentiment Score in the Lexicon.*

Figure 4 reveals the distribution of Neutral, Positive and Negative sentiment values for a lexicon after being translated to English.

**Positive sentiment.** The bar chart shows that there are more positive words on the lexicon. There are approximately 2500 positive words in the lexicon, and this makes the positive sentiment dominant class in this lexicon. This skew is because of the positivity being many compared to the other words.

**Neutral and negative sentiment.** Words categorised under this sentiment are fewer compared to the positive sentiment. There are about 200-300 words under the neutral and negative category, and this indicates that there are few words to create a negative or neutral sentence. This imbalance means more positive emotional sentences can be created rather than negative or neutral sentences. More neutral and negative words were manually added to balance the distribution and the skewness of the bar chart.

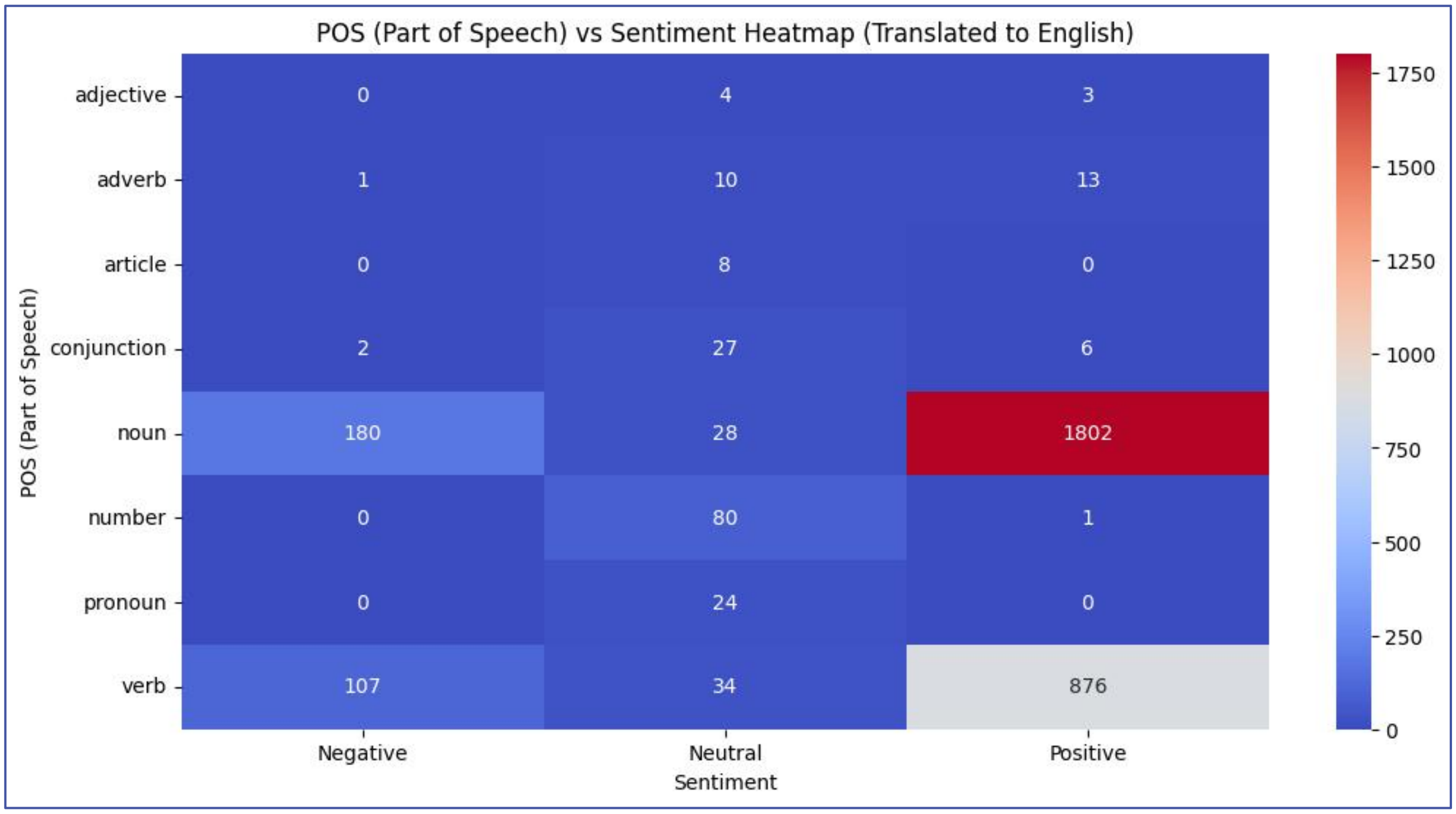

*Figure 5 Heatmap Showing Parts of Speech Against Sentiment.*

The heatmap shown in Figure 5 provides a visual representation of the relationship between Parts of Speech (POS) and sentiment classifications for the lexicon. Negative, neutral, and positive within a translated lexicon results are shown by the heatmap. On the vertical axis, various POS categories, such as nouns, verbs, and adjectives, are displayed, with colour intensity indicating the volume of words in each sentiment category (Figure 6). Notably, nouns dominate the positive sentiment category, with over 1800 classified as positive, aligning with earlier sentiment distribution, findings that favoured positivity. In contrast, only about 180 nouns are categorised as negative, and a mere 28 as neutral.

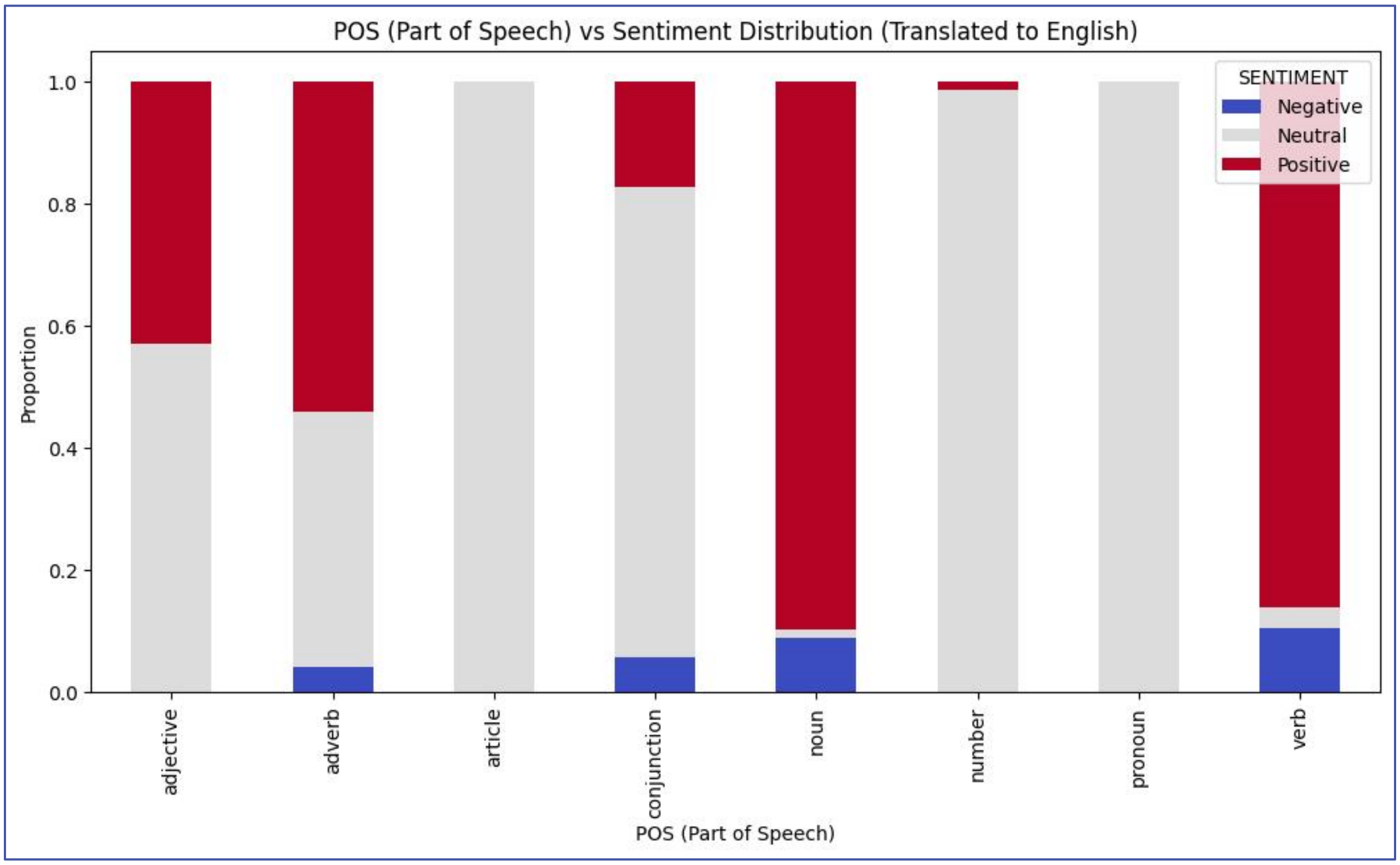

*Figure 6 POS Against Sentiment Distribution.*

This prevalence of positive nouns suggests that the lexicon predominantly reflects favourable objects, people, or concepts. Verbs also exhibit a positive sentiment trend, with more than 876 classified positively, while negative and neutral verbs are less common, numbering 107 and 34, respectively. This indicates that the actions represented in the lexicon are generally perceived positively. In contrast, adjectives and adverbs show minimal representation across sentiment categories, with only a few words classified as positive or neutral (Figure 6). This lack of descriptive terms points to the limitations in the lexicon. Additionally, other POS categories, such as conjunctions, pronouns, articles, and numbers, show minimal sentiment representation, which is expected as these words primarily serve grammatical functions rather than conveying sentiment on their own. This stacked bar chart shows the proportions of negative, neutral, and positive sentiments across different POS categories (Figure 6). The stacked bar chart reveals how different parts of speech (POS) express varying sentiment negative, neutral, and positive. This analysis aligns well with our earlier findings. Notably, nouns and verbs emerge as positivity, with verbs standing out for their overwhelmingly upbeat tone. This suggests that action-oriented language tends to be favourable within this dataset. In addition, adjectives and adverbs present a more mixed of sentiments. While they show both positive and neutral sentimentality, adverbs also introduce a slight hint of negativity. This indicates that although these descriptive words add emphasis, they do not always clearly signal a positive or negative sentiment, adding a layer of complexity to how we interpret them in sentiment analysis. Articles and pronouns remain firmly neutral, given their typical role in providing structure rather than emotional nuance (Figure 6). Conjunctions also show a slight presence in both positive and neutral sentiments but generally lean towards neutrality, reinforcing their function as connectors rather than sentiment bearers. Overall, the prevalence of neutral sentiment in certain POS types underscores their structural role, while nouns and verbs more directly influence emotional expression (Figure 7).

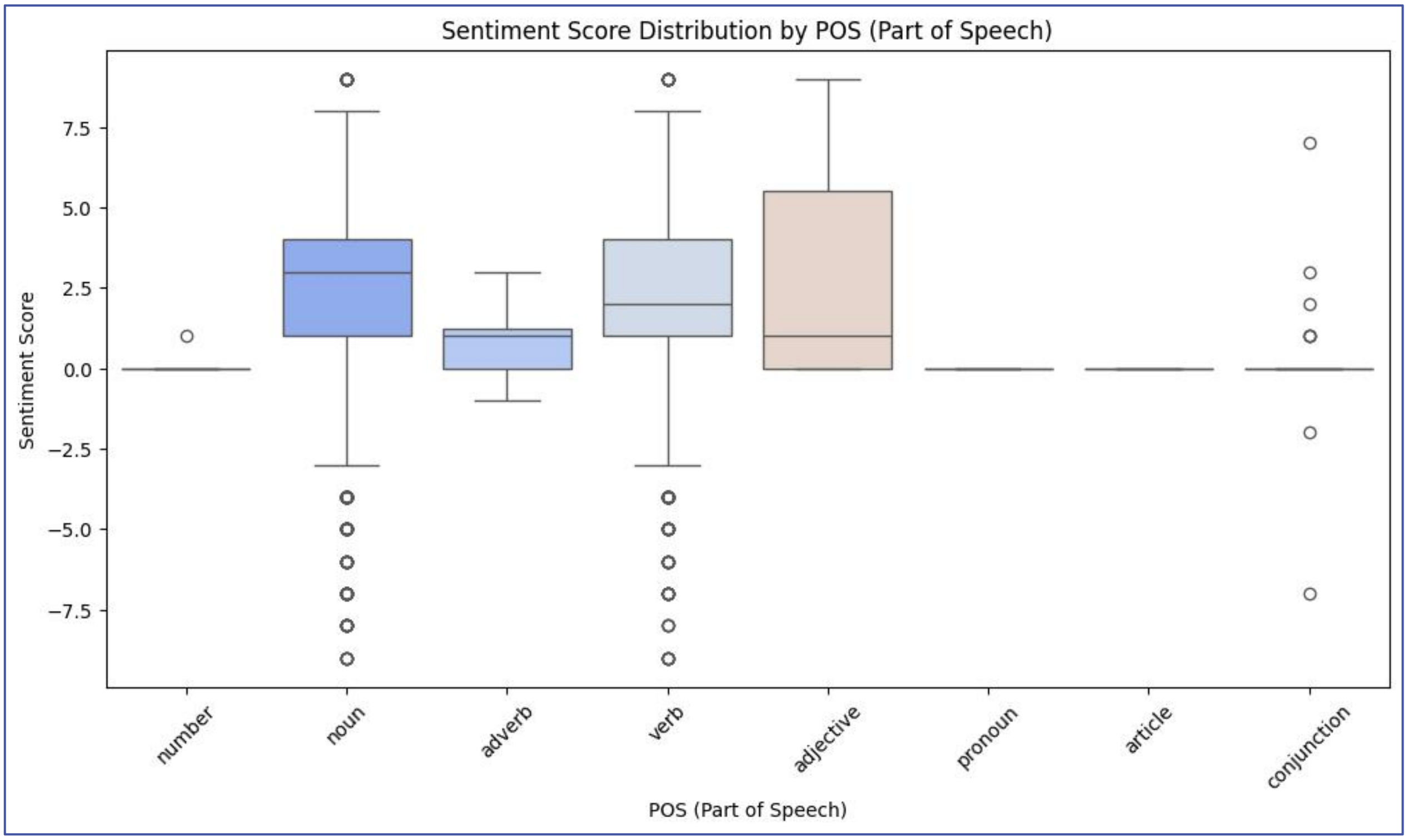

*Figure 7 Box Plot Showing the Sentiment Score Variability Within Each POS Category.*

The box plot shown in Figure 7 offers valuable insights into the variability of sentiment scores across different Parts of Speech (POS) categories. Nouns and verbs emerge as key players, with nouns demonstrating a wide range of positive sentiment, while verbs show a generally high positive median but also include some notable negative outliers (Figure 7). This indicates that nouns consistently convey positive sentiment, whereas verbs, although predominantly positive, exhibit some variability likely due to the contextual differences in the actions they describe. Adjectives also reveal a high median sentiment score, reflecting their role in expressing strong sentiments, often positive though there is some variability in their usage. In contrast, adverbs display a narrower range around a positive sentiment, with occasional negative outliers suggesting that while they tend to be positive, they can introduce negative nuances depending on the context (Figure 7). Functional words such as conjunctions, pronouns, and articles show little variance in sentiment, clustering tightly around neutral scores. This aligns with their primary grammatical role, rather than contributing emotional weight. Overall, the distributional insights from this plot highlight that nouns, adjectives, and verbs play significant roles in shaping sentiment, while functional words like articles and pronouns have minimal impact on sentiment expression (Figure 8).

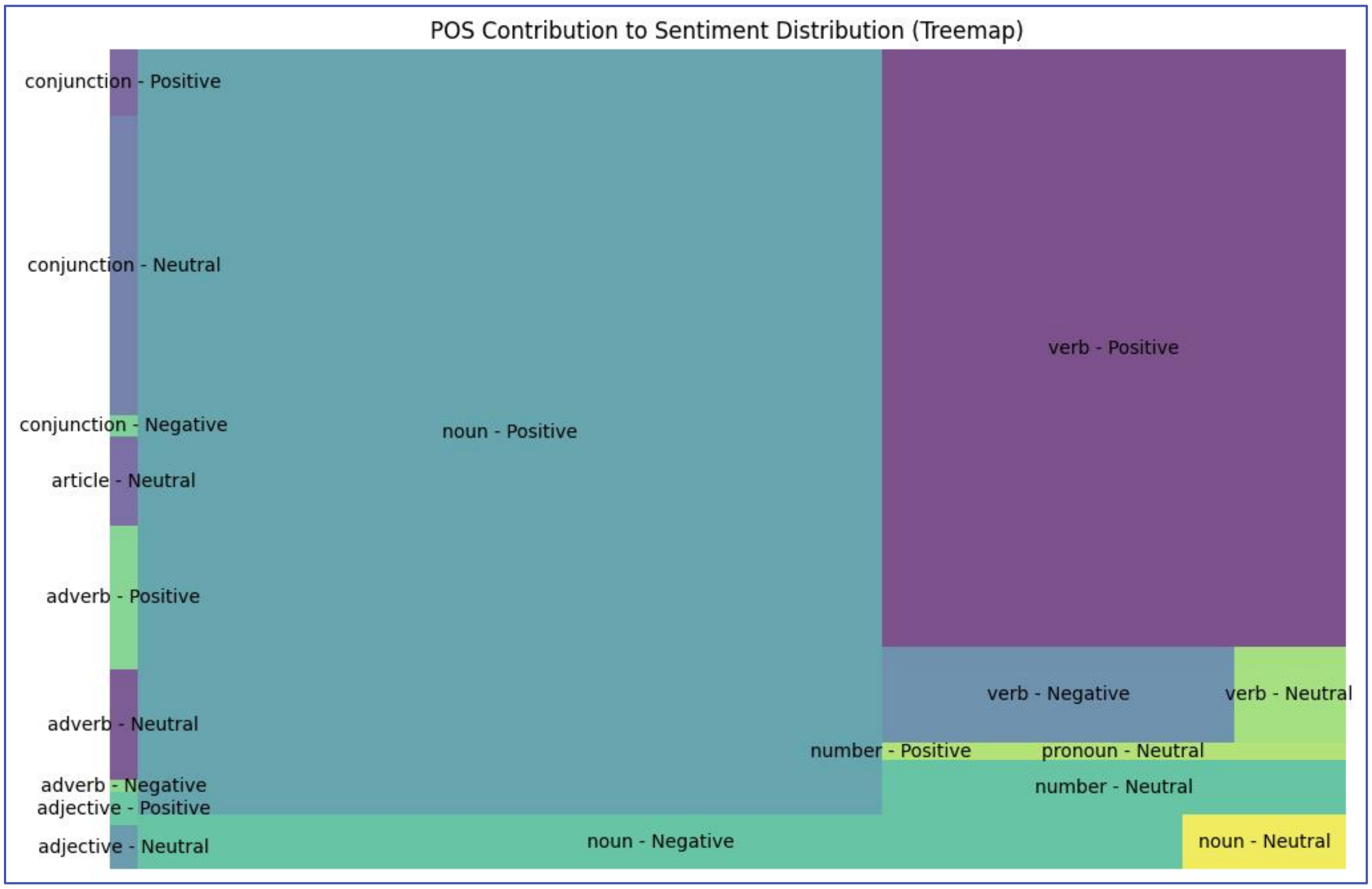

*Figure 8 Sentiments per Word based on Significance.*

The Lexicon contains different sentiments per word based on the significance in the languages based on cultural significance (preserving the meaning of the word across different languages). So Figure 8 visualisations look at the difference between the sentiment for different languages. The treemap offers a visual breakdown of each Part of Speech (POS) type's contribution to overall sentiment, categorised into positive, negative, and neutral sentiments (Figure 8). Nouns occupy a significant area within both the positive and neutral sentiment sections, highlighting their role as primary carriers of sentiment in the Lexicon. This observation aligns with their high frequency and substantial contributions noted in other analyses. Similarly, verbs constitute a large portion of the positive sentiment area, reinforcing the earlier finding that action words in this Lexicon tend to skew positively.

This trend suggests that the Lexicon favours verbs that express constructive actions and positive outcomes. Adjectives and adverbs appear in both the positive and neutral sections, with smaller representations in the negative area, confirming their mixed sentiment contributions. Furthermore, they generally support positive sentiment, they also add to neutrality and occasionally introduce negativity. In contrast, articles, pronouns, and conjunctions remain largely neutral, confirming their limited influence on sentiment analysis, consistent with their grammatical function rather than a semantic one. This analysis underscores how different parts of speech contribute to sentiment in distinct ways, revealing the underlying structure of language where content words drive emotional meaning while function words primarily provide grammatical structure. Together, figures 7-10 visualisations illustrate that nouns and verbs are the main drivers of sentiment, especially positive sentiment, indicating that content words serve as the primary bearers of emotional meaning. Adjectives and adverbs provide nuanced sentiment, often leaning positive but exhibiting some variability, which reflects their role in enhancing sentiment intensity and emphasis.

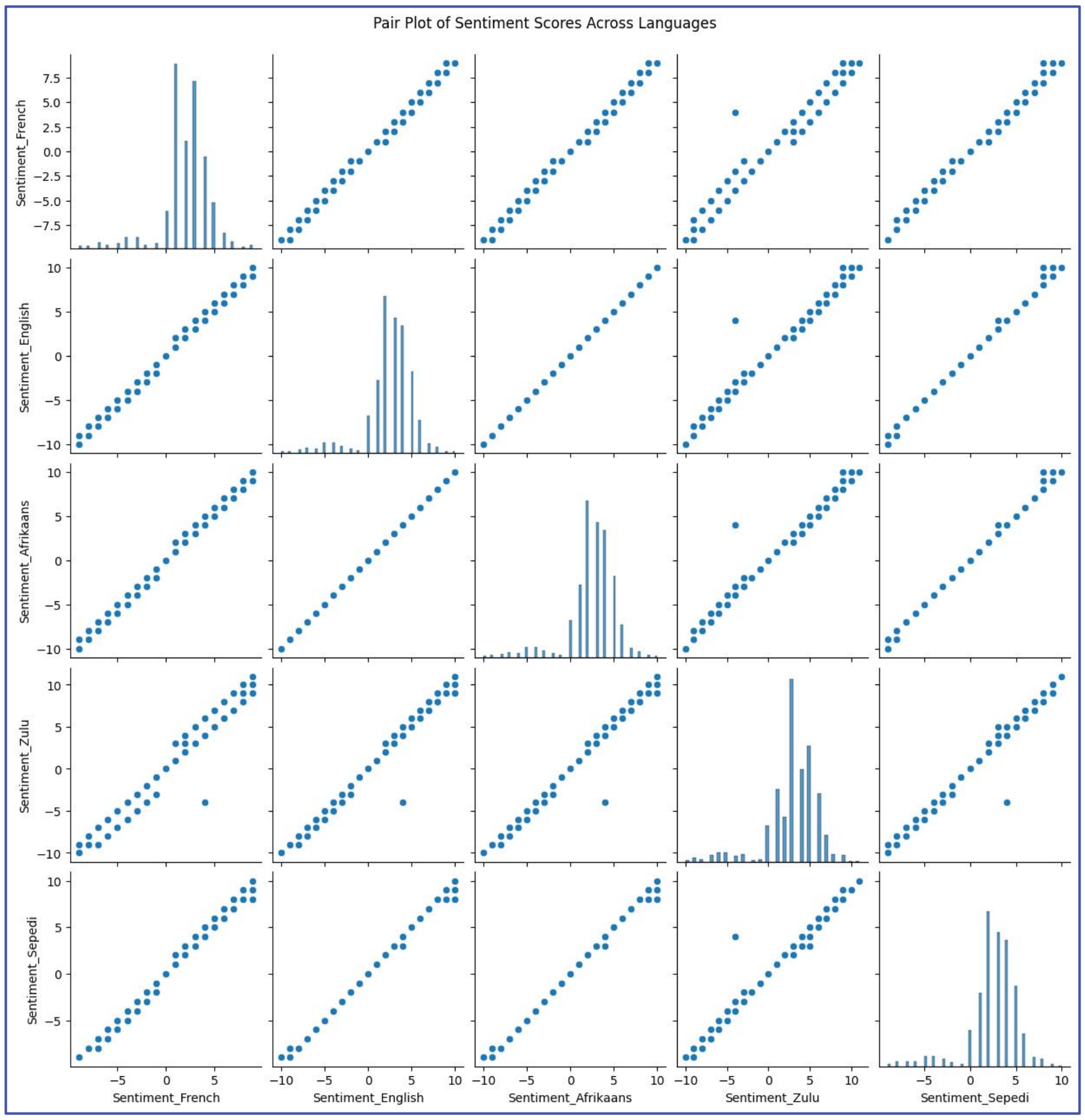

*Figure 9 Sentiment Score Against Languages.*

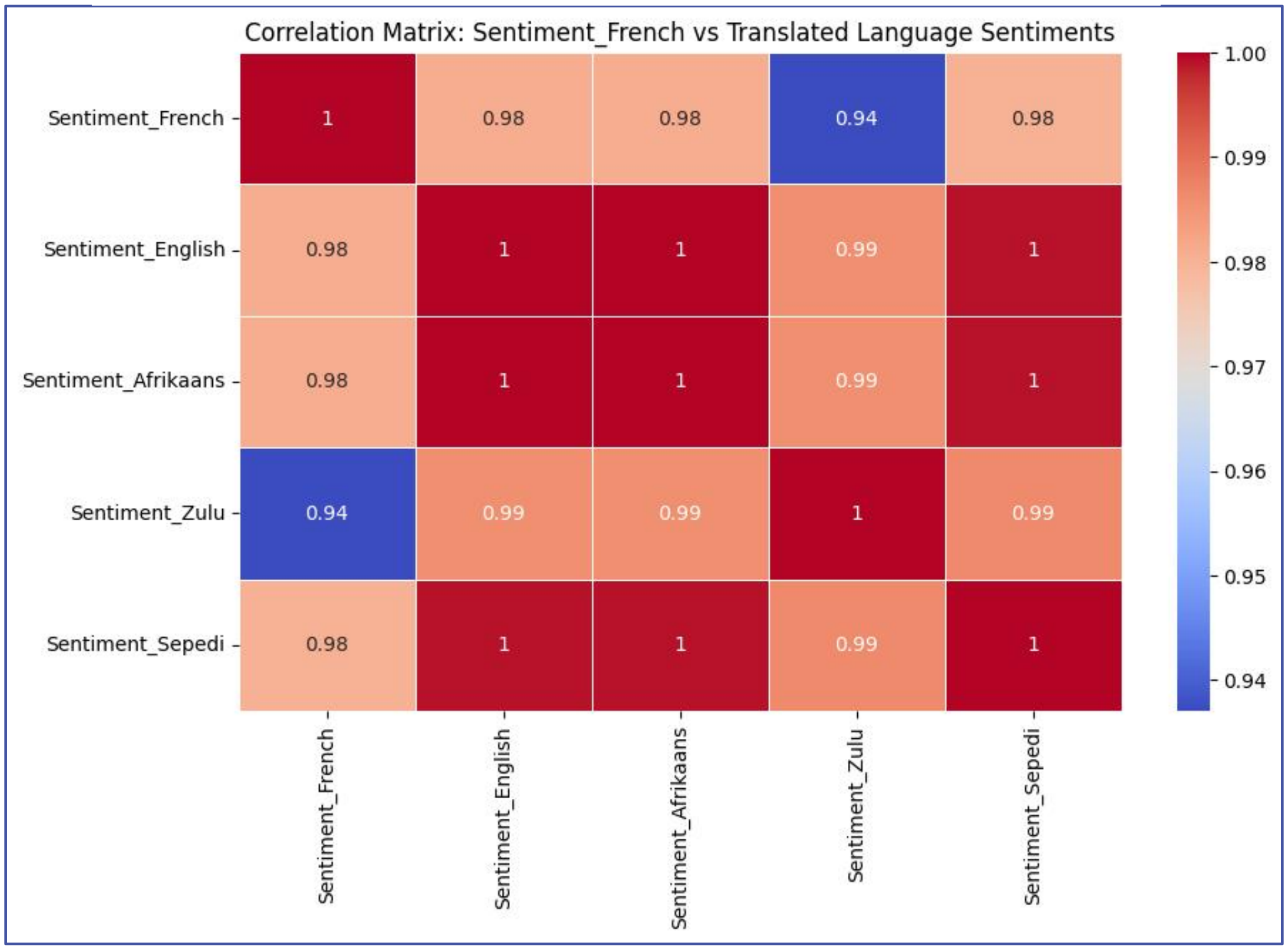

*Figure 10 French Sentiment vs Translated Language Sentiments.*

The confusion matrix shows high correlations across translations in French, English, Afrikaans, and Sepedi, with values ranging from 0.98 to 1 (Figure 10). This indicates effective translation of sentiments across these languages. The slightly lower correlation between French and Zulu suggests minor differences in sentiment interpretation or expression between these two languages. However, overall, the sentiment scores are well-preserved across translations (Figure 11).

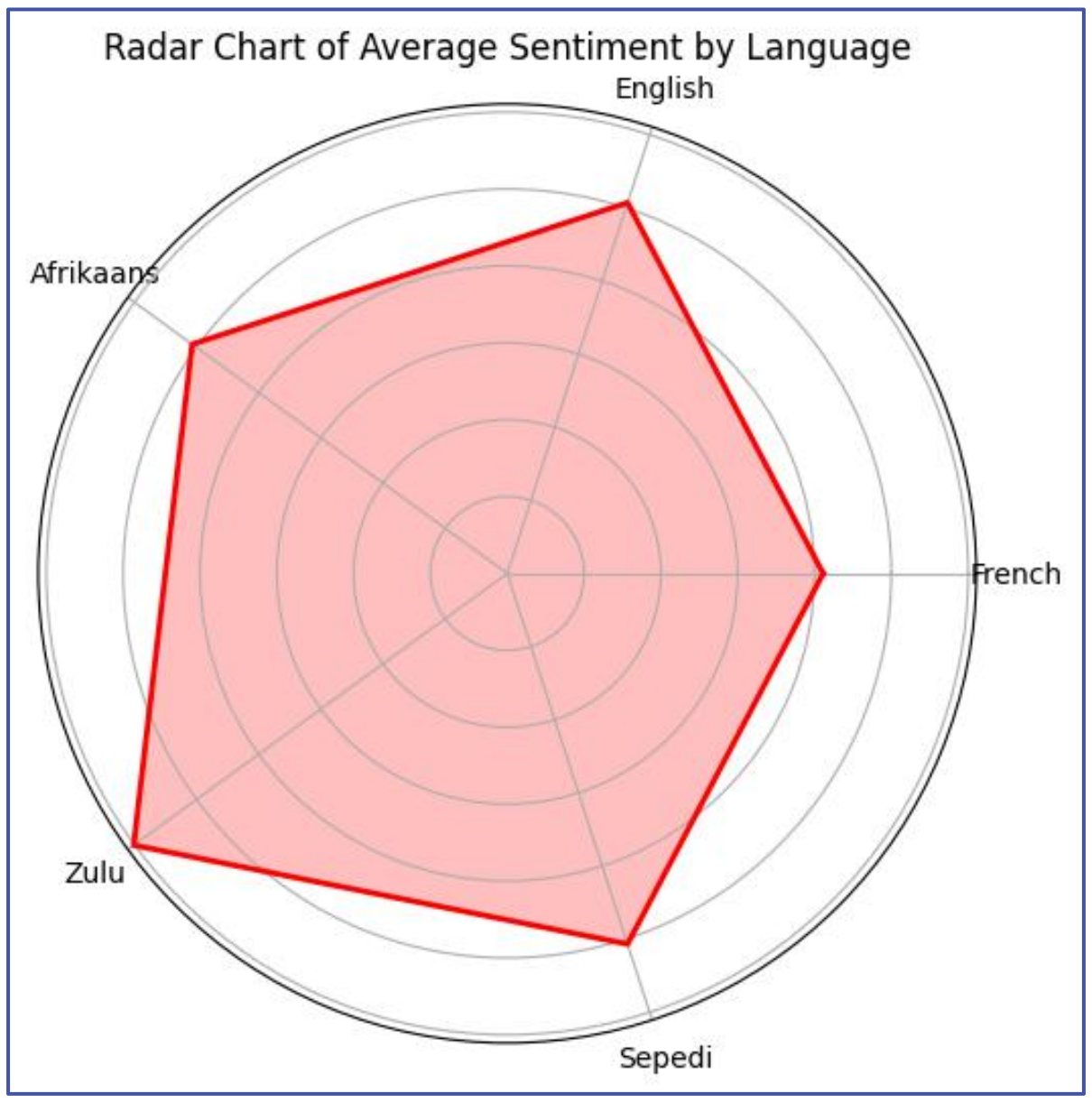

*Figure 11 Chart Showing the Distribution of Sentiment.*

The radar chart is mainly evenly distributed amongst the different languages, which further confirms that sentiment translation is relatively consistent across these languages, with only minor variations (figures 11-13).

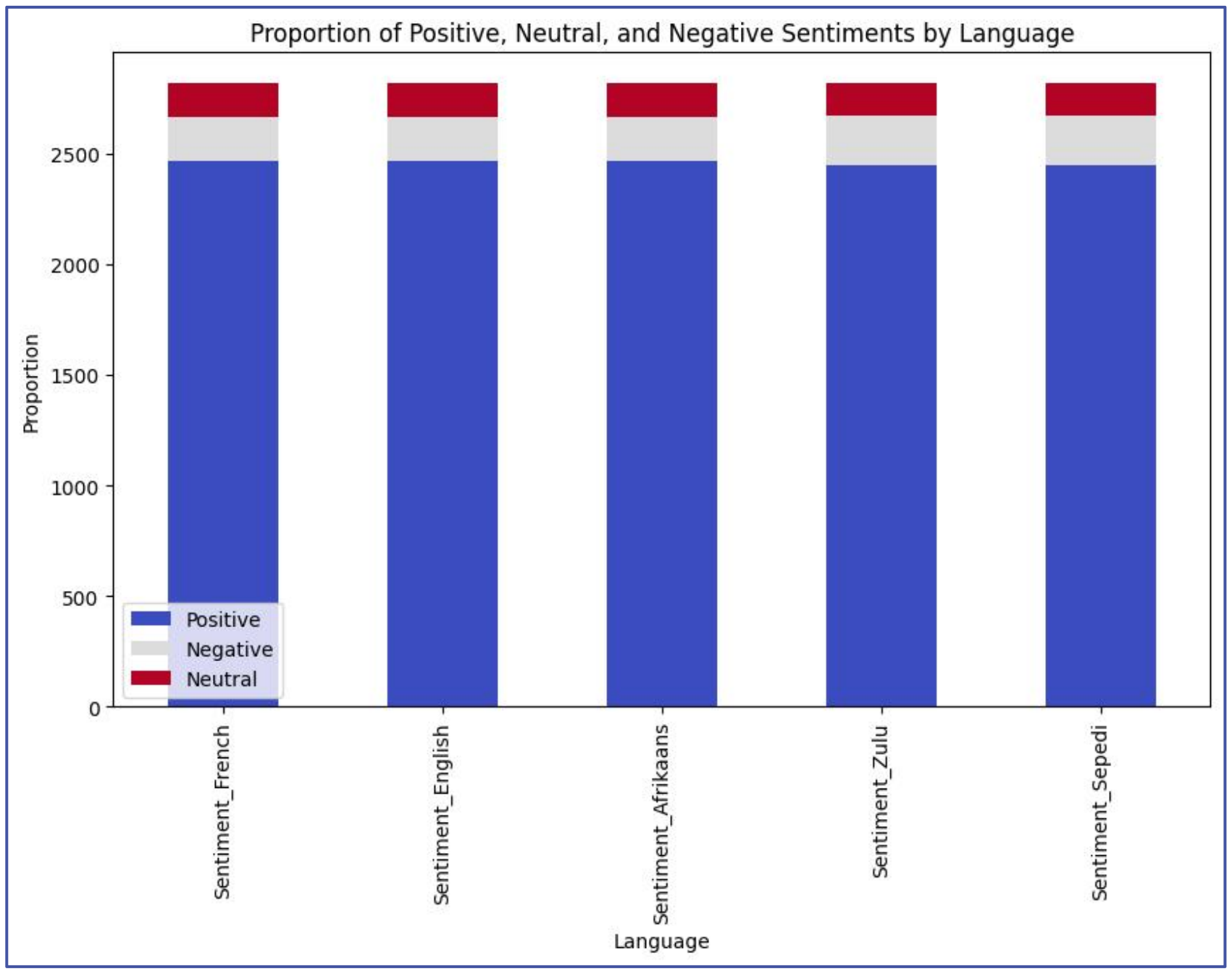

*Figure 12 Proportions of Sentiment by Language.*

Figures 13-14 show the distribution of the sentiment scores across the different languages, in the Lexicon. The sentiment scores represent the positive value or negative value associated with the word. A positive sentiment score indicates a positive sentiment, while a negative score indicates a negative sentiment. The density plot shows the sentiment score for each language. The data is skewed slightly to the left, indicating that each language tends to words that have a neutral to positive sentiment score. The distribution across the languages is relatively similar, differing slightly in the density and spread of the sentiment scores.

The French language has the highest peaks, indicating a higher density of positive words between the 0 to 5 sentiment score. The Zulu language has a higher density of the higher positive words, between a sentiment score of 5-7. The lexicon shows a considerably lower density of words with a negative sentiment score, across all languages.

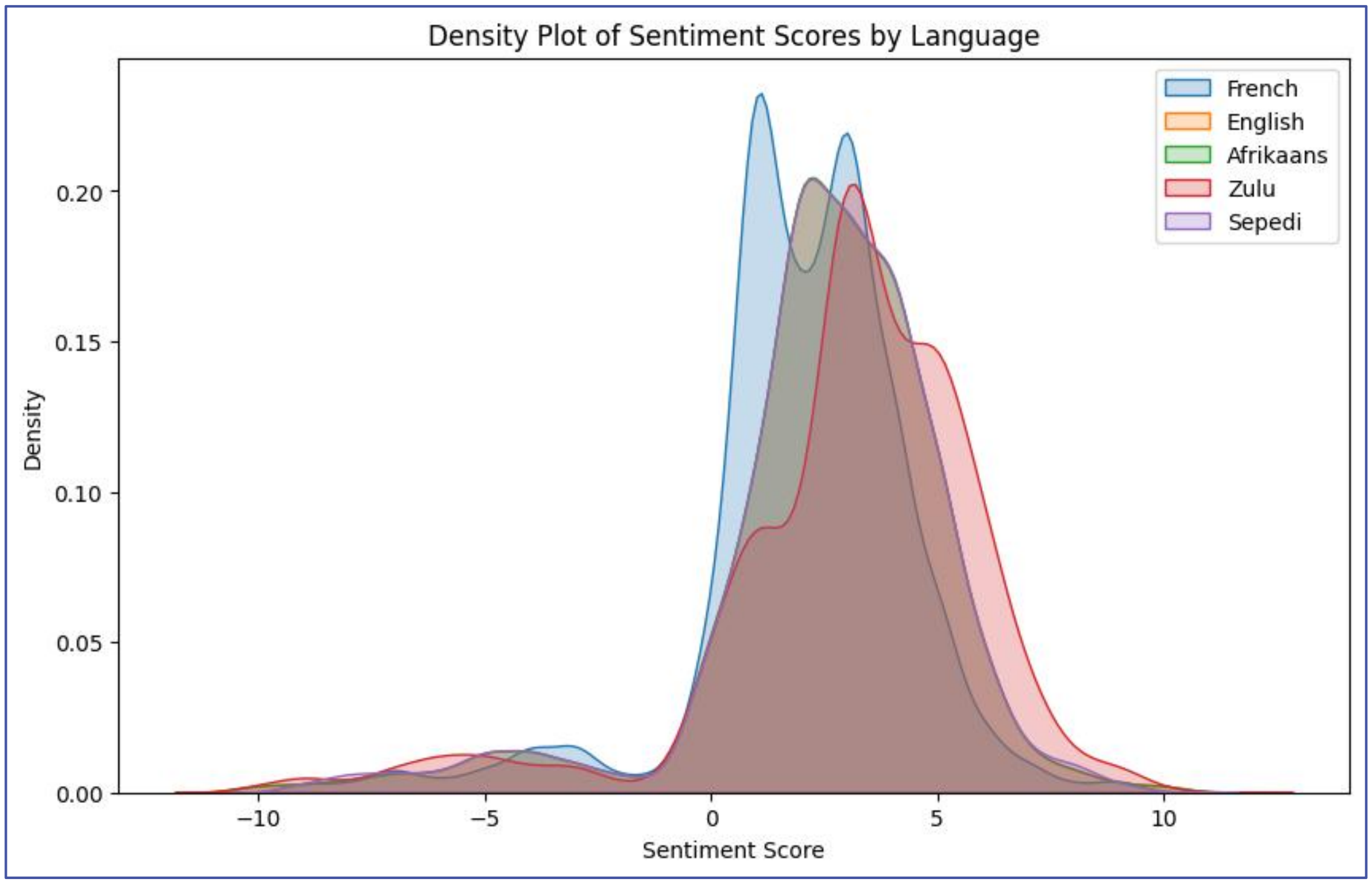

*Figure 13 Sentiments by Language.*

# 4.2 Sentiment Analysis

### 4.2.1 Source Language: Afrikaans

| | sentence | source_language | target_language | translated_text |
|---|---|---|---|---|
| 0 | Ek vertrou haar | afrikaans | zulu | mina thembeka ake |
| 1 | Ek vertrou haar | afrikaans | sepedi | ke tshepo yena |
| 2 | Ek vertrou haar | afrikaans | english | i trust her |
| 3 | Bestuur na die stad. | afrikaans | english | drive to the city |

### Table 2: Sentiment Analysis Results

| | total_score_avg | word_scores_avg | sentiment_avg | total_score_v2 | word_scores_v2 | sentiment_v2 | vader_compound | vader_sentiment |
|---|---|---|---|---|---|---|---|---|
| 0 | 2.600000 | ek:0; vertrou:0; haar:2.6 | positive | 2.600000 | ek:0; vertrou:0; haar:2.6 | positive | 0.0000 | neutral |
| 1 | 2.600000 | ek:0; vertrou:0; haar:2.6 | positive | 2.600000 | ek:0; vertrou:0; haar:2.6 | positive | 0.0000 | neutral |
| 2 | 2.600000 | ek:0; vertrou:0; haar:2.6 | positive | 2.600000 | ek:0; vertrou:0; haar:2.6 | positive | 0.0000 | neutral |
| 3 | 4.666667 | bestuur:0; na:0; die:0.0; stad:4.666666666666667 | positive | 4.666667 | bestuur:0; na:0; die:0; stad:4.666666666666667 | positive | -0.5994 | negative |

*Figure 14 Sentiment Performed by V2 Sentiment Analysis.*

Figure 14 shows translated texts with positive sentiment performed by the V2 advanced sentiment analysis. In comparison to this, the Vader sentiment analysis shows neutral and negative sentiment for the above sentences. The Vader sentiment is a built in tool to perform sentiment analysis, but it is not correctly identifying sentiments which could be because of the languages used in the expanded lexicon, therefore, the V2 advanced sentiment analysis seems to be doing well in translating and identifying sentiments for the Afrikaans language (Figure 15).

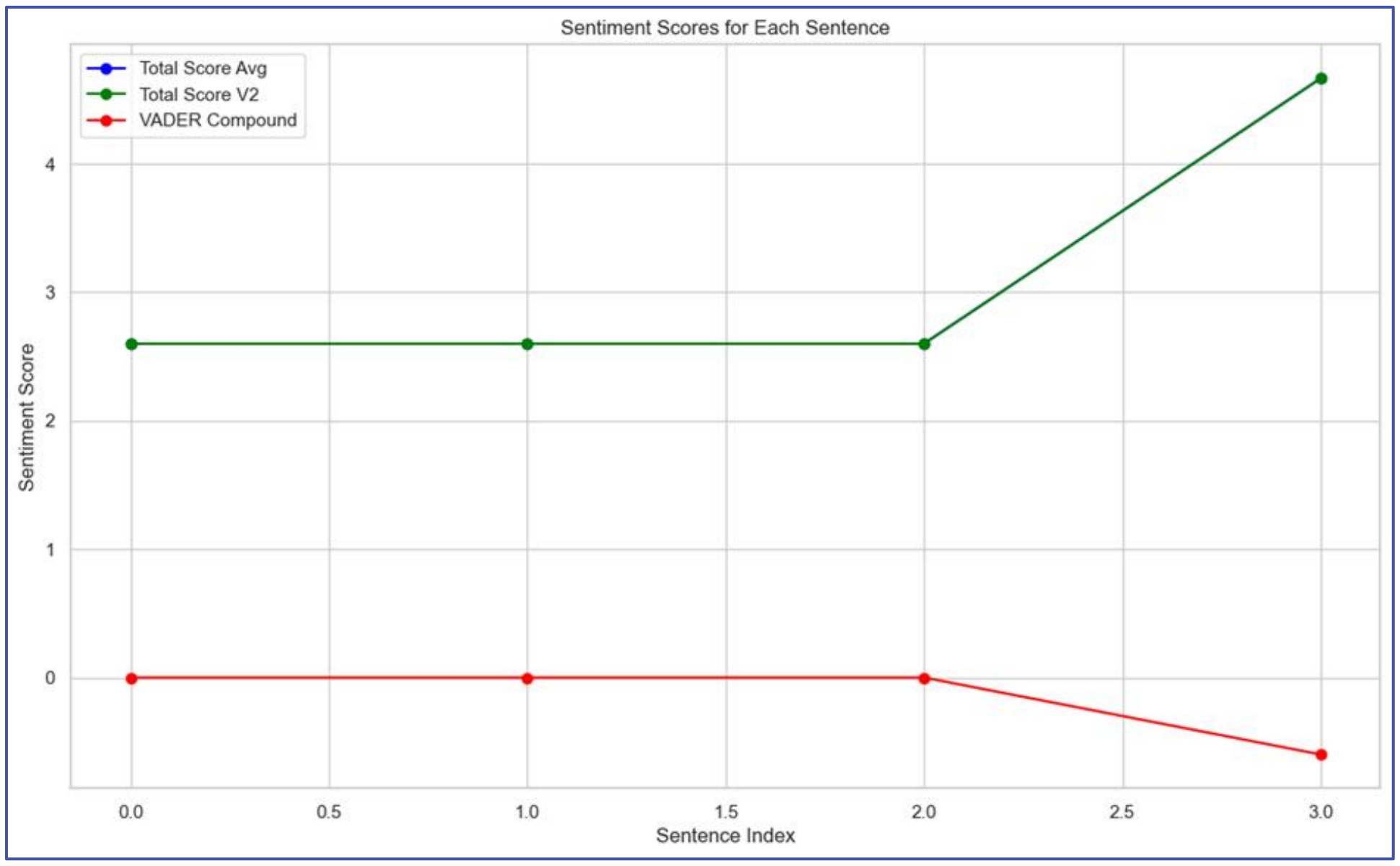

*Figure 15 Sentiment Score for Each Sentence, for Vader and V2.*

Figure 15 depicts a difference in the sentiment analysis performed by the Vader compound and the V2 advanced sentiment analysis. The Vader compound shows a negative sentiment, the V2 advanced sentiment analysis shows a positive sentiment (Figure 16).

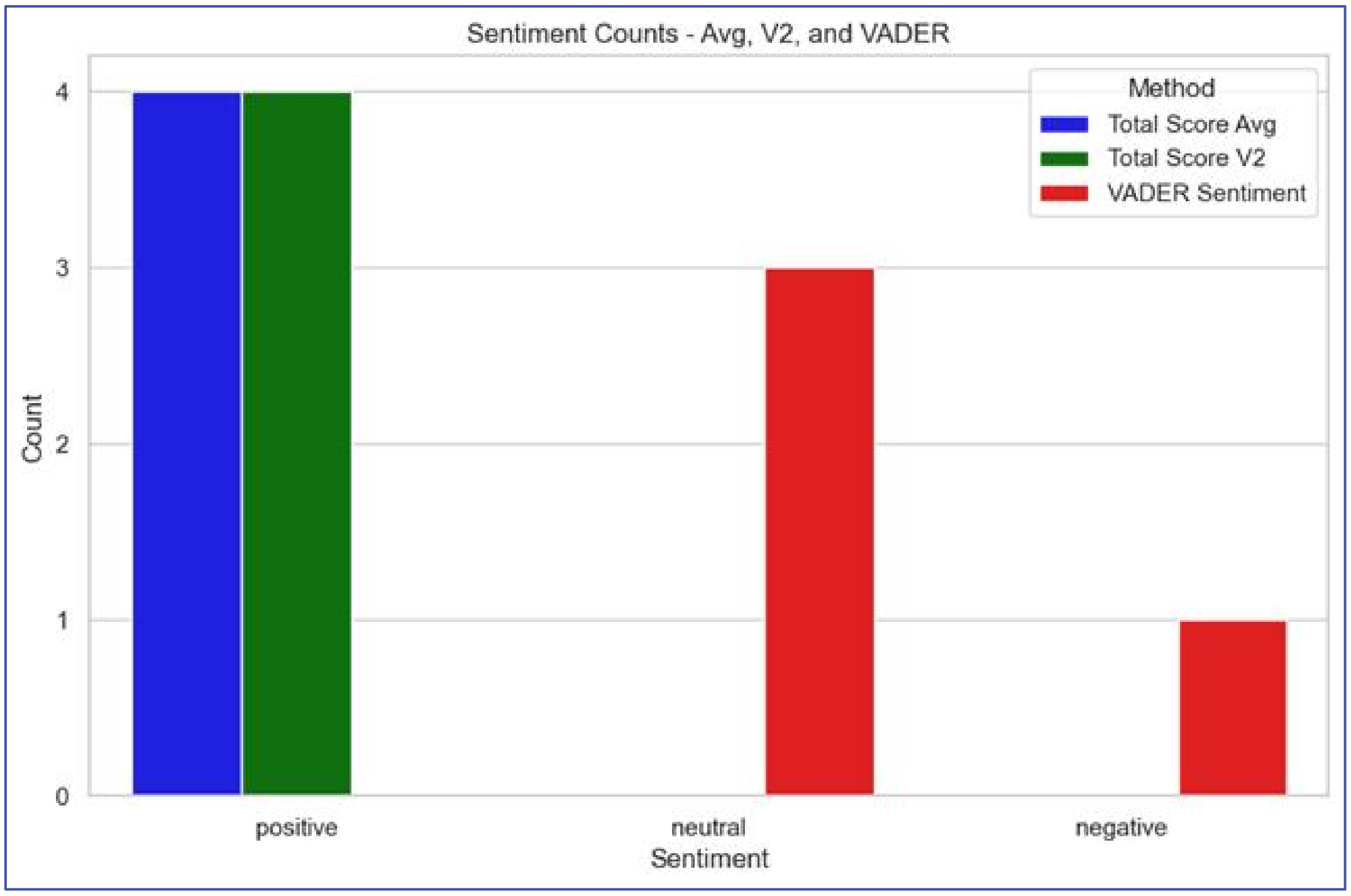

*Figure 16 Sentiment for V2 and Vader.*

The total average score of sentiment is positive, which is correctly identified by the V2 advanced sentiment analysis, while the Vader sentiment seems to struggle in identifying the sentiments as it classified it as neutral and negative. The V2 advanced sentiment analysis is able to correctly identify sentiments translated from the Afrikaans language (Figure 17).

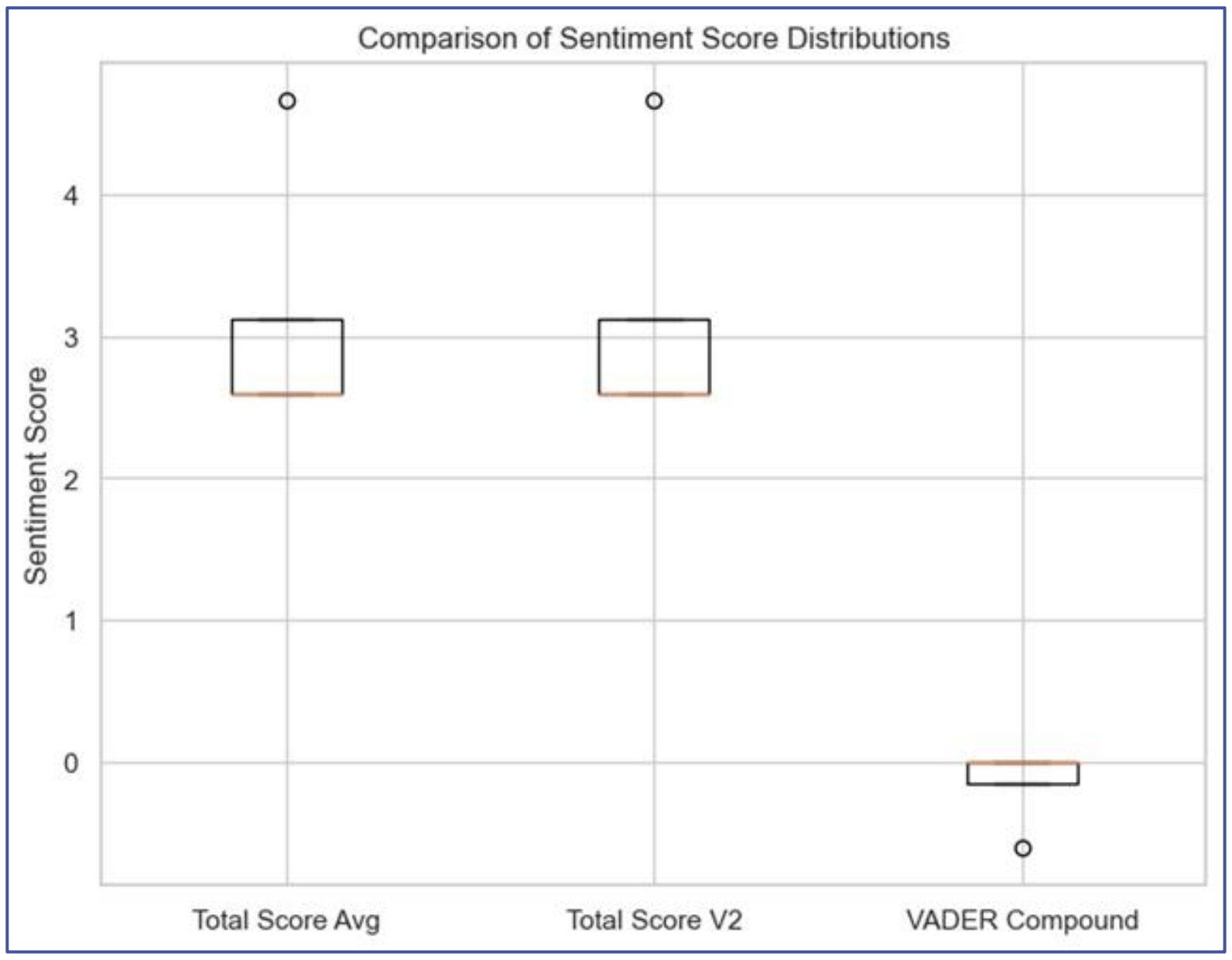

*Figure 17 Comparison of Sentiment Score Distributions.*

The V2 advanced sentiment analysis correctly identified the sentiments in the sentences given, while the Vader compound struggled to correctly identify the sentiment (figures 18-19).

### 4.2.2 Source Language: English

### Table 1: Translations

| | sentence | source_language | target_language | translated_text |
|---|---|---|---|---|
| 0 | I want food. | english | zulu | mina funa ukudla |
| 1 | I want to dance with you. | english | zulu | mina funa ukudansa nabo wena |
| 2 | Come to uncle Josh. | english | zulu | fika uku umalume josh |
| 3 | I am happy today. | english | zulu | nginguye jabulile namhlanje |
| 4 | What do you want to watch? | english | zulu | yini yenza wena funa uku buka |
| 5 | Waiting to dance to accompany you. | english | zulu | lindizayo ukudansa ukuhambisana wena |
| 6 | Have you eaten? | english | zulu | hamba wena udleke |
| 7 | Thank you. | english | sepedi | re a leboga |
| 8 | I want food. | english | sepedi | ke o nyaka dijo |
| 9 | Follow me home for fun. | english | sepedi | latelela nna gae bakeng sa monate |
| 10 | Come to uncle Josh. | english | sepedi | etla go fihla go: malome josh |
| 11 | I am happy today. | english | sepedi | ke nna thabile lehono |
| 12 | Waiting to dance to accompany you. | english | sepedi | letile go bina go felegetša wena |
| 13 | Thank you. | english | french | merci |
| 14 | Have a good day | english | french | avoir une bonnejournée |
| 15 | I want food. | english | french | je vouloir aliment |
| 16 | I want to dance with you. | english | french | je vouloir danser avec vous |
| 17 | Come to uncle Josh. | english | french | venir à oncle josh |
| 18 | Waiting to dance to accompany you. | english | french | attente danser accompagner vous |
| 19 | Thank you. | english | ciluba | tuasakadila |

*Figure 18 English Sentences Translated to Different Target Languages.*

### Table 2: Sentiment Analysis Results

| | total_score_avg | word_scores_avg | sentiment_avg | total_score_v2 | word_scores_v2 | sentiment_v2 | vader_compound | vader_sentiment |
|---|---|---|---|---|---|---|---|---|
| 0 | 6.500000 | i:0.0; want:3.5; food:3.0 | positive | 7.000000 | i:0; want:4; food:3 | positive | 0.0772 | positive |
| 1 | 5.900000 | i:0.0; want:3.5; to dance:2.4; with:0.0; you:0.0 | positive | 6.400000 | i:0; want:4; to dance:2.4; with:0; you:0 | positive | 0.0772 | positive |
| 2 | 7.666667 | come:3.3333333333333335; to:0.0; uncle:4.33333... | positive | 7.666667 | come:3.3333333333333335; to:0; uncle:4.3333333... | positive | 0.0000 | neutral |
| 3 | 9.166667 | i am:3.0; happy:4.5; today:1.6666666666666667 | positive | 10.500000 | i am:3; happy:5; today:2.5 | positive | 0.5719 | positive |
| 4 | 6.750000 | what:0.0; do:1.5; you:0.0; want:3.5; to:0.0; w... | positive | 7.750000 | what:0; do:2.0; you:0; want:4; to:0; watch:1.75 | positive | 0.0772 | positive |
| 5 | 8.900000 | waiting:2.5; to dance:2.4; to accompany:4.0; y... | positive | 9.400000 | waiting:3; to dance:2.4; to accompany:4; you:0 | positive | 0.0000 | neutral |
| 6 | 6.500000 | have:3.0; you:0.0; eaten:3.5 | positive | 9.000000 | have:3; you:0; eaten:6 | positive | 0.0000 | neutral |
| 7 | 5.000000 | thank you:5.0 | positive | 5.000000 | thank you:5 | positive | 0.3612 | positive |
| 8 | 6.500000 | i:0.0; want:3.5; food:3.0 | positive | 7.000000 | i:0; want:4; food:3 | positive | 0.0772 | positive |
| 9 | 9.000000 | follow:2.0; me:0.0; home:5.0; for:1.0; fun:1.0 | positive | 10.000000 | follow:2; me:0; home:6; for:1; fun:1 | positive | 0.5106 | positive |
| 10 | 7.666667 | come:3.3333333333333335; to:0.0; uncle:4.33333... | positive | 7.666667 | come:3.3333333333333335; to:0; uncle:4.3333333... | positive | 0.0000 | neutral |
| 11 | 9.166667 | i am:3.0; happy:4.5; today:1.6666666666666667 | positive | 10.500000 | i am:3; happy:5; today:2.5 | positive | 0.5719 | positive |
| 12 | 8.900000 | waiting:2.5; to dance:2.4; to accompany:4.0; y... | positive | 9.400000 | waiting:3; to dance:2.4; to accompany:4; you:0 | positive | 0.0000 | neutral |
| 13 | 5.000000 | thank you:5.0 | positive | 5.000000 | thank you:5 | positive | 0.3612 | positive |
| 14 | 7.000000 | have:3.0; a:0.0; good day:4.0 | positive | 7.000000 | have:3; a:0; good day:4 | positive | 0.4404 | positive |
| 15 | 6.500000 | i:0.0; want:3.5; food:3.0 | positive | 7.000000 | i:0; want:4; food:3 | positive | 0.0772 | positive |
| 16 | 5.900000 | i:0.0; want:3.5; to dance:2.4; with:0.0; you:0.0 | positive | 6.400000 | i:0; want:4; to dance:2.4; with:0; you:0 | positive | 0.0772 | positive |
| 17 | 7.666667 | come:3.3333333333333335; to:0.0; uncle:4.33333... | positive | 7.666667 | come:3.3333333333333335; to:0; uncle:4.3333333... | positive | 0.0000 | neutral |
| 18 | 8.900000 | waiting:2.5; to dance:2.4; to accompany:4.0; y... | positive | 9.400000 | waiting:3; to dance:2.4; to accompany:4; you:0 | positive | 0.0000 | neutral |

*Figure 19 Sentiment Analysis Results.*

The sentences have positive sentiments according to the sentiment average values, the V2 advanced sentiment analysis is able to correctly identify the sentiment analysis for the given sentences , while the Vader sentiment struggles to correctly identify the sentiments in the given texts (Figure 19). There are a number of reasons why the Vader sentiment is not performing well , it is due to the nature of the sentences and the lack of punctuation. Vader sentiment struggles in correctly identifying sentiments behind these sentences (Figure 20).

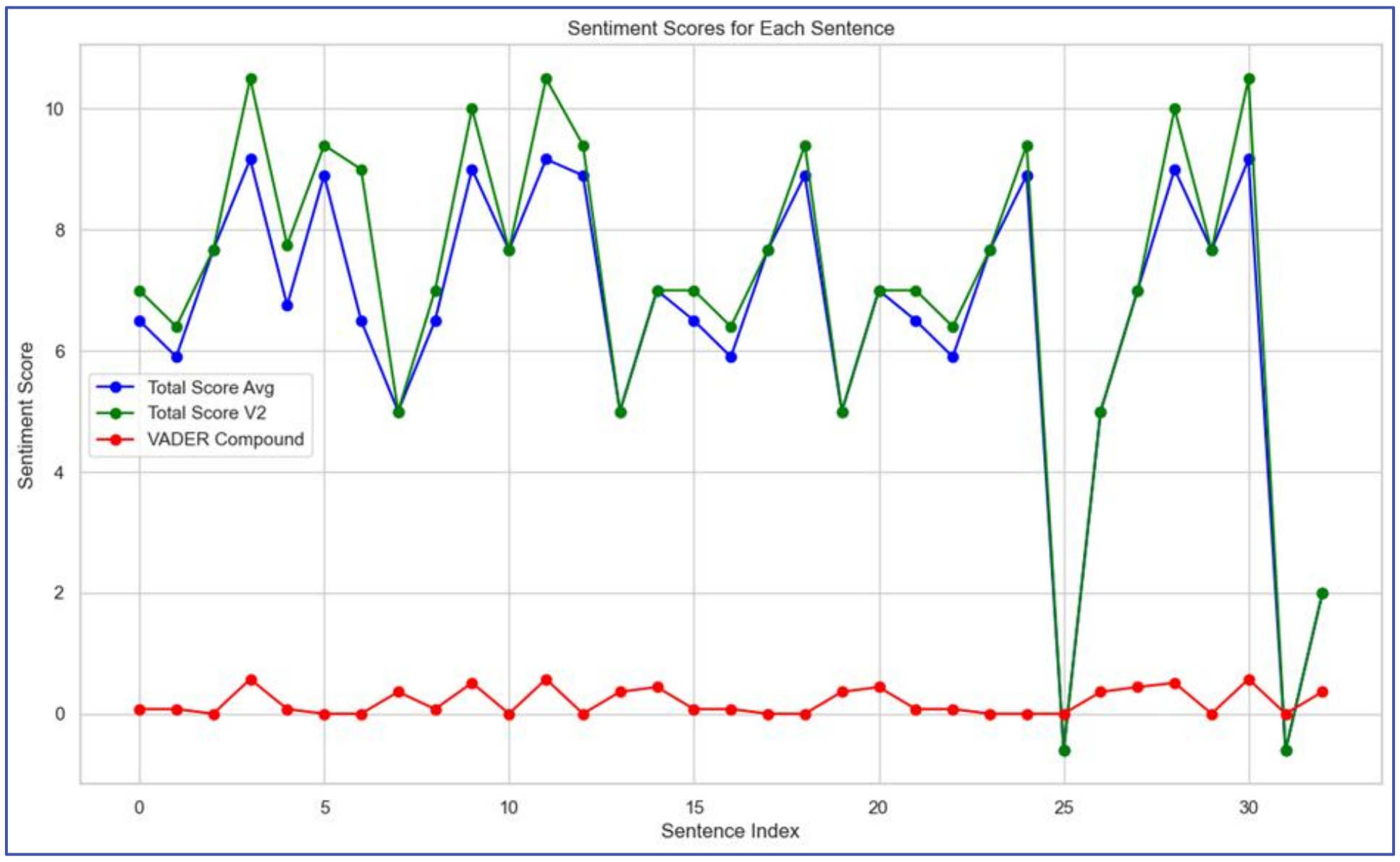

*Figure 20 Sentiment Scores for Each Sentence.*

The V2 advanced sentiment analysis is able to correctly identify the sentiments as either positive or negative, however, it is not extremely accurate in identifying the sentiment score , but performs well overall, while the Vader compound struggles to correctly identify the sentiment and its scores. The Vader compound performs poorly in identifying the sentiment and their scores (Figure 21).

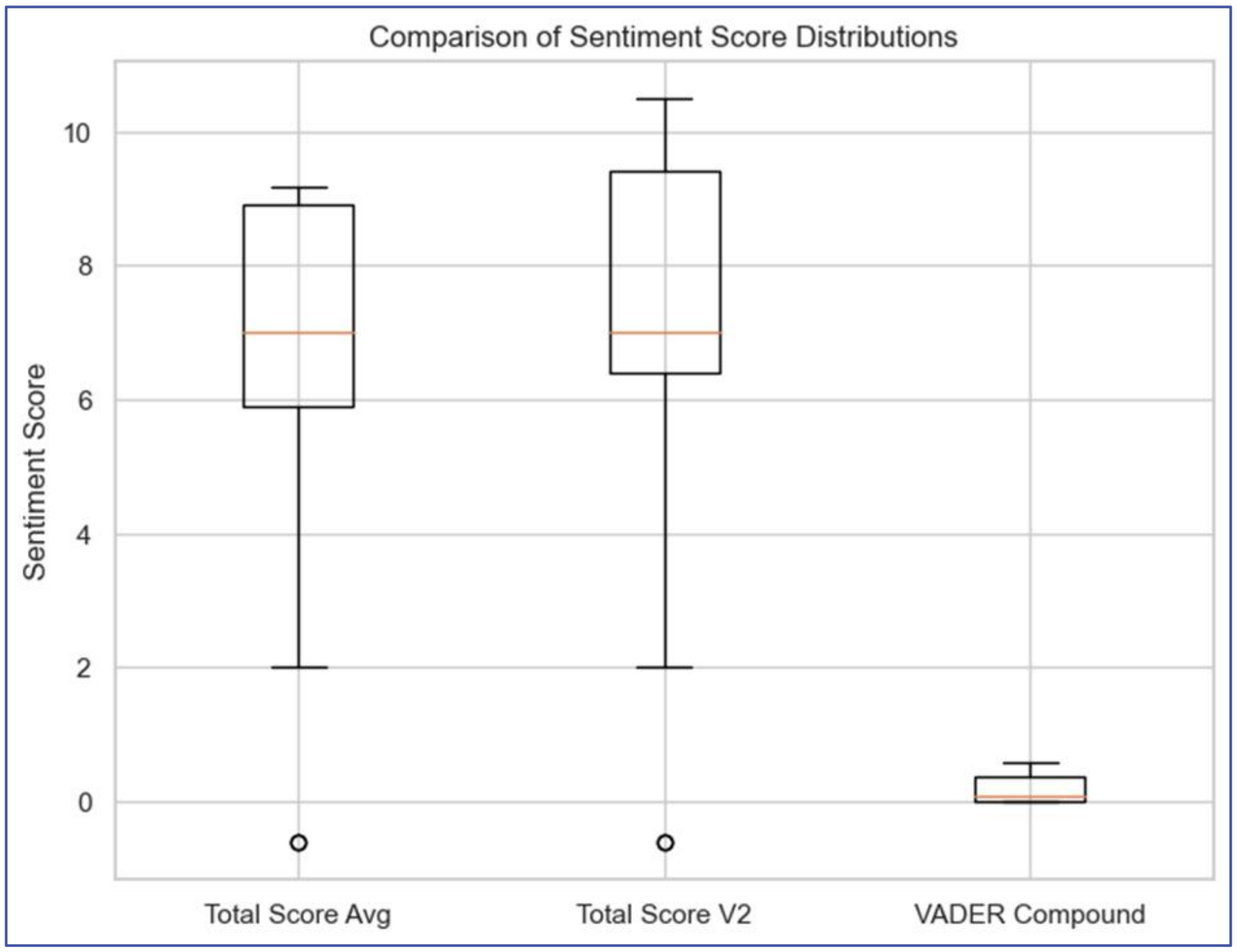

*Figure 21 Comparison of Sentiment Score Distributions.*

The distribution of sentiments is poorly distributed for the Vader compound due to its poor performance. Vader is identifying sentiments, while the distribution of V2 advanced sentiment analysis is more spread out compared to the total score average. The V2 advanced sentiment analysis is able to correctly identify the overall sentiment but struggled with the specific scores which results in the spread-out distribution observed in figures 21 and 22.

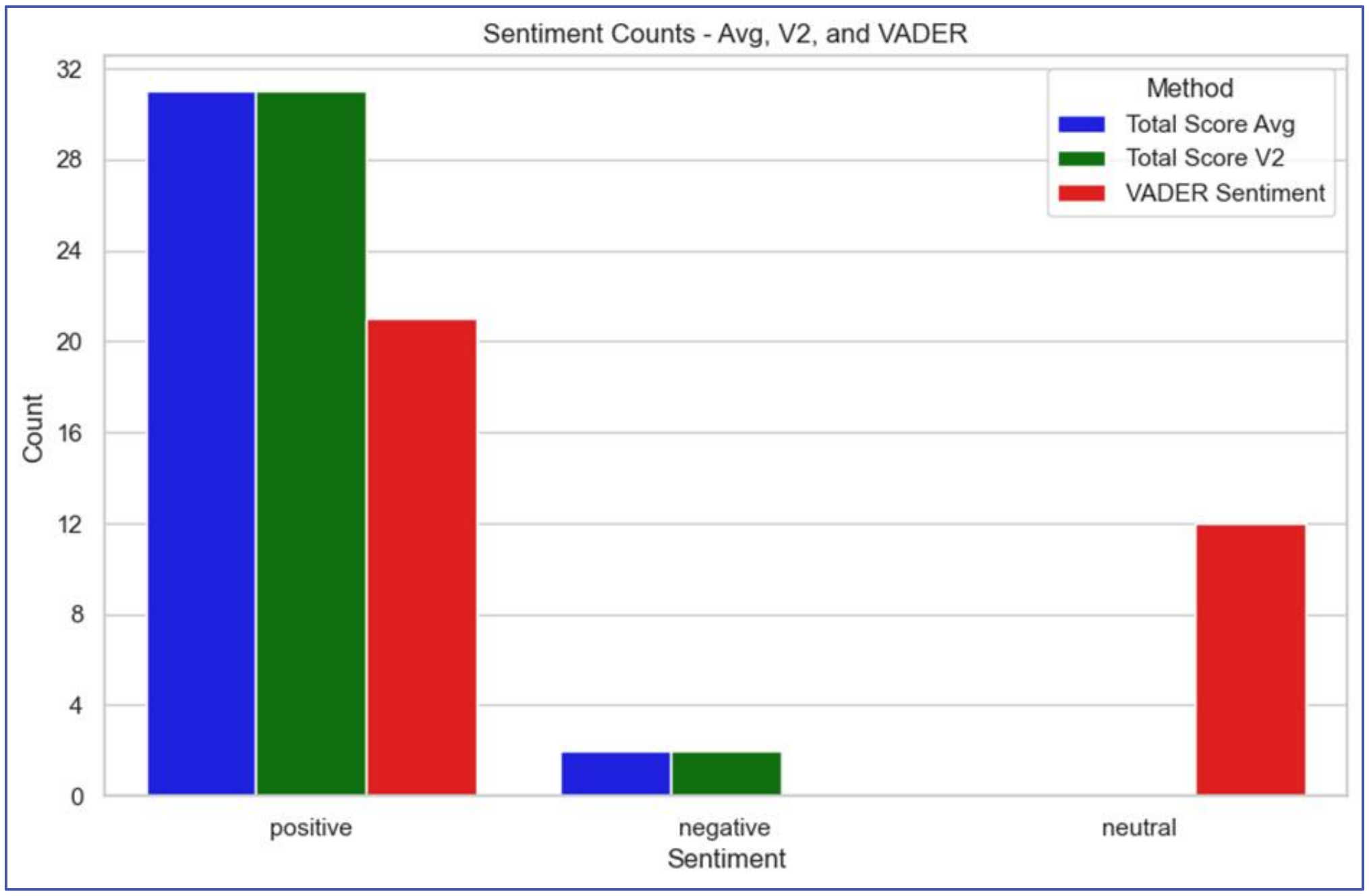

*Figure 22 Sentiment Counts.*

The V2 advanced sentiment analysis was able to correctly identify the sentiments in the sentences according to the total score average, while the Vader sentiment struggled to correctly identify the sentiments and misclassified some as neutral that should have been positive (Figure 22). The V2 advanced sentiment analysis is able to correctly identify the sentiments as either positive or negative, while the Vader sentiment analysis struggles by identifying the sentiments as neutral (figures 23-24).

### 4.2.3 Source Language: Sepedi

### Table 1: Translations

| | sentence | source_language | target_language | translated_text |
|---|---|---|---|---|
| 0 | Go tšhaba go wa. | sepedi | zulu | ukwesaba ukuwa |
| 1 | Go letela ngwedi. | sepedi | zulu | ukulindela inyanga |
| 2 | Ke motho le go tšhaba kotlo. | sepedi | zulu | mina umuntu futhi ukwesaba inhlawulo |
| 3 | O rata go letela tše mpe. | sepedi | zulu | uthanda ukulindela ezimbi |
| 4 | Go tšhaba go wa. | sepedi | zulu | ukwesaba ukuwa |
| 5 | Go tšhaba go wa. | sepedi | english | to fear to fall |
| 6 | Go letela ngwedi. | sepedi | english | to expect moon |
| 7 | O rata go letela tše mpe. | sepedi | english | you like to expect bad things |
| 8 | Go tšhaba go wa. | sepedi | afrikaans | tot vrees om te val |

### Table 2: Sentiment Analysis Results

| | total_score_avg | word_scores_avg | sentiment_avg | total_score_v2 | word_scores_v2 | sentiment_v2 | vader_compound | vader_sentiment |
|---|---|---|---|---|---|---|---|---|
| 0 | -1.000000 | go tšhaba:0; go wa:-1.0 | negative | -1.00 | go tšhaba:0; go wa:-1 | negative | 0.0 | neutral |
| 1 | 3.250000 | go letela:1.0; ngwedi:2.25 | positive | 3.25 | go letela:1; ngwedi:2.25 | positive | 0.0 | neutral |
| 2 | 2.166667 | ke:0.0; motho:1.0; le:0.0; go tšhaba:0; kotlo:... | positive | 4.00 | ke:0; motho:1; le:0; go tšhaba:0; kotlo:3.0 | positive | 0.0 | neutral |
| 3 | 1.000000 | o rata:9.0; go letela:1.0; tše mpe:-9.0 | positive | 1.00 | o rata:9; go letela:1; tše mpe:-9 | positive | 0.0 | neutral |
| 4 | -1.000000 | go tšhaba:0; go wa:-1.0 | negative | -1.00 | go tšhaba:0; go wa:-1 | negative | 0.0 | neutral |
| 5 | -1.000000 | go tšhaba:0; go wa:-1.0 | negative | -1.00 | go tšhaba:0; go wa:-1 | negative | 0.0 | neutral |
| 6 | 3.250000 | go letela:1.0; ngwedi:2.25 | positive | 3.25 | go letela:1; ngwedi:2.25 | positive | 0.0 | neutral |
| 7 | 1.000000 | o rata:9.0; go letela:1.0; tše mpe:-9.0 | positive | 1.00 | o rata:9; go letela:1; tše mpe:-9 | positive | 0.0 | neutral |
| 8 | -1.000000 | go tšhaba:0; go wa:-1.0 | negative | -1.00 | go tšhaba:0; go wa:-1 | negative | 0.0 | neutral |

*Figure 23 Translation and Sentiment Analysis Results.*

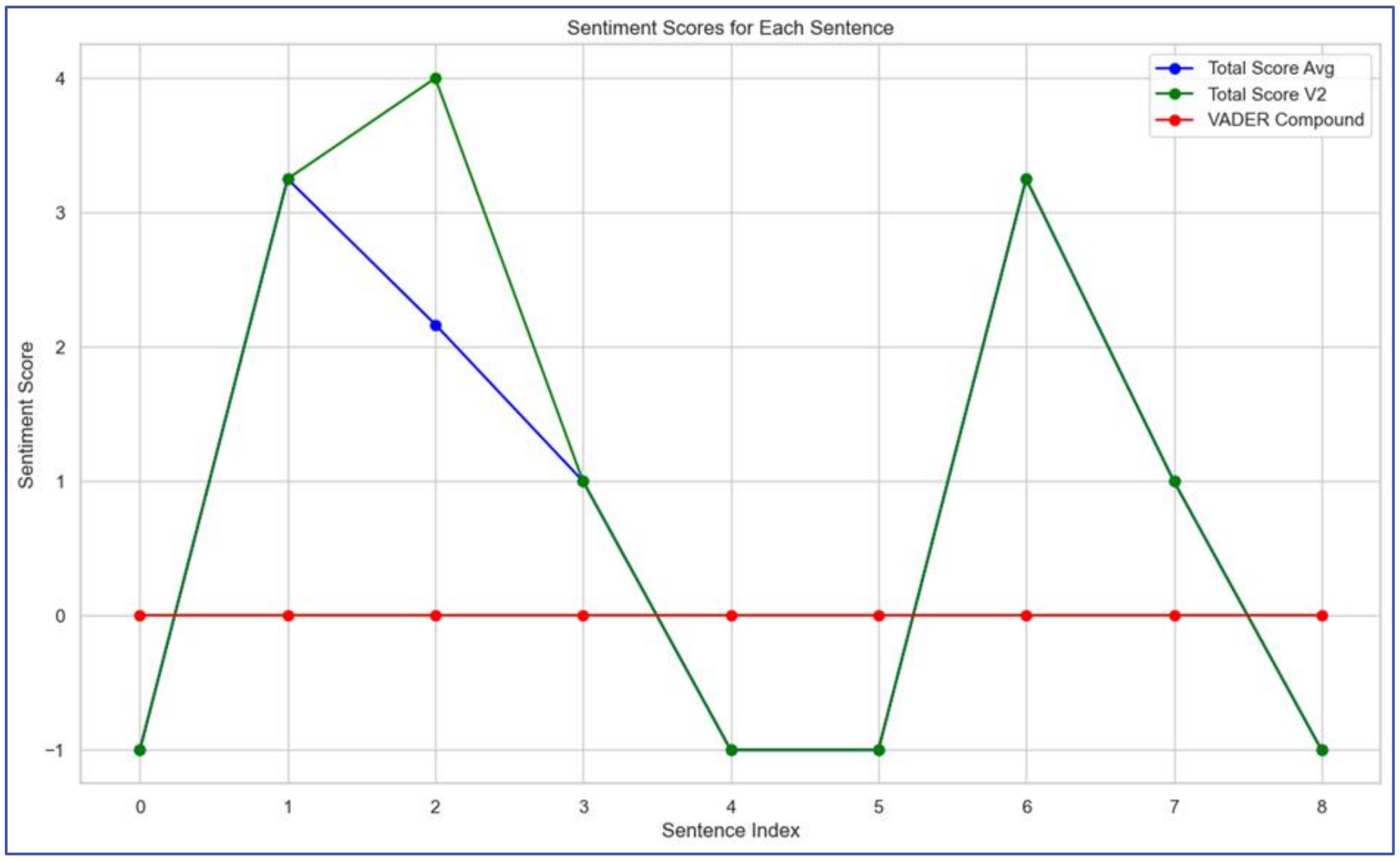

*Figure 24 Sentiment Score for Each Sentence.*

The Vader compound was not able to identify any of the sentiments as all the sentiments are classified as neutral , while the V2 sentiment struggles in some sentences, but shows a good performance of sentiment analysis overall compared to the total score average.

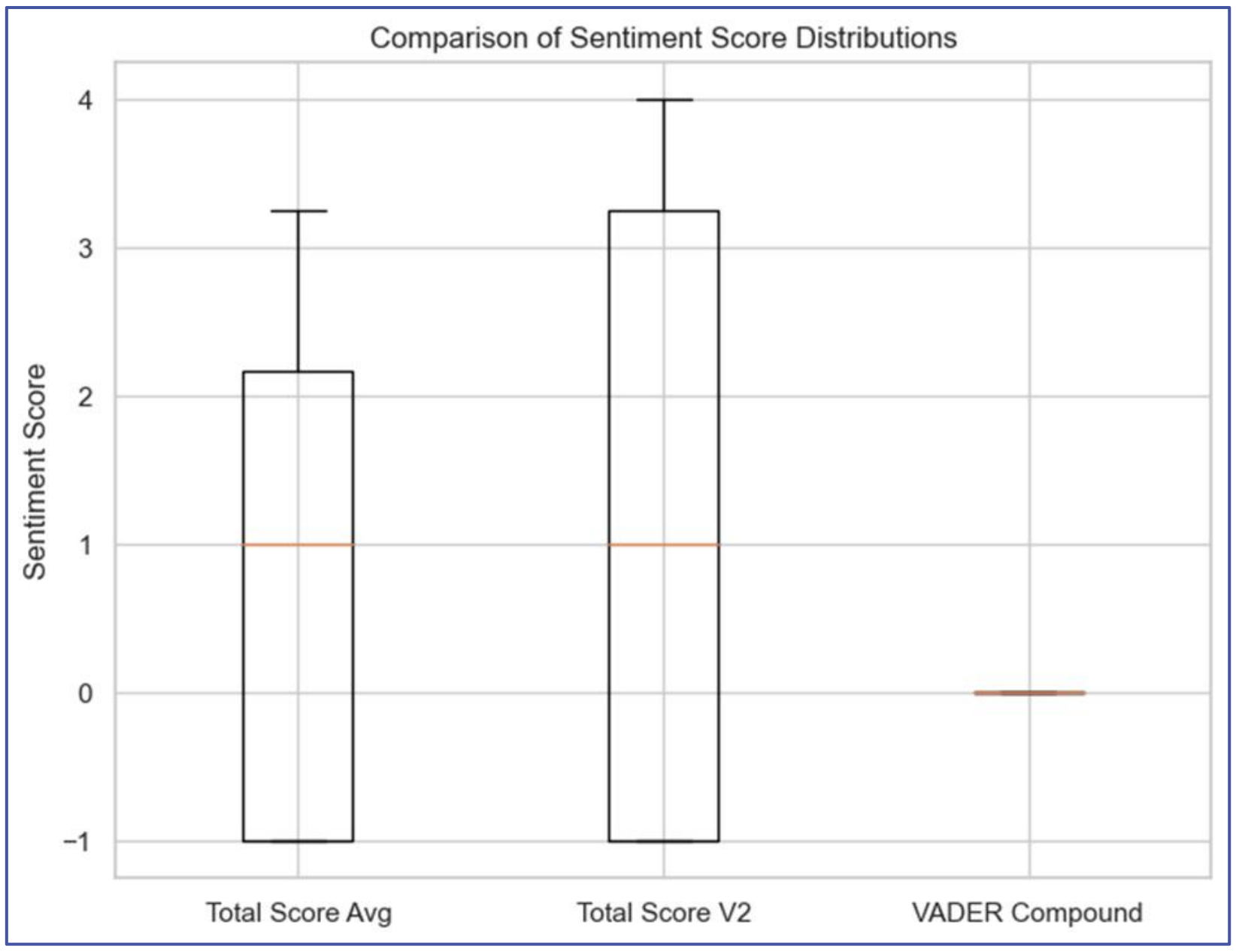

*Figure 25 Sentiment Score Distributions.*

There is no distribution of sentiment analysis for the Vader compound, as it classified all the sentiments as neutral, while the V2 sentiment is more distributed compared to the total score average, due to some misclassifications, however, the mean is the same in the V2 sentiment analysis as well as the total score average.

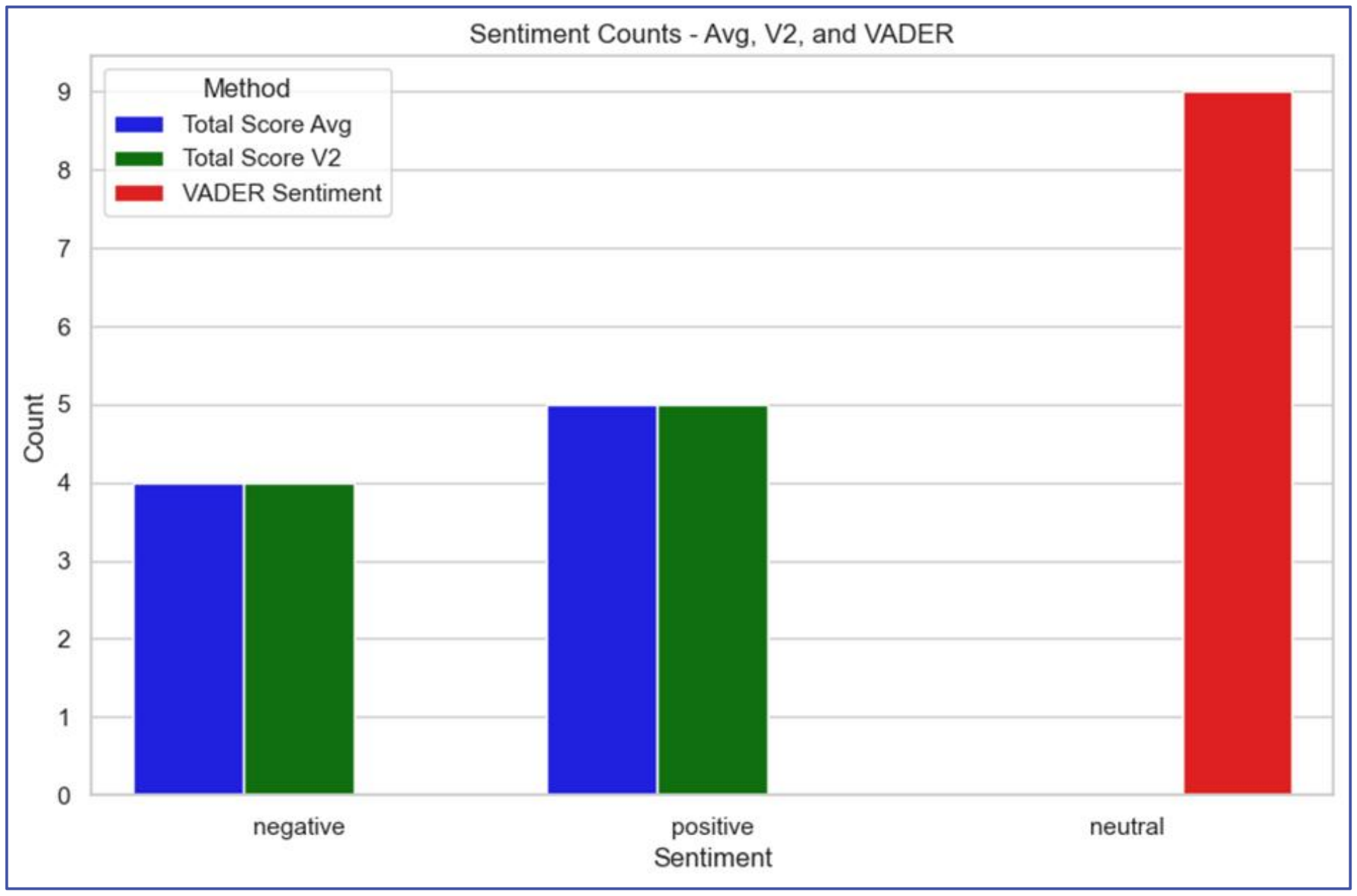

*Figure 26 Sentiment Counts.*

The V2 was able to correctly classify the sentiments as either positive or negative, while the Vader sentiment classified all the sentiments as neutral (figures 27-28).

### 4.2.4 Source Language: Zulu

### Table 1: Translations

| | sentence | source_language | target_language | translated_text |
|---|---|---|---|---|
| 0 | Ngiyabathanda abantu abazinakekelayo | zulu | english | i love them women those who are caring |
| 1 | kubalulekile ukuzinakekela | zulu | english | it's important to take care of oneself |
| 2 | yenza okufunayo | zulu | english | make what you want |
| 3 | uphatha kabi abantu | zulu | english | you treat badly women |
| 4 | uthanda izinto ezimbi | zulu | english | you like things bad things |
| 5 | Ngiyabathanda abantu abazinakekelayo | zulu | sepedi | ke a ba rata basadi bao ba kgathalago |
| 6 | Mase ubona engathi umuthanda kakhulu muyeke | zulu | sepedi | ge o bona bjalo ka o a mo rata ka bontši mo lese |
| 7 | yenza okufunayo | zulu | sepedi | dira se o nyakago |
| 8 | uphatha kabi abantu | zulu | sepedi | o swara gampe basadi |
| 9 | uthanda izinto ezimbi | zulu | sepedi | o rata dilo tše mpe |
| 10 | kubalulekile ukuzinakekela | zulu | afrikaans | dit is belangrik om jouself te versorg |
| 11 | yenza okufunayo | zulu | afrikaans | maak wat jy wil hê |

*Figure 27 Translations.*

### Table 2: Sentiment Analysis Results

| | total_score_avg | word_scores_avg | sentiment_avg | total_score_v2 | word_scores_v2 | sentiment_v2 | vader_compound | vader_sentiment |
|---|---|---|---|---|---|---|---|---|
| 0 | 11.75 | ngiyabathanda:0; abantu:2.75; abazinakekelayo:9.0 | positive | 11.75 | ngiyabathanda:0; abantu:2.75; abazinakekelayo:9 | positive | 0.0 | neutral |
| 1 | 5.00 | kubalulekile:0.0; ukuzinakekela:5.0 | positive | 5.00 | kubalulekile:0; ukuzinakekela:5 | positive | 0.0 | neutral |
| 2 | 2.50 | yenza:2.5; okufunayo:0.0 | positive | 3.00 | yenza:3; okufunayo:0 | positive | 0.0 | neutral |
| 3 | 2.75 | uphatha:5.0; kabi:-5.0; abantu:2.75 | positive | 2.75 | uphatha:5; kabi:-5; abantu:2.75 | positive | 0.0 | neutral |
| 4 | 12.00 | uthanda:9.0; izinto:3.0; ezimbi:0 | positive | 12.00 | uthanda:9; izinto:3; ezimbi:0 | positive | 0.0 | neutral |
| 5 | 11.75 | ngiyabathanda:0; abantu:2.75; abazinakekelayo:9.0 | positive | 11.75 | ngiyabathanda:0; abantu:2.75; abazinakekelayo:9 | positive | 0.0 | neutral |
| 6 | -2.00 | mase:0; ubona:0.0; engathi:0.0; umuthanda:-3.0... | negative | -2.00 | mase:0; ubona:0; engathi:0; umuthanda:-3; kakh... | negative | 0.0 | neutral |
| 7 | 2.50 | yenza:2.5; okufunayo:0.0 | positive | 3.00 | yenza:3; okufunayo:0 | positive | 0.0 | neutral |
| 8 | 2.75 | uphatha:5.0; kabi:-5.0; abantu:2.75 | positive | 2.75 | uphatha:5; kabi:-5; abantu:2.75 | positive | 0.0 | neutral |
| 9 | 12.00 | uthanda:9.0; izinto:3.0; ezimbi:0 | positive | 12.00 | uthanda:9; izinto:3; ezimbi:0 | positive | 0.0 | neutral |
| 10 | 5.00 | kubalulekile:0.0; ukuzinakekela:5.0 | positive | 5.00 | kubalulekile:0; ukuzinakekela:5 | positive | 0.0 | neutral |
| 11 | 2.50 | yenza:2.5; okufunayo:0.0 | positive | 3.00 | yenza:3; okufunayo:0 | positive | 0.0 | neutral |

*Figure 28 Sentiment Analysis Results.*

The Vader sentiment once again classifies the sentiments as neutral, while the V2 advanced sentiment analysis correctly classified the sentiments in the sentences translated from the Zulu language to the target languages (figures 27-28). The V2 sentiment analysis correctly identifies the sentiments as either positive or negative but has a few sentences that do not match the total score average, although correctly identified as positive or negative (figures 29-30). The distribution of sentiment scores is almost identical in the V2 sentiment analysis and the total score average, while there is no distribution for the Vader compound as the sentences are classified as neutral . The Vader compound struggles to identify the sentiments from languages such as Zulu and Sepedi and classifies the sentiments as neutral (figures 30-31). The Vader sentiment classifies the sentiments as neutral, while the V2 advanced sentiment analysis correctly classified the sentiments in the sentences to the target languages to either positive or negative.

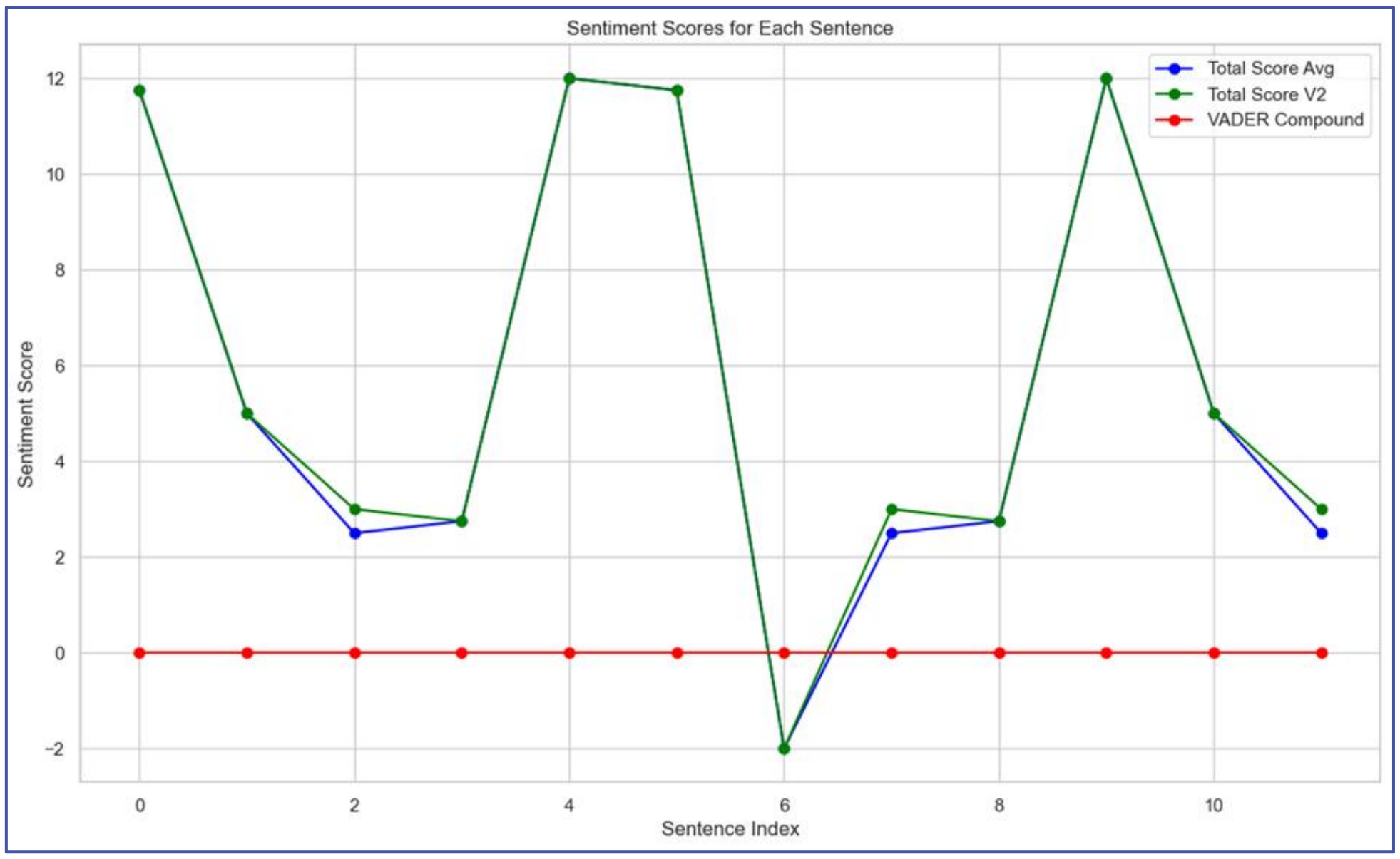

*Figure 29 Sentiment Score for Each Sentence.*

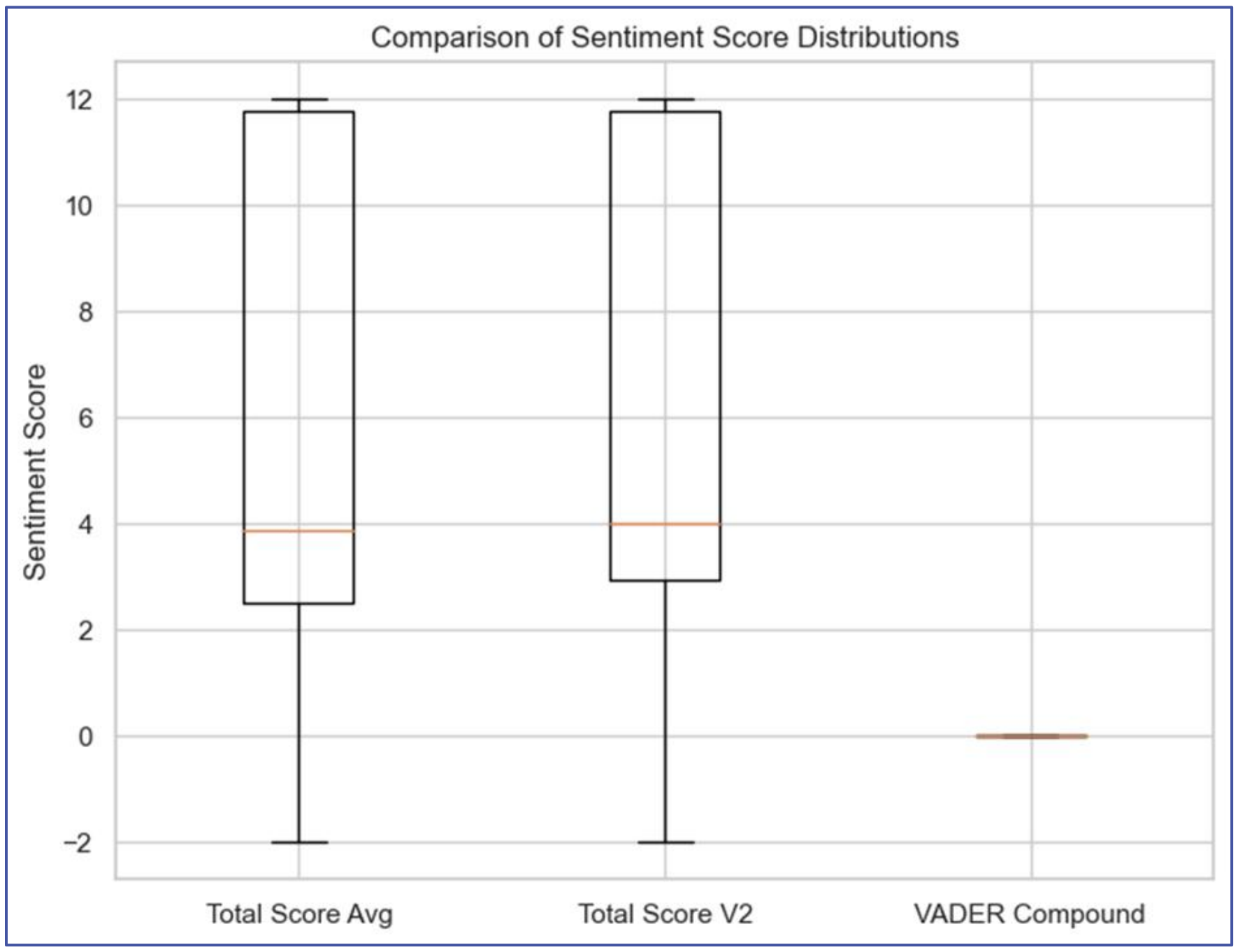

*Figure 30 Distribution of Sentiment Scores.*

The Random Forest model managed to achieve an accuracy of 66% on our Lexicon (Figure 31). The confusion metrics shows that classes like 'Mot' and 'nombre' are predicted well on the model with a recall and precision value of 0.96 and 0.68 (Figure 31). This indicates that these classes are identified correctly on the model. The precision, however, varies significantly across classes, with many classes for example, "adjectif," "adverb," "pronompersonnel", having zero precision, indicating that the model did not successfully predict these classes. "Mot" and "nombre" have relatively higher precision scores, suggesting that when the model predicted these classes, it was often correct. The F1-score, which balances precision and recall, is low for most classes, with a weighted average of 0.56. The model struggles particularly with classes like "verbe," which has an F1-score of only 0.12, indicating poor performance in both precision and recall for this class (Table 1). This could mean that the model might need more data or features to improve prediction for underperforming classes.

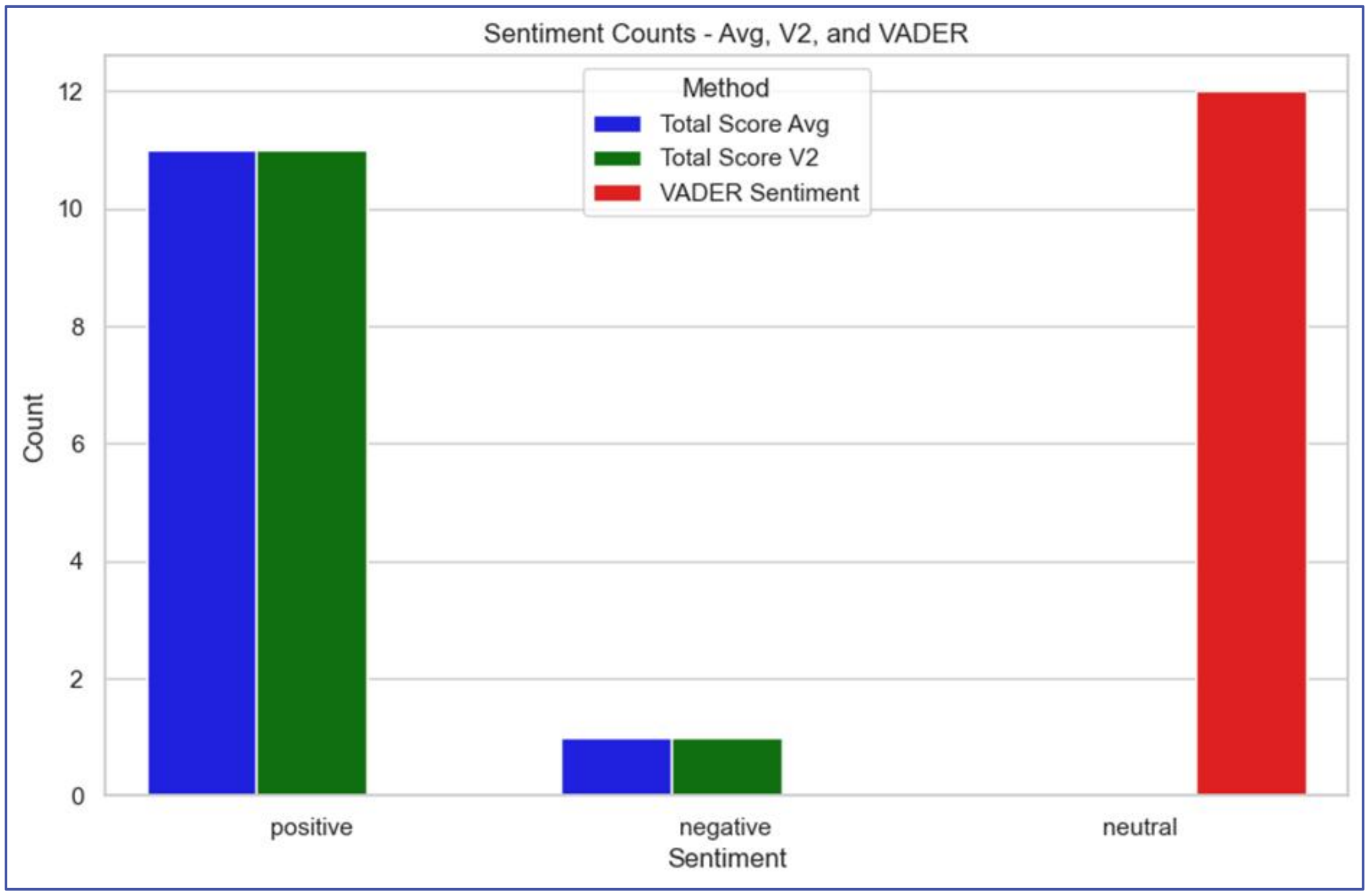

*Figure 31 Sentiment Counts.*

## 4.3 Machine Learning

### 4.3.1 Random Forest

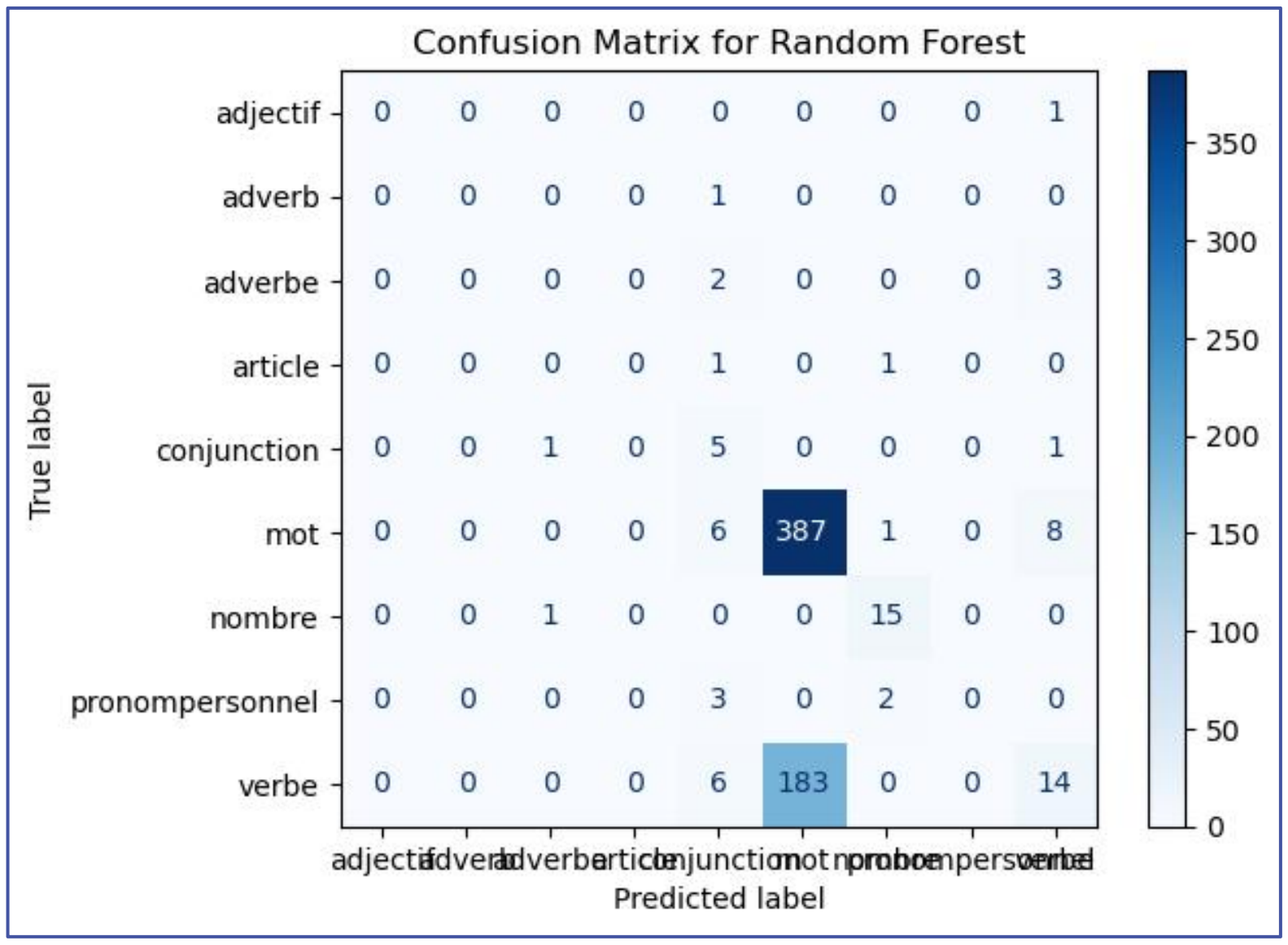

*Figure 33 Confusion Matrix for RF.*

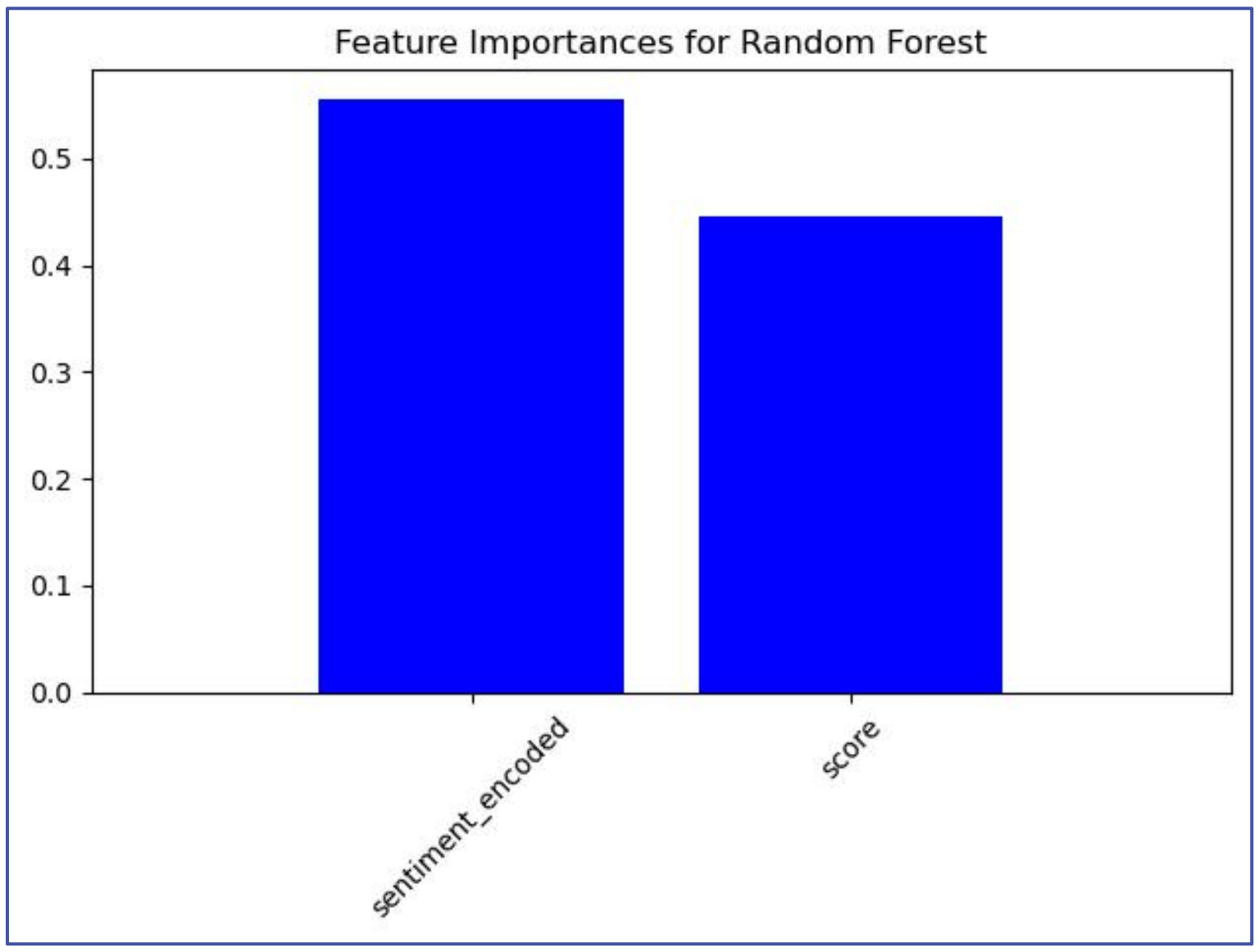

*Figure 32 Feature Importance for RF.*

The use of receiver operating characteristics (ROC) curves further explains the variation between the nature of the words as individual classes affecting the discrimination thresholds for the model. Overall, Random Forest classes aggregated had an area under curve (AUC) of 0.96 indicating an impressive performance across all the classes (Figure 34-35). While the prediction focus is on the sentiment, understanding the impact of the nature of the words is equally important.

| | Precision | Recall | F1-score | Support |
|---|---|---|---|---|
| adjectif | 0.0 | 0.0 | 0.0 | 1 |
| adverb | 0.0 | 0.0 | 0.0 | 1 |
| adverbe | 0.0 | 0.0 | 0.0 | 5 |
| article | 0.0 | 0.0 | 0.0 | 2 |
| conjunction | 0.21 | 0.71 | 0.32 | 7 |
| mot | 0.68 | 0.96 | 0.8 | 402 |
| nombre | 0.79 | 0.94 | 0.86 | 16 |
| pronompersonnel | 0.0 | 0.0 | 0.0 | 5 |
| verbe | 0.52 | 0.07 | 0.12 | 203 |
| accuracy | | | 0.66 | 642 |
| macro avg | 0.24 | 0.3 | 0.23 | 642 |
| weighted avg | 0.61 | 0.66 | 0.56 | 642 |

*Table 1 Performance Metrics for RF.*

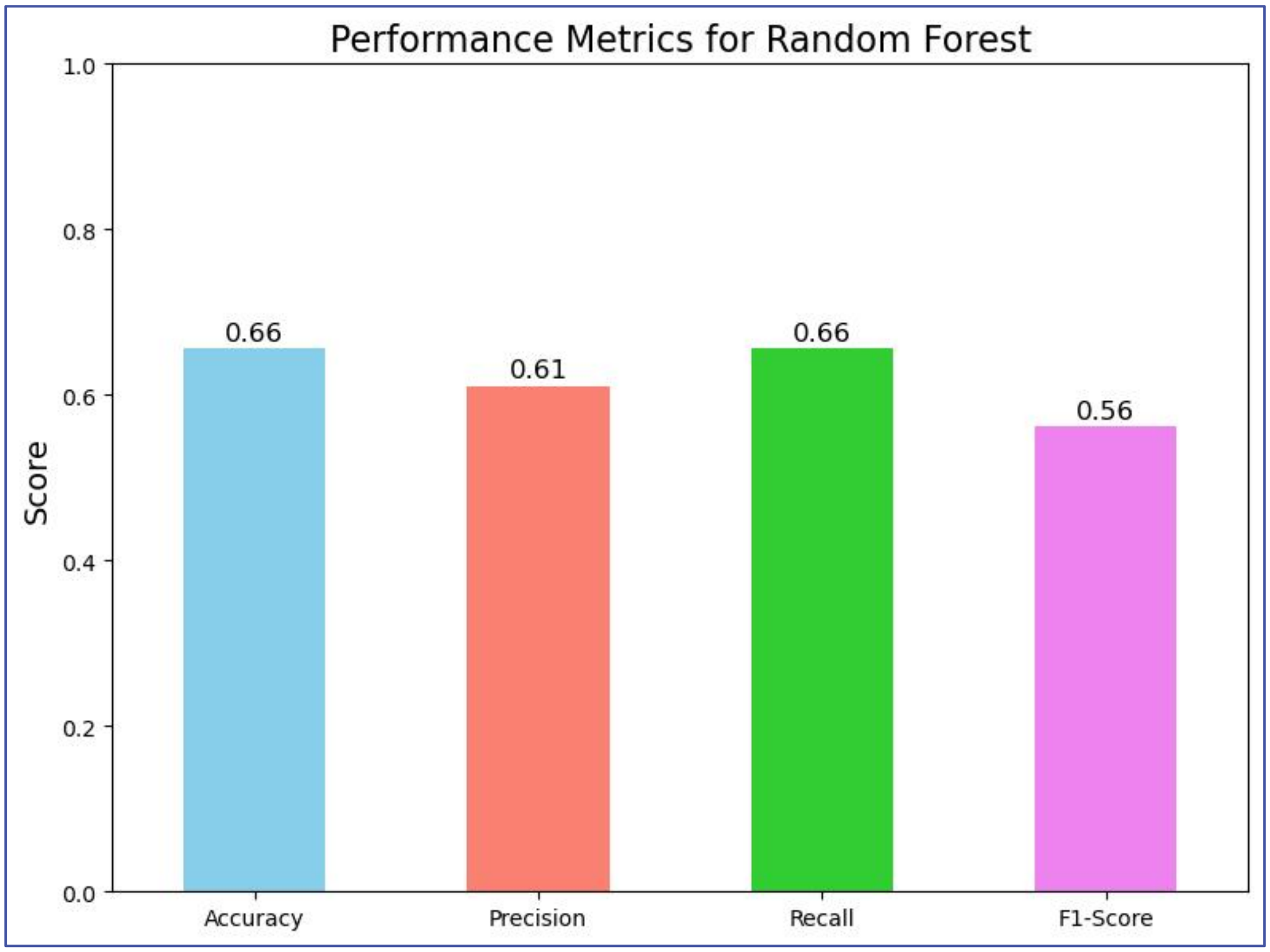

*Figure 34 Performance Metrics for RF.*

Class adjectif has shown poor performance with an AUC of 0.11 (Figure 35). Similarly, class verbe and class adverb indicated poor performance with the AUC of 0.49 and 0.58 respectively. Class '*mot*' has however indicated a rather moderate performance with the AUC of 0.64 (Figure 35). Majority of the classes have performed exceptionally well with the following AUCs per class: 0.96 for class '*nombre*', 0.97 for class '*article*' and '*pronompersonel*', 0.98 for class '*conjunction*' and 0.99 for class '*adverb or adverbe in French*'. Through analysis of the class performances, we get that the various parts of speech do impact the sentiment of the sentences by either amplifying or downplaying the sentiment based on the context.

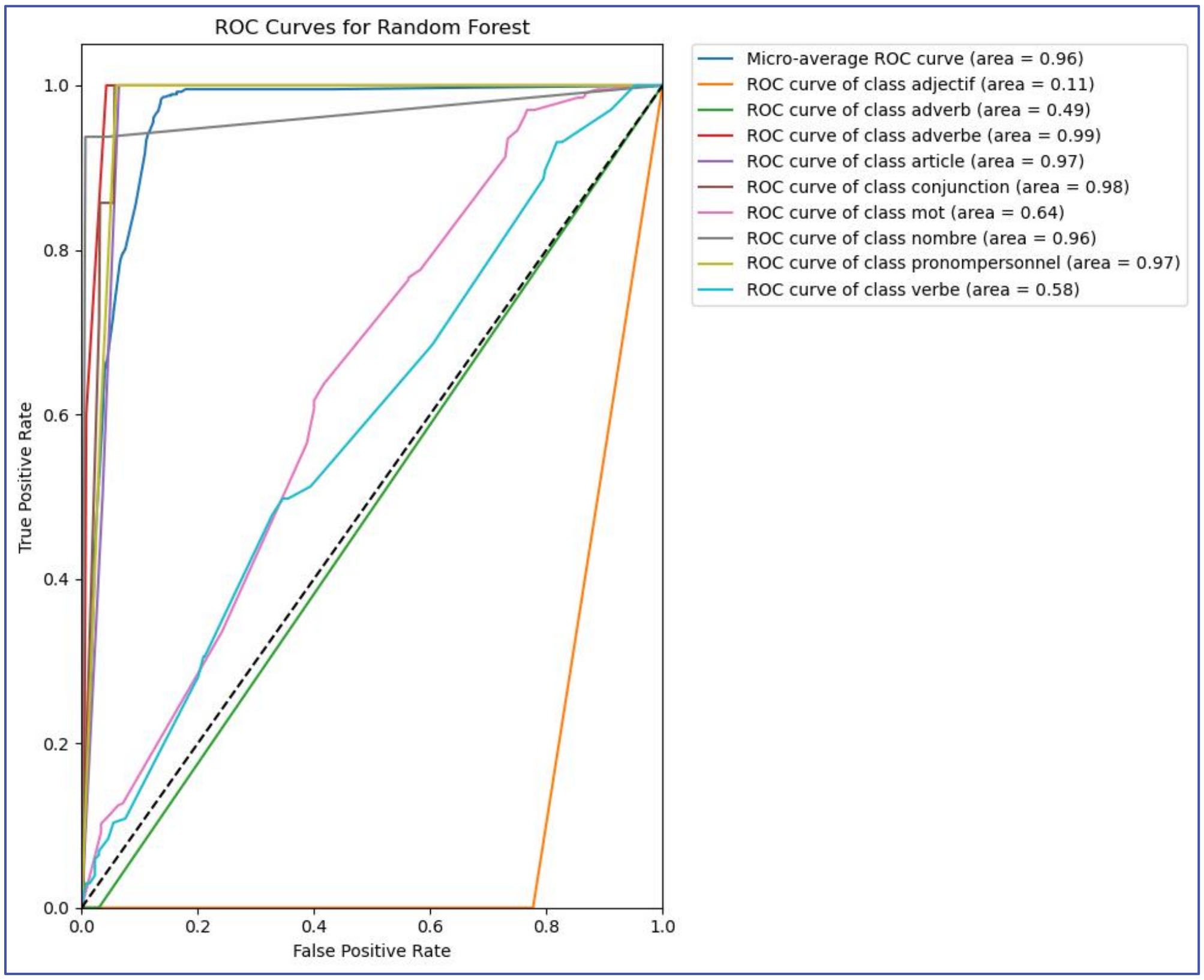

*Figure 35 ROC Curve for RF.*

The SVM precision focuses on the measures of true positives against all true positives obtained from the model prediction. Classes adjectif, adverb, adverbe, article, and pronompersonnel have not been predicted correctly at all (Figure 36-37). This is more likely due to the lower number of the classes during the randomized sampling processes. By looking at the support, it is clear that while these classes were present in the sampled dataset, they were of the least weight (Figure 36-37). Classes such as 'mot', 'nombre' and 'verbe' were present in large proportions and were classified with the weak to moderate precision (Table 2). Nombre class had about 2.49% portion of the Lexicon and was classified with high precision of 0.79 indicating the effectiveness of the model to identify the class well. The model overall aggregates the predictions very well when focusing on all classes altogether with the AUC of 0.95 (Figure 38-39). This shows that with all the classes combined, the model has learnt the dataset well and is able to differentiate sentiments with high accuracy such that the positively predicted sentiments are indeed positive from the Lexicon. However, getting to individual classes, the class 'adverb' and 'adjectif' perform poorly with the AUCs of 0.05 and 0.11 respectively (Figure 38-39). This indicates that while the model is able to distinguish the sentiments with all classes combined, it struggles to predict the sentiments when dealing with the classes 'adverb' and 'adjectif'. Classes 'verbe' and 'mot' are in large proportion of the dataset being second largest and largest portions of the Lexicon but still not easily classified with the AUCs just being 0.54 and 0.63 showing that the model still finds it difficult to accurately predict the sentiment.

### 4.3.2 SVM

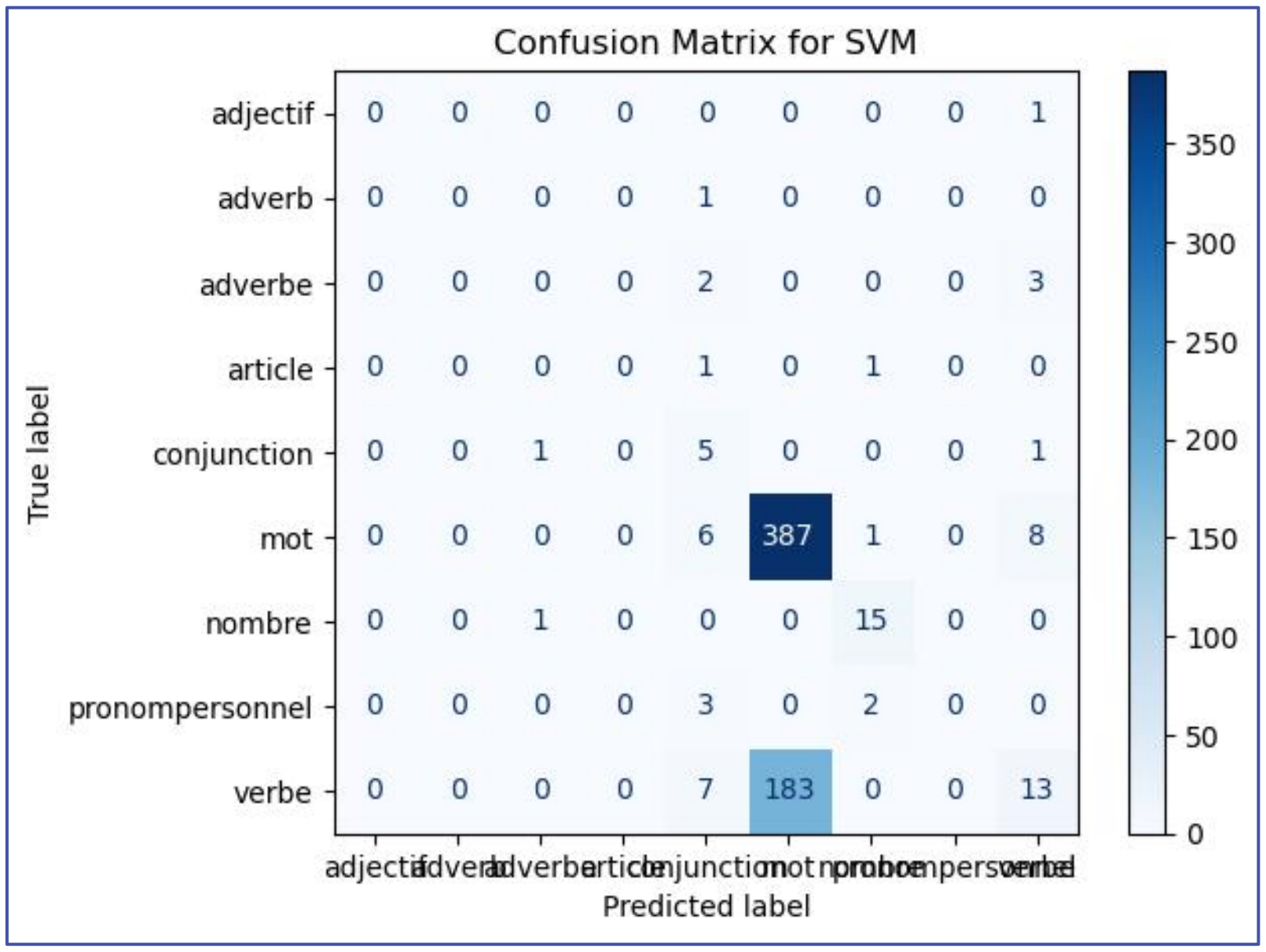

Figure 36 Confusion Matrix for SVM.

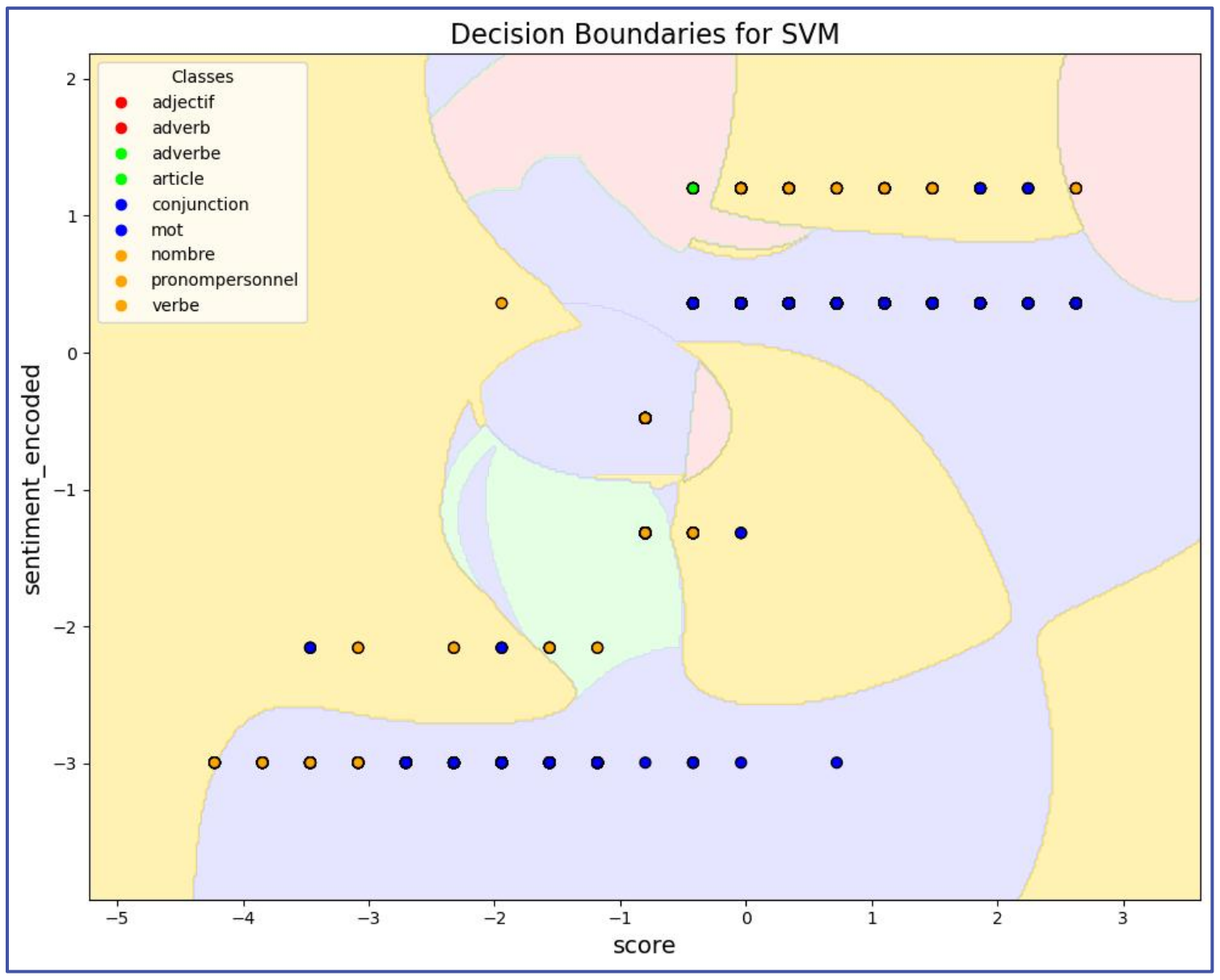

Figure 37 Decision Boundaries for SVM.

| | Precision | Recall | F1-score | Support |
|---|---|---|---|---|
| adjectif | 0.0 | 0.0 | 0.0 | 1 |
| adverb | 0.0 | 0.0 | 0.0 | 1 |
| adverbe | 0.0 | 0.0 | 0.0 | 5 |
| article | 0.0 | 0.0 | 0.0 | 2 |
| conjunction | 0.2 | 0.71 | 0.31 | 7 |
| mot | 0.68 | 0.96 | 0.8 | 402 |
| nombre | 0.79 | 0.94 | 0.86 | 16 |
| pronompersonnel | 0.0 | 0.0 | 0.0 | 5 |
| verbe | 0.5 | 0.06 | 0.11 | 203 |
| accuracy | | | 0.65 | 642 |
| macro avg | 0.24 | 0.3 | 0.23 | 642 |
| **weighted avg** | 0.61 | 0.65 | 0.56 | 642 |

Table 2 Performance Metrics for SVM.

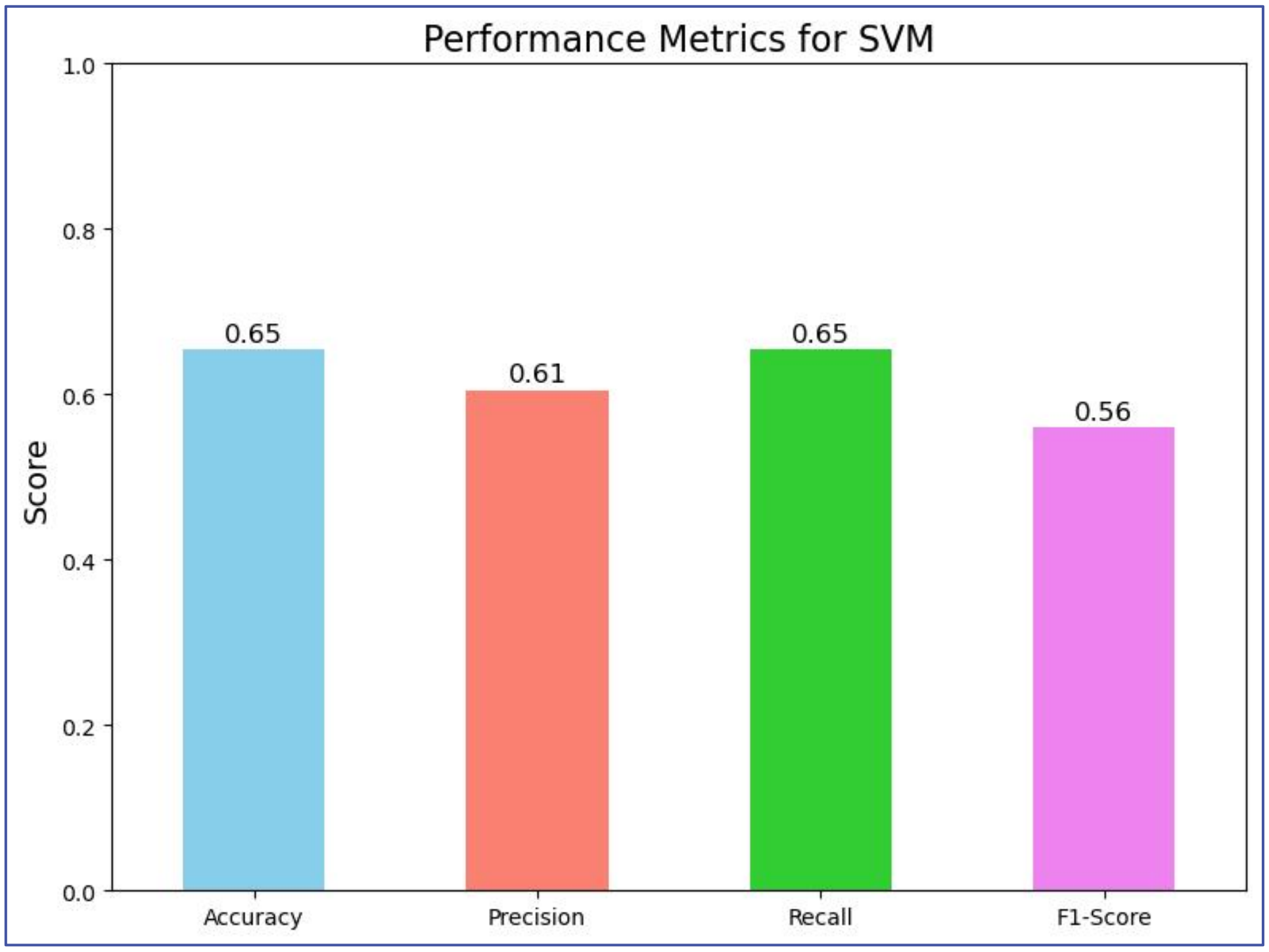

*Figure 38 Performance Metrics for SVM.*

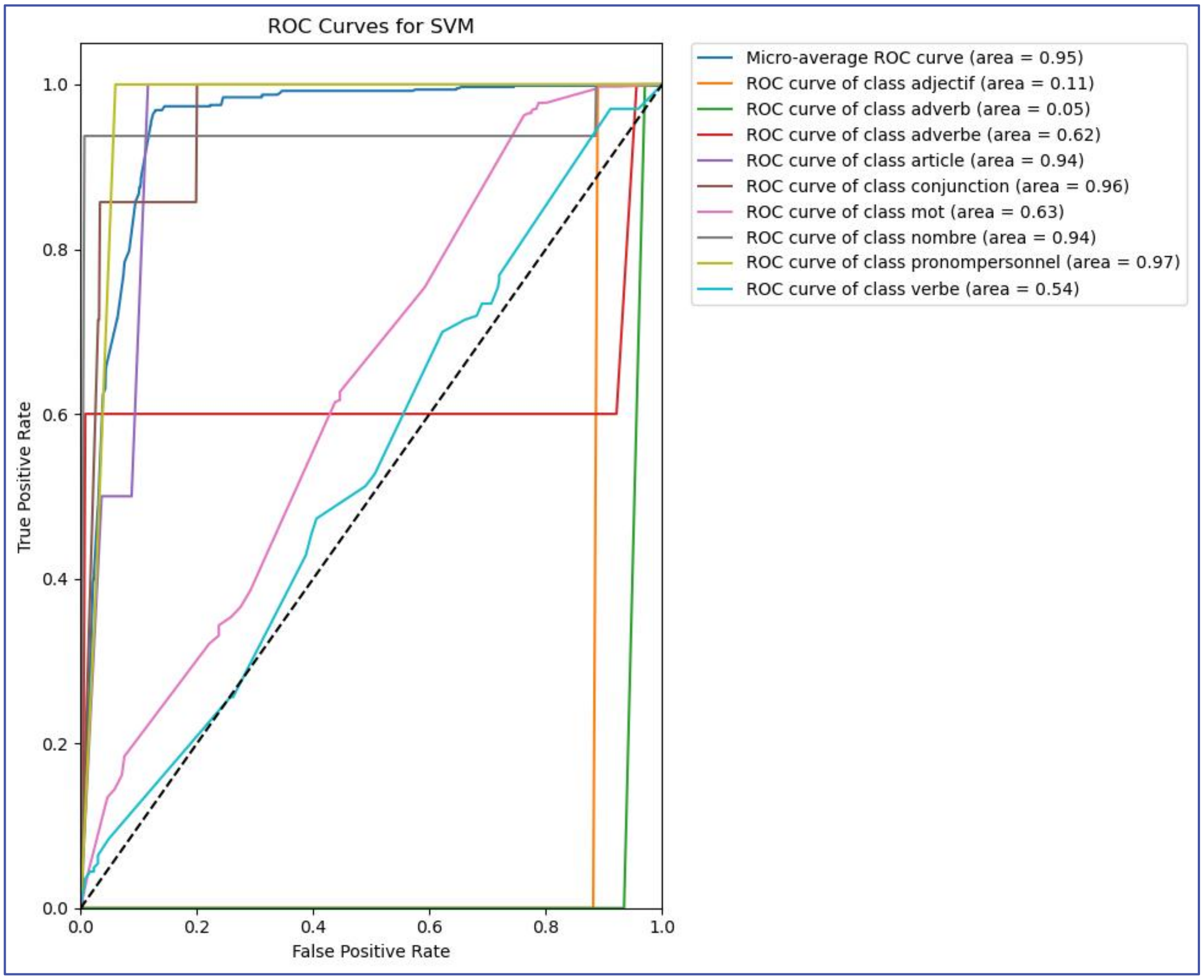

*Figure 39 ROC Curve for SVM.*

Class 'verbe' however was one of the low portions in the dataset sampled and still manages to predict the sentiment moderately well with just AUC of 0.64 (Figure 39). When looking at the proportion from the sampled data, we can say that the model is able to moderately handle the class imbalances but not with optimal performance. Class ‘nombre’, ‘article’, ‘conjunction’ and ‘pronompersonnel’ have however performed extremely well individually with AUCs of 0.94 for ‘nombre’ and ‘article’, 0.96 and 0.97 showing better performance of the model with high accuracy for the true positives against the actual true positive sentiments. The Naive Bayes confusion matrix illustrated in Figure 40 and 41 shows that the model is struggling with certain classes. For example, for precision and recall of categories like adverbs, adjectives, and conjunctions, the model shows zeros. There is also a poor performance for personal pronouns and verbs revealed by the model. For instance, the model shows recall of 0,12 and 0,6 for these categories (Table 3). Furthermore, the model shows a high performance for ‘mot/word’ classes with a recall of 0,86 and a precision of 0,67. The macro average compared to the weighted average shows low values for precision (0,21), F1-score (0,22) and recall (0,28). Thus, this highlights the imbalance of classes in the model and ‘Mot’ class that performs better in the Gaussian Naive Bayes model. The verbs and personal pronouns show that the model is struggling with contextual nuances of differentiating verbs and pronouns.

### 4.3.3 Gaussian Naïve Bayes

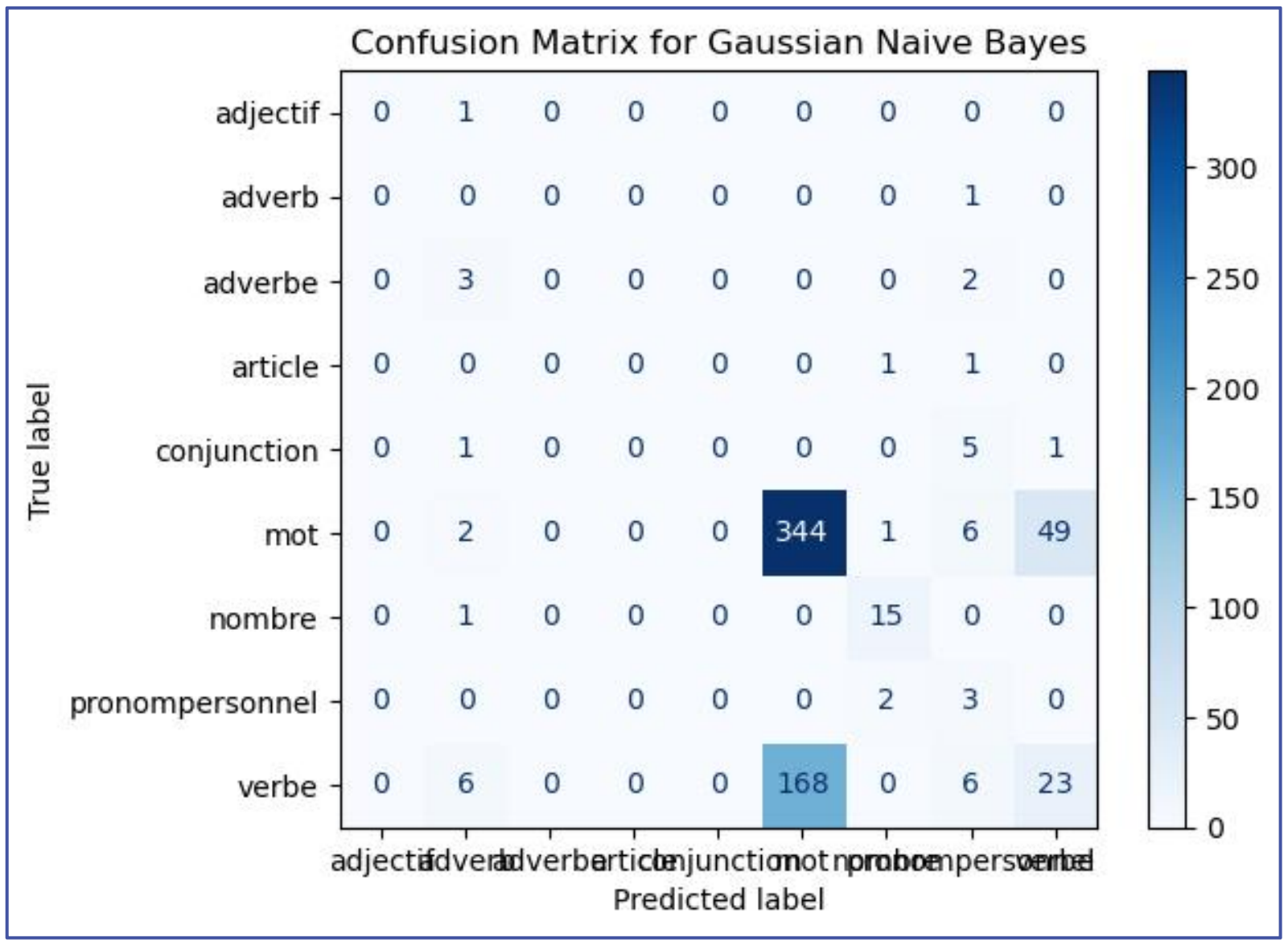

*Figure 40 Confusion Matrix for GNB.*

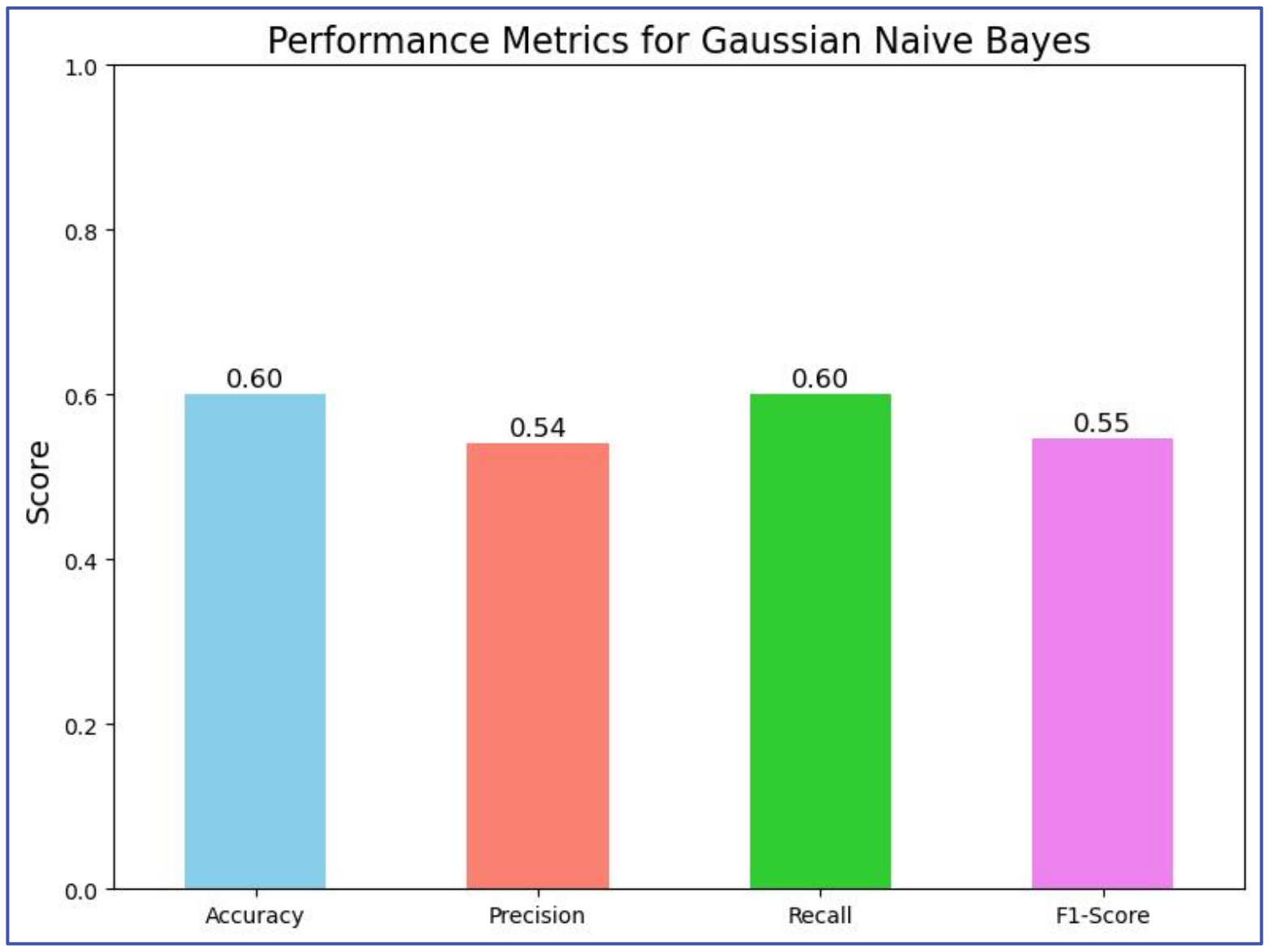

*Figure 41 Performance Metrics for GNB.*

| | Precision | Recall | F1-Score | Support |
|---|---|---|---|---|
| adjectif | 0.0 | 0.0 | 0.0 | 1 |
| adverb | 0.0 | 0.0 | 0.0 | 1 |
| adverbe | 0.0 | 0.0 | 0.0 | 5 |
| article | 0.0 | 0.0 | 0.0 | 2 |
| conjunction | 0.0 | 0.0 | 0.0 | 7 |
| mot | 0.67 | 0.86 | 0.75 | 402 |
| nombre | 0.79 | 0.94 | 0.86 | 16 |
| pronompersonnel | 0.12 | 0.6 | 0.21 | 5 |
| verbe | 0.32 | 0.11 | 0.17 | 203 |
| accuracy | | | 0.6 | 642 |
| macro avg | 0.21 | 0.28 | 0.22 | 642 |
| weighted avg | 0.54 | 0.6 | 0.55 | 642 |

*Table 3 Performance Metrics for Gaussian Naive Bayes (GNB).*

The model exhibits inconsistent performance across various linguistic categories due to disparities in precision and recall, especially evident in how it processes classes with different frequencies. For example, while "nombre" (number) achieves a high recall rate (0.94) and moderate precision (0.79), showing it is often recognised but sometimes confused with other classes, "verbe" (verb) shows significant weakness with an F1-score of just 0.17 (Table 3). This suggests the model performs unevenly, excelling in some areas but struggling to accurately classify less frequent classes. Such results indicate a tendency to favour high-frequency classes over sparse ones, compromising the model's reliability. The data imbalance further skews performance metrics, with a noticeable gap between macro and weighted averages. While weighted averages slightly elevate the overall scores due to frequent class representation, macro averages demonstrate the model's poor handling of underrepresented classes (Table 3). The limited support for classes like "adjectif" and "adverb" results in inadequate generalisation across categories. Improving the model's effectiveness may require addressing this imbalance, possibly by rebalancing data or exploring alternative methods better suited to sparse classes (Figure 42).

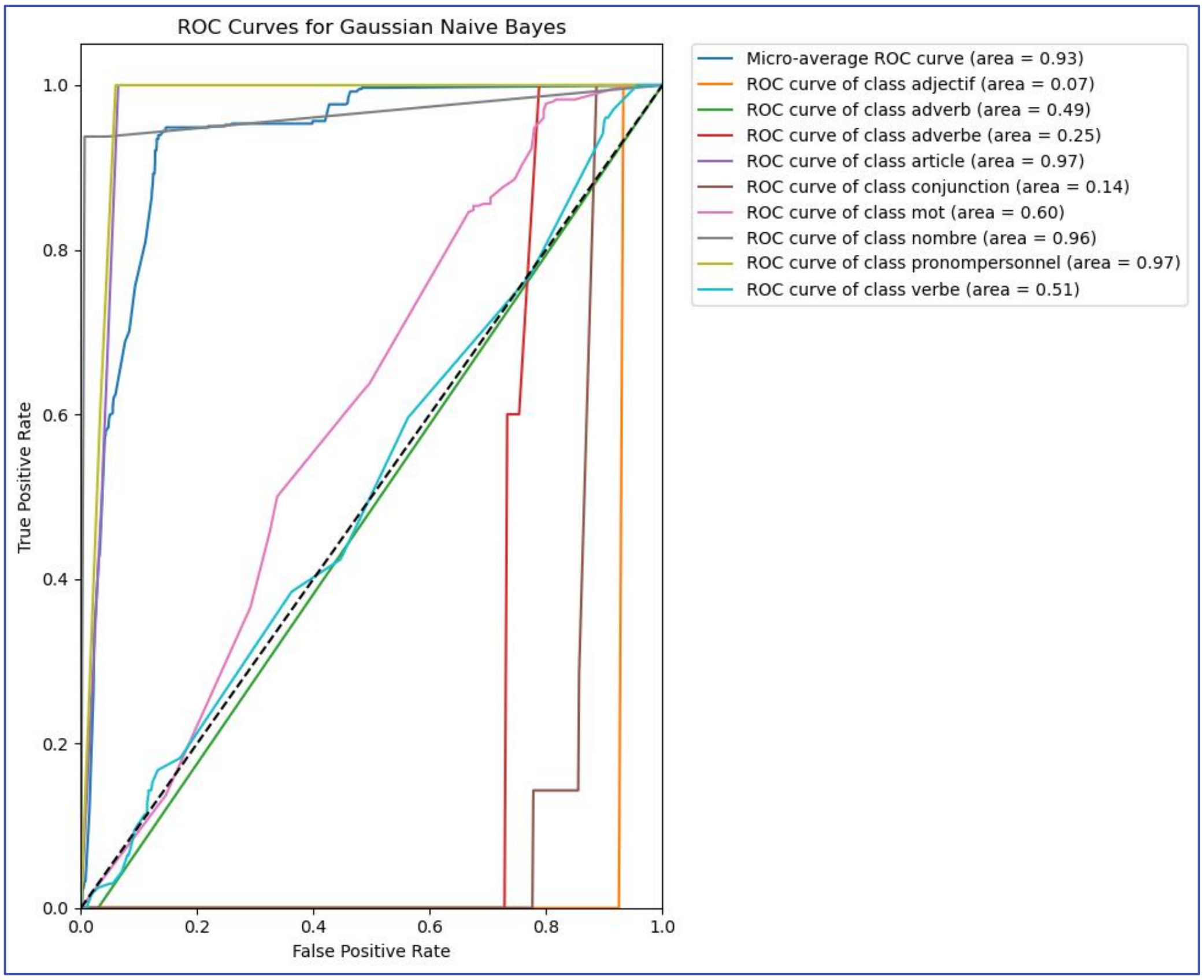

*Figure 42 ROC Curve for GNB.*

The ROC curve for Gaussian Naive Bayes evaluates true positive rates against false positives across thresholds, revealing a moderate effective classification capability but with notable limitations. A moderately high area under the curve score suggests some capacity to distinguish between classes, yet the model's independence assumption among features likely limits the performance. Class imbalance affects the curve, as high-frequency classes inflate the model's apparent effectiveness, thereby masking poorer accuracy for minority categories. The curve also highlights areas where adjusting thresholds could improve recall or precision based on specific needs, such as adopting a more lenient threshold for harder-to-detect classes like verbs or stricter settings for common terms. Additionally, bias in classification favours classes with greater support like "mot" and "nombre," reducing the true positive rate for less frequent classes. Thus, the ROC analysis suggests moderate performance overall, with potential for improvement by recalibrating thresholds, particularly to enhance specificity for underrepresented categories (Figure 42).

### 4.3.4 Decision Tree

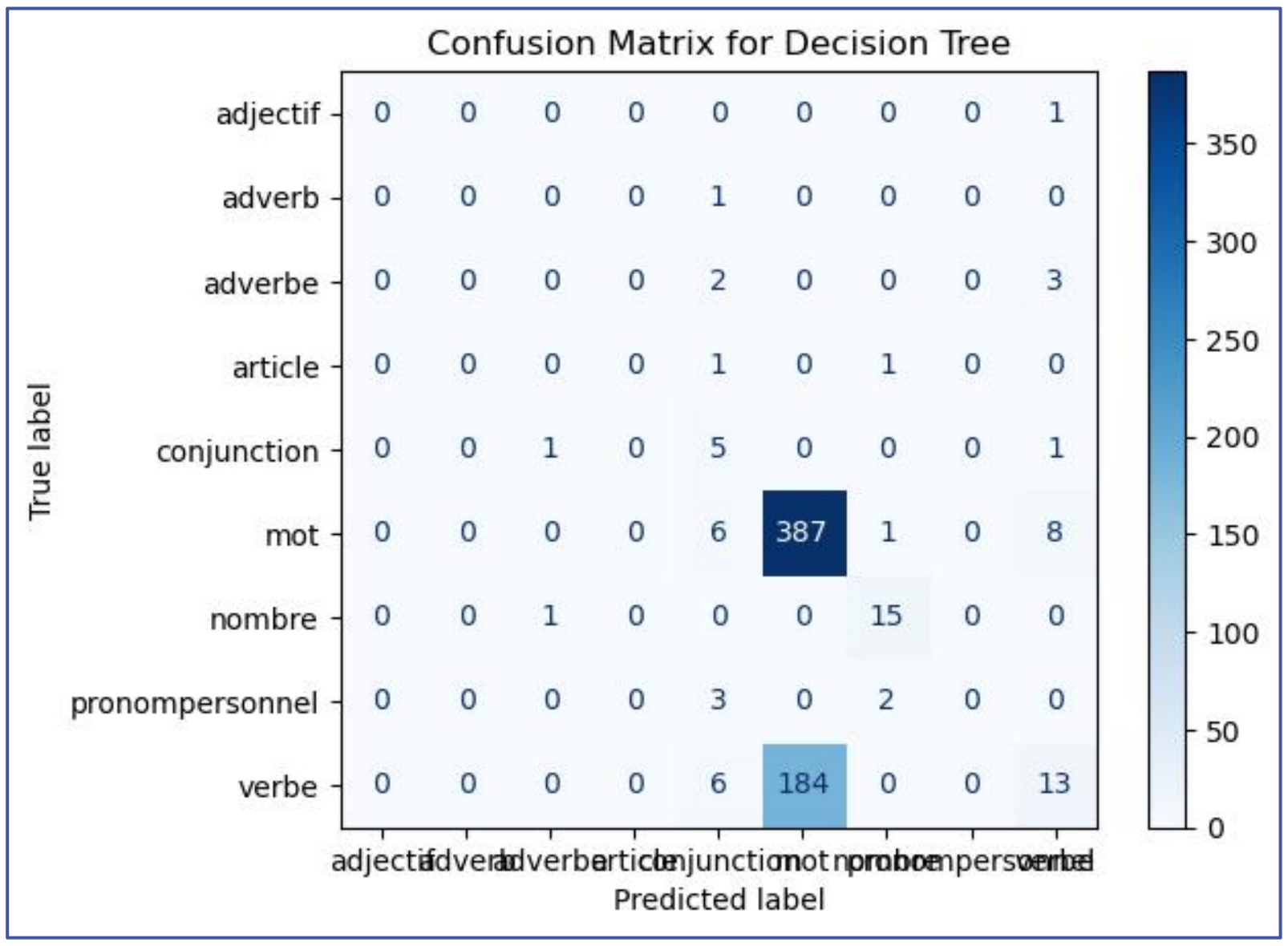

*Figure 43 Confusion Matrix for Decision Tree.*

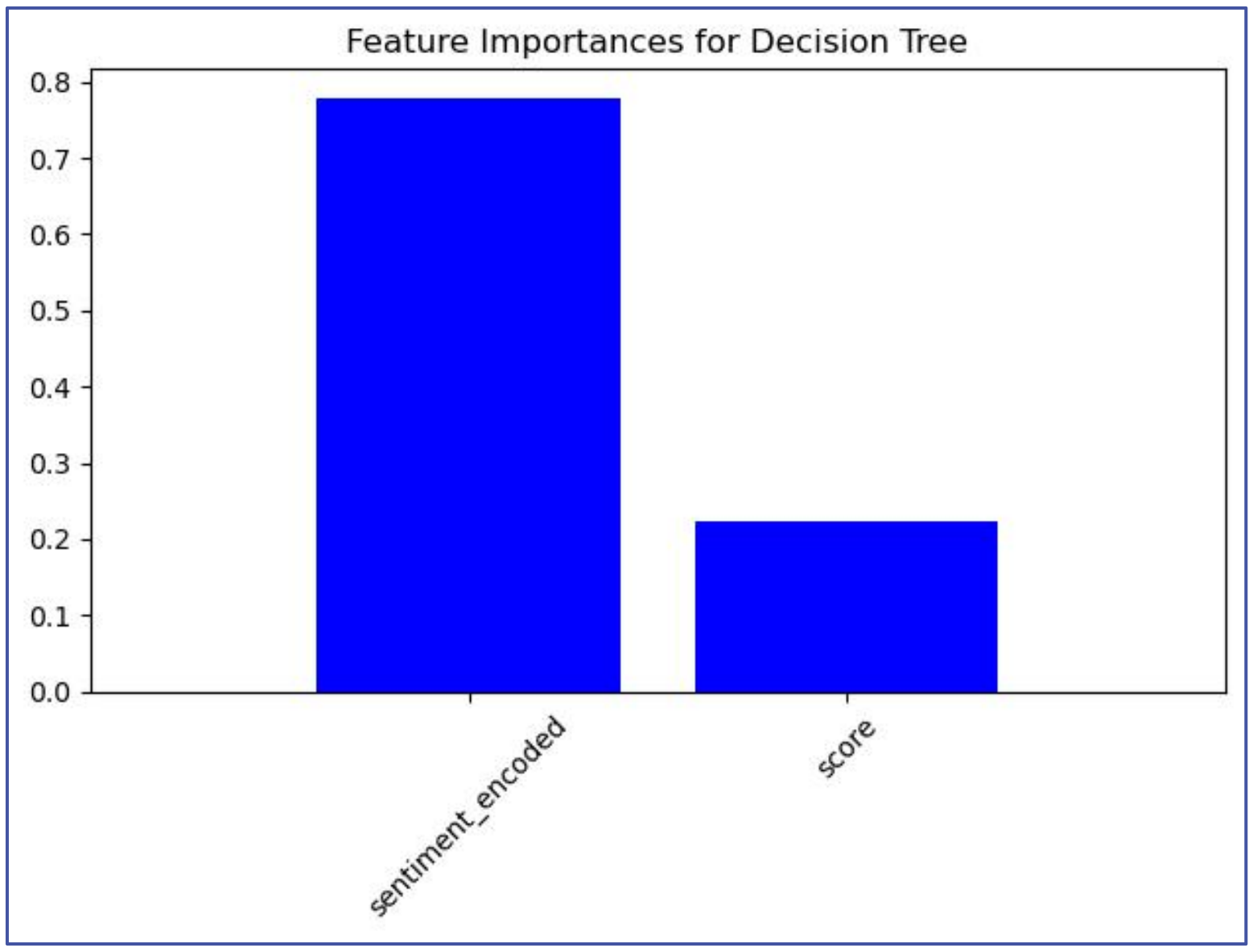

*Figure 44 Feature Importance for Decision Tree.*

The confusion matrix shown in Figure 43 presents the model performance over the different classes within the decision tree. The model had poor performance over most classes. Adjectif, adverb, adverbe, article, and pronompersonnel had no correctly identified values. This is most likely due to the lack of samples available in these classes, with each class listed having poor support. The model performed well in the mot, nombre and verbe classes, identifying 387 mot values, 15 nombre values and 184 verbe values correctly. Overall, the model performed poorly, however it was able to identify mot, nombre and verbe values (Table 5).

| | Precision | Recall | F1-Score | Support |
|---|---|---|---|---|
| adjectif | 0.0 | 0.0 | 0.0 | 1 |
| adverb | 0.0 | 0.0 | 0.0 | 1 |
| adverbe | 0.0 | 0.0 | 0.0 | 5 |
| article | 0.0 | 0.0 | 0.0 | 2 |
| conjunction | 0.21 | 0.71 | 0.32 | 7 |
| mot | 0.68 | 0.96 | 0.8 | 402 |
| nombre | 0.79 | 0.94 | 0.86 | 16 |
| pronompersonnel | 0.0 | 0.0 | 0.0 | 5 |
| verbe | 0.5 | 0.06 | 0.11 | 203 |
| accuracy | | | 0.65 | 642 |
| macro avg | 0.24 | 0.3 | 0.23 | 642 |
| weighted avg | 0.6 | 0.65 | 0.56 | 642 |

*Table 5 Performance Metrics for Decision Tree.*

Table 5 shows the accuracy metrics for the Decision Tree model. Adjectif, adverb, adverbe, and pronompersonnel had 0.0 across all metrics, indicating very poor model performance. However, this can be attributed to the lack of support for these classes. Each of the named classes had support values less than 10, indicating that the model had a very small sample size of classes. This resulted in poor performance. The model performed slightly better in the conjunction class, with a precision value of 0.21, a recall value of 0.71 and an f1 score of 0.32. This indicates that the model identifies true conjunctions 71% of the time. However, the precision is low, suggesting that a large number of the conjunction predictions were incorrectly classified (Figure 45).

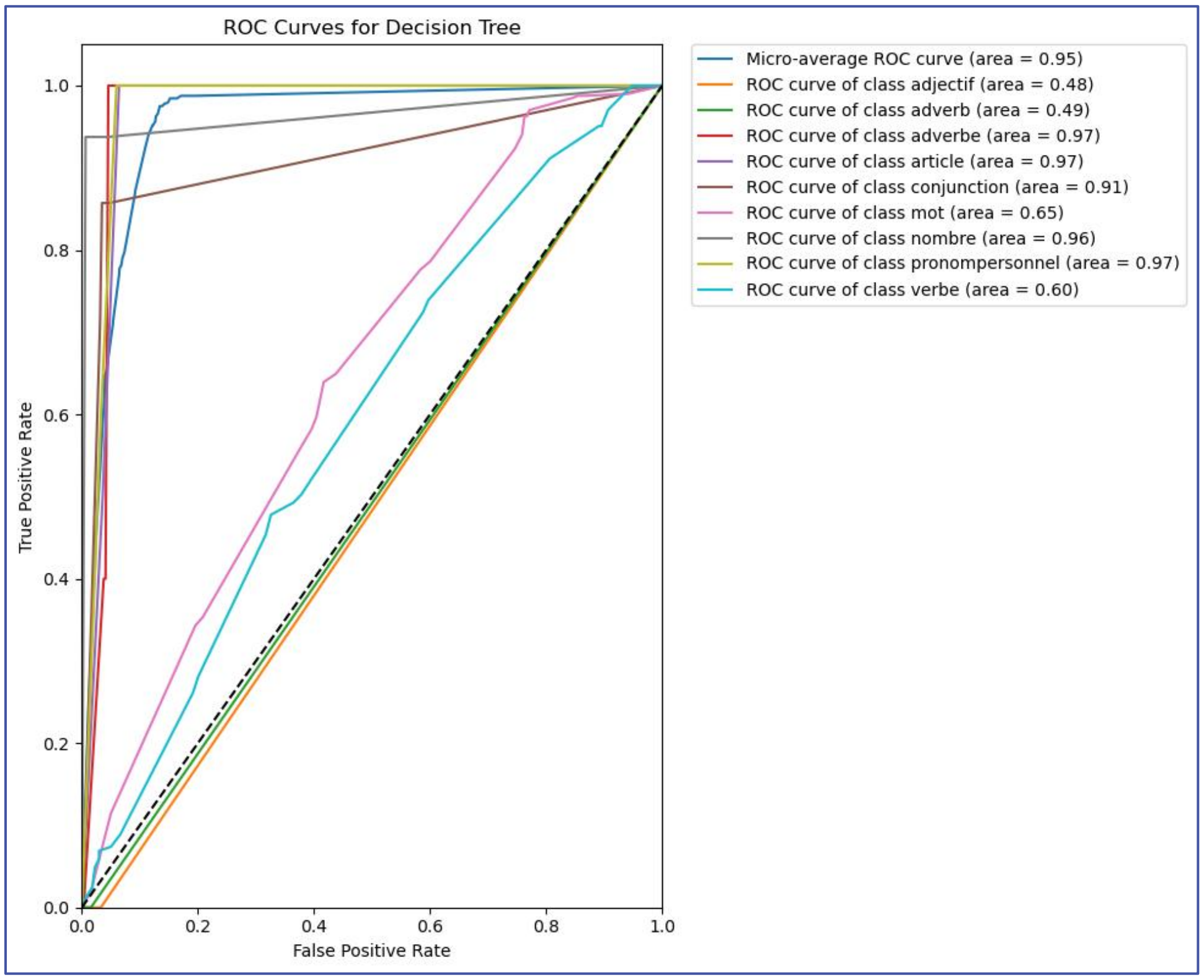

*Figure 45 ROC Curve for Decision Tree.*

The model performs fairly well in the ‘mot’ class, with a precision of 0.68, recall of 0.96 and an F1 score of 0.8 (Table 5). The precision indicates when the model predicts a value as belonging to the “mot” class, it is correct 68% of the time. However, the model still identifies other classes as the “mot” class. The recall of 0.96 indicates that the model identifies nearly all the “mot” values, with very few negatives. The F1 score is a balance of both precision and recall, and a score of 0.8 shows that the model performs fairly well, with regards to the “mot” class (Table 5). The model performs well in terms of the ‘nombre’ class, with a with a precision of 0.79, recall of 0.94 and an F1 score of 0.86 (Table 5). This shows that the model can effectively identify values in the ‘nombre’ class 79% of the time, while still identifying values as the nombre class, when they belong to different classes. A recall of 0.96 indicates that the model identifies nearly all the ‘nombre’ values. The F1 score is a balance of both precision and recall, and a score of 0.86 shows that the model performs fairly well when classifying values in the nombre class. The model performs poorly in the ‘verbe’ class, with a precision of 0.5, a recall of 0.06, and an F1 score of 0.11. It identifies values in the ‘verbe’ class 50% of the time, with a number of false positives. A recall of 0.06 indicates very poor performance when identifying values from the ‘verbe’ class. Overall, the model performs with a weighted average of 0.6, indicated fair model performance. It performs poorly in most of the classes, due to the lack of support. However, in the classes with adequate support, the model performs better. The Lexicon does not have enough representation over the different classes, resulting in poorer model performance. Decision trees with multiple classes performs very well and can predict the polarities of the sentiments with moderately high accuracy based on its root to leaf node approach in decision making covering about 0.95 of AUC (Figure 45). Individual classes adjectif and adverb performed poorly with the AUCs of 0.48 and 0.49 indicating that the model struggles with predicting the poorly represented class (Figure 46). Class ‘verbe’ and ‘mot’ had moderate success in predicting the true positives covering just about 0.60 and 0.64 of AUC. The majority of the classes performed well covering good areas. This has shown the bias towards the majority of the well-presented classes that were sampled.

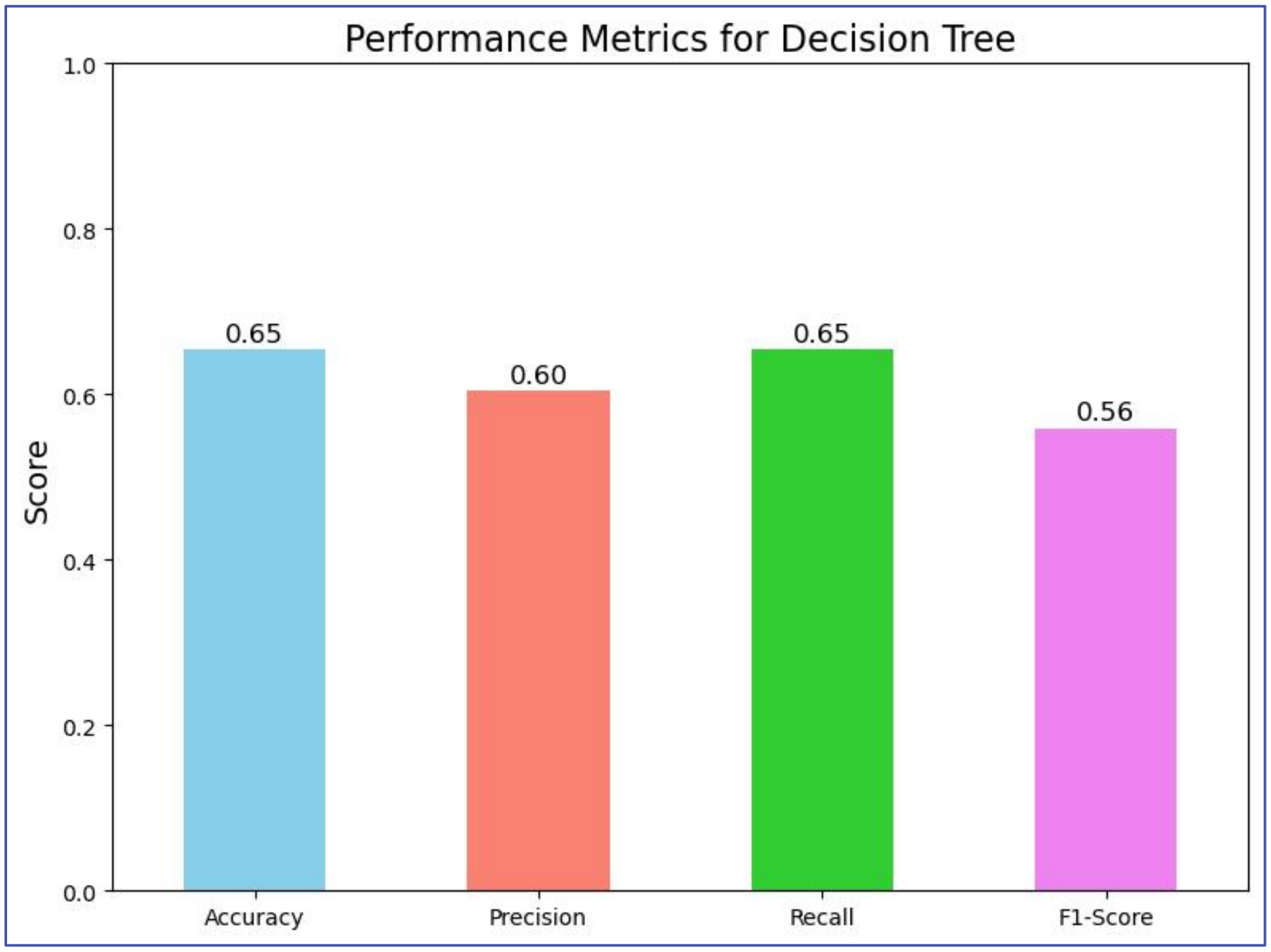

*Figure 46 Performance Metrics for Decision Tree.*

### 4.3.5 Comparison

| Model | Accuracy | Precision | Recall | F1-Score |
|---|---|---|---|---|
| Random Forest | 0.655763 | 0.611037 | 0.655763 | 0.561988 |
| SVM | 0.654206 | 0.605091 | 0.654206 | 0.559285 |
| Decision Tree | 0.654206 | 0.604437 | 0.654206 | 0.558882 |
| Gaussian Naive Bayes | 0.599688 | 0.54098 | 0.599688 | 0.547012 |

*Table 6 Model Comparison.*

The performance matrix highlights the strengths and weaknesses of Random Forest, SVM, Decision Tree, and Gaussian Naive Bayes models. Random Forest leads with the very best accuracy (0.66) and F1-score, demonstrating robustness against elegance imbalance because of its ensemble method, which improves generalisation and performance consistency (Table 6). SVM carefully follows, with an accuracy of 0.654. This model successfully handles complicated limitations, even though it could war with pretty imbalanced classes, as its binary margin optimisation limits adaptability to numerous elegance frequencies (Table 6). Gaussian Naive Bayes, with the lowest accuracy (0.599), is hindered by its simplicity and reliance on feature independence, which does not completely suit the lexicon. This model's moderate precision makes it a less optimal choice for tasks requiring nuanced classification- (Table 6). Trade-offs in precision and recall suggest SVM and Decision Tree slightly sacrifice precision for higher recall, helpful in cases where capturing true positives is prioritised. While Gaussian Naive Bayes might work when precision is critical, Random Forest generally offers the most balanced performance across metrics, making it the preferred choice for this task (Figure 47).

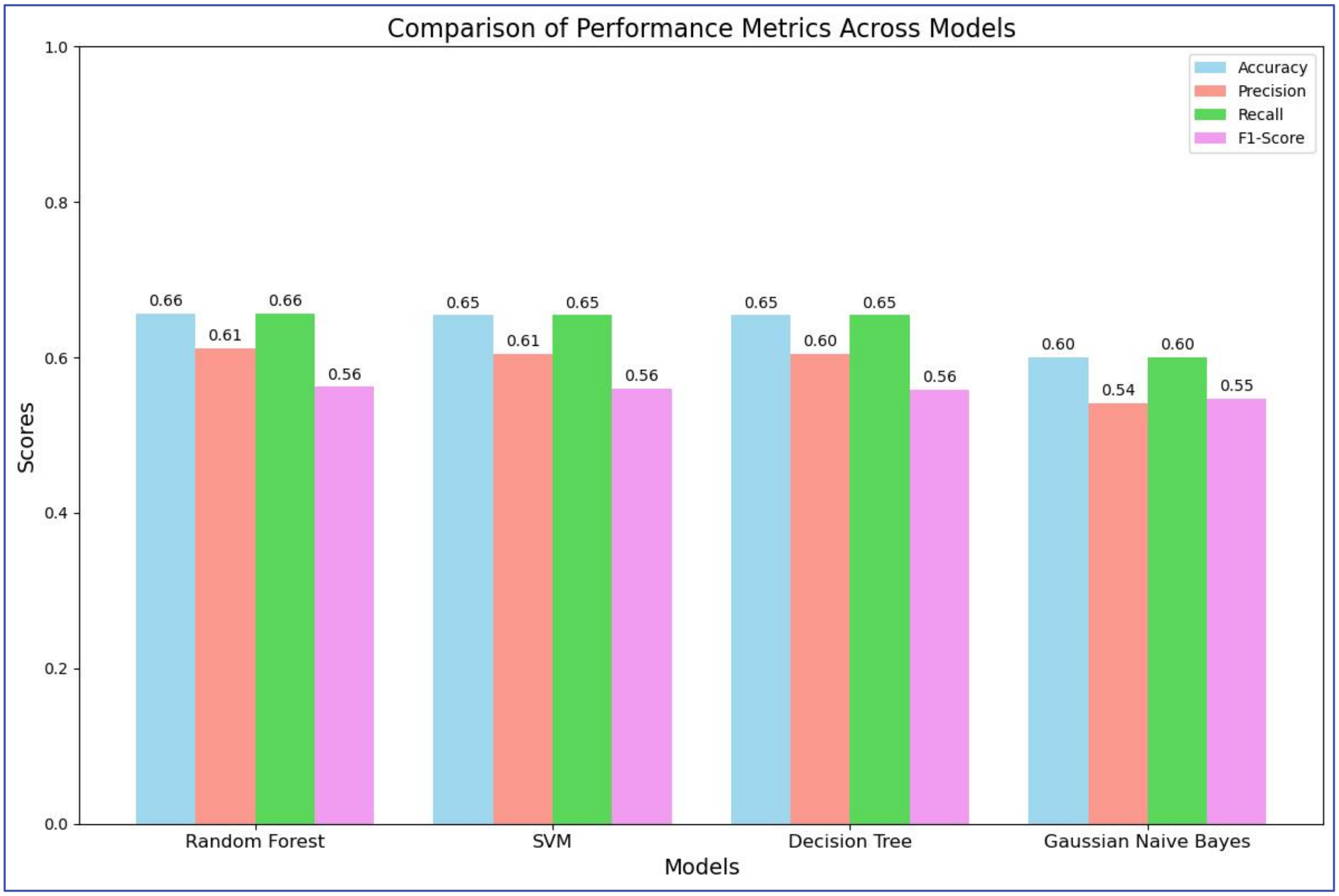

*Figure 47 Model Comparison.*

## 4.4 Aspect-Based Sentiment Analysis

### 4.4.1 BERT Model Performance

| Sentiment | Precision | Recall | F1-Score | Support |
|---|---|---|---|---|
| Negative | 0.98 | 1 | 0.99 | 151 |
| Neutral | 0 | 0 | 0 | 3 |
| Positive | 1 | 1 | 1 | 154 |
| Accuracy | | 0.99 | | 308 |
| Macro avg | 0.66 | 0.67 | 0.66 | 308 |

| Weighted avg | 0.98 | 0.99 | 0.99 | 308 |
|---|---|---|---|---|

Table 7 BERT Performance Metrics.

In Aspect-based sentiment analysis, a BERT model was used in order to classify the sentiment of a word contextually. Table 7 shows the accuracy metrics for the BERT model. In terms of the negative sentiment, the model has a high precision of 0.98. This indicates that the BERT model identified the negative data points correctly 98% of the time. The negative sentiment had a recall of 1, indicating that the model correctly identified all the negative sentiment cases 100% of the time (Table 7). The F-1 Score is the harmonious mean of both the precision and the recall. A F-1 score of 0.99 indicates that the BERT model performed very well in identifying the negative sentiments. The model correctly identified 151 negative sentiments as negative (Table 7). The neutral sentiment score had a value of 0 across all accuracy metrics. This indicates that the model did not successfully classify any values as neutral. However, there were only 3 cases of a neutral sentiment within the model. This could suggest the need for more neutral cases to be included, or that a resulting sentiment of zero is not common. The positive sentiment has a score of 1 across all accuracy metrics. This indicates perfect performance for the positive sentiment. It identified all 154 positive sentiment cases and classified them correctly. Overall, the model performance suggest that BERT can effectively distinguish between positive and negative sentiment values, contextually (Figure 48-49). It has difficulty with the neutral sentiment, likely due to the small sample size. The high weighted average across all metrics indicates a well performing model.

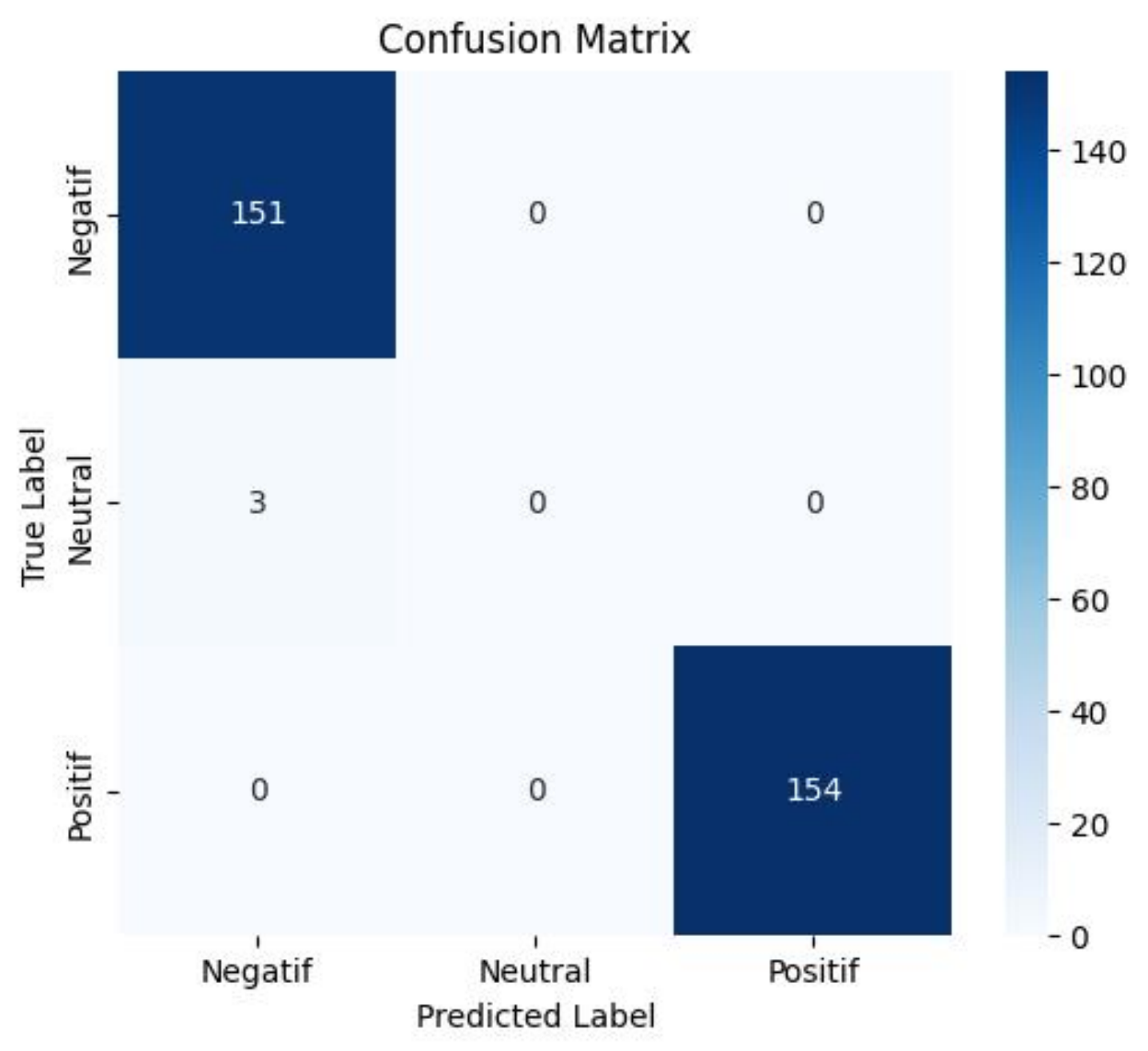

Figure 48 Confusion Matrix for BERT.

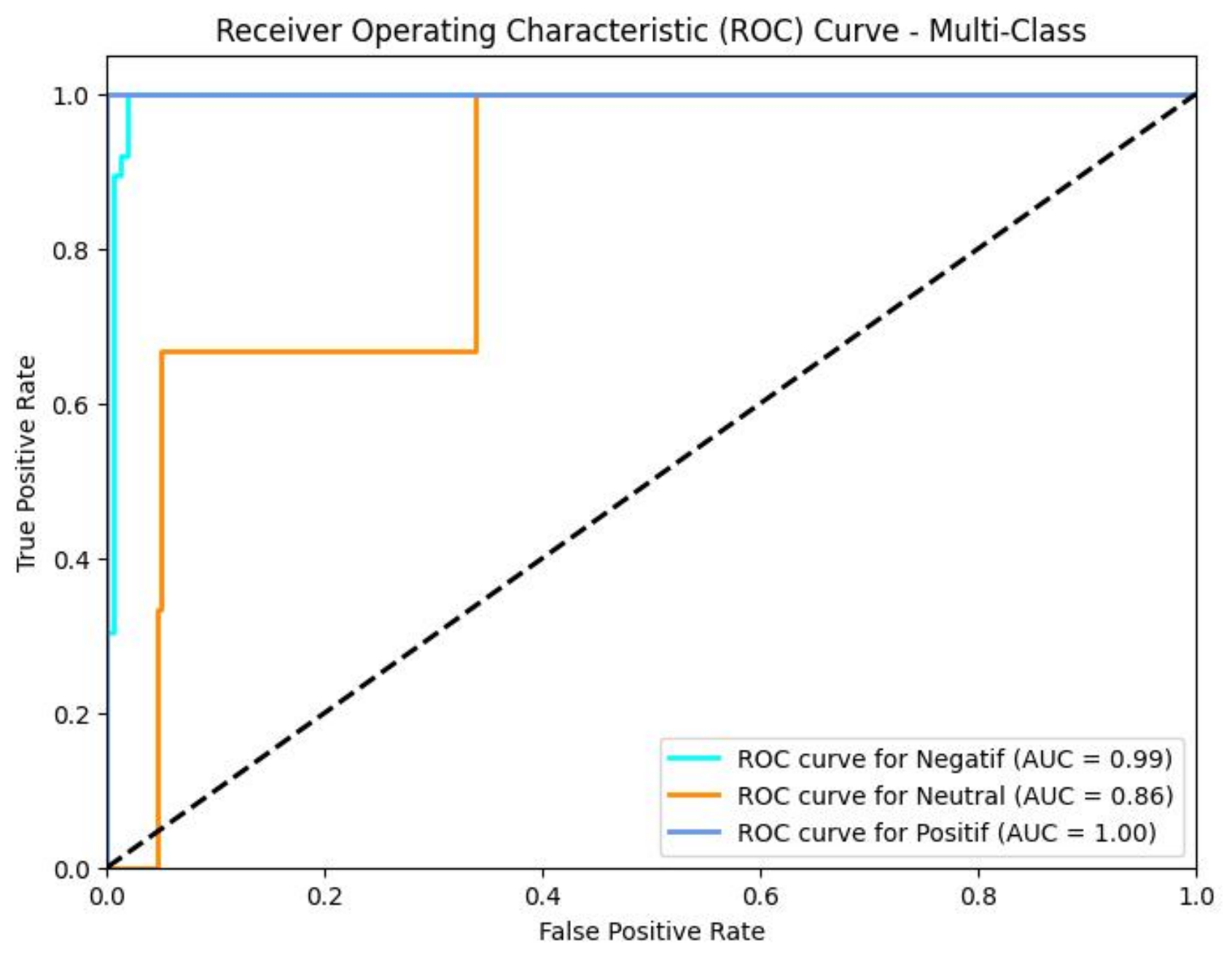

Figure 49 ROC Curve for BERT.

Figures 48-49 show the Confusion Matrix for the BERT model. In terms of the negative sentiment, the model correctly identified 151 instances of true negative sentiments. There were no false positives as all negative instances were classified as negative. The neutral sentiment had no correctly identified neutral values, and they were classified as negative sentiments. This suggests that the model struggles to identify neutral instances. The positive sentiment had all 154 instances classified correctly, showing perfect classification accuracy. There were no instances of false positives. The curve shows the True Positive Rate against the False Positive Rate. The Area Under the Curve indicates the model performance, with higher values indicating a better model performance. The negative curve has an AUC of 0.99, indicating that the model can nearly perfectly identify negative sentiment values (Figure 49). The AUC for the neutral classes is 0.86, indicating that the model struggles more with the neutral sentiment, which is to be expected due to the lower number of neutral samples in the Lexicon. The positive sentiment AUC is 1, indicating that the model perfectly classifies positive sentiment values (Figure 49).

### 4.4.2 Integrated Gradients XAI

Four sentences from the testing dataset were randomly selected to be processed with integrated gradients as an XAI technique to identify the attribution of different words in the sentence to the prediction of the sentiment for the target word. These sentences are shown in Table 8.

| **Marked Sentence** | **Predicted Sentiment** | **Attribution** | **Convergence Delta** |
|---|---|---|---|
| [TARGET] accuse [/TARGET] was instrumental in organising the event. | Positif (Confidence: 0.63) | 0.08 | -0.8459 |
| [TARGET] earth [/TARGET] is the third planet from the sun. | Negatif (Confidence: 0.93) | 0.0962 | 1.8049 |
| [TARGET] accuse [/TARGET] has been involved in multiple community projects. | Negatif (Confidence: 0.70) | 0.0534 | 1.007 |
| [TARGET] earth [/TARGET] provides the necessary resources for life. | Positif (Confidence: 0.83) | 0.0033 | -2.9139 |

*Table 8 Testing Sentences for Sentiment Analysis.*

The heatmap illustrated in Figure 50 shows the attributions for "[TARGET] accuse [/TARGET] was instrumental in organising the event." The heatmap depicts that the highest attributions are for the words *accuse* and *event*, while the rest of the words have an attribution of around 0.5. The lowest attribution is for "*was*".

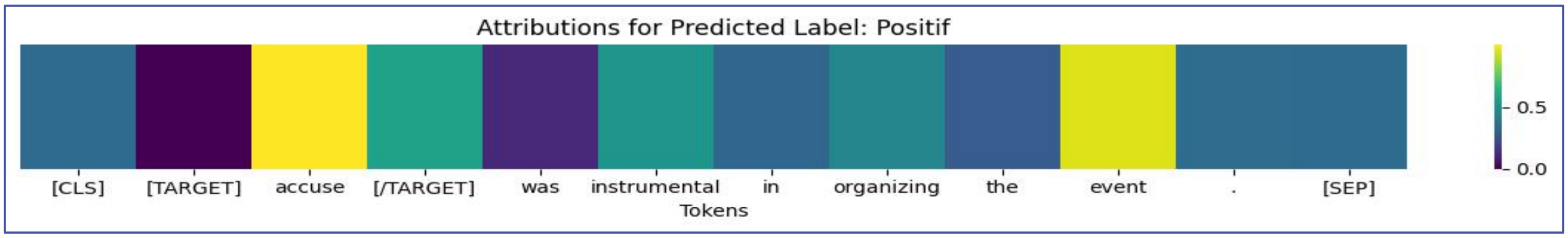

*Figure 50 Attribution Heatmap for Sentence 1.*

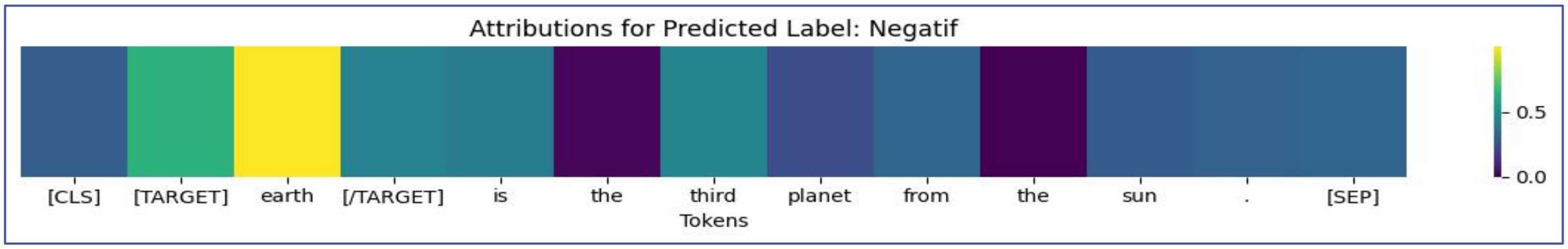

*Figure 51 Attribution Heatmap for Sentence 2.*

Figure 51 shows the attributions for “[TARGET] earth [/TARGET] is the third planet from the sun” The heatmap shows that the highest attribution is for the word *earth*, while the rest of the words have an attribution of around 0.5. The lowest attribution is for “*the*” with an attribution of 0.0.

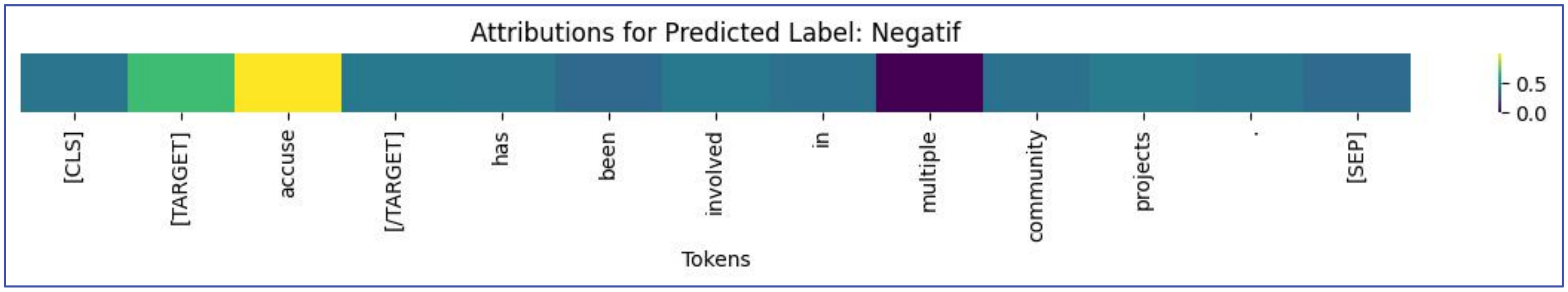

*Figure 52 Attribution Heatmap for Sentence 3.*

Figure 52 illustrates the attributions for “[TARGET] accuse [/TARGET] has been involved in multiple community projects.” The heatmap shows that the highest attribution is for the word “accuse”, while the rest of the words have an attribution of around 0.5. The lowest attribution is for “multiple” with an attribution of 0.0.

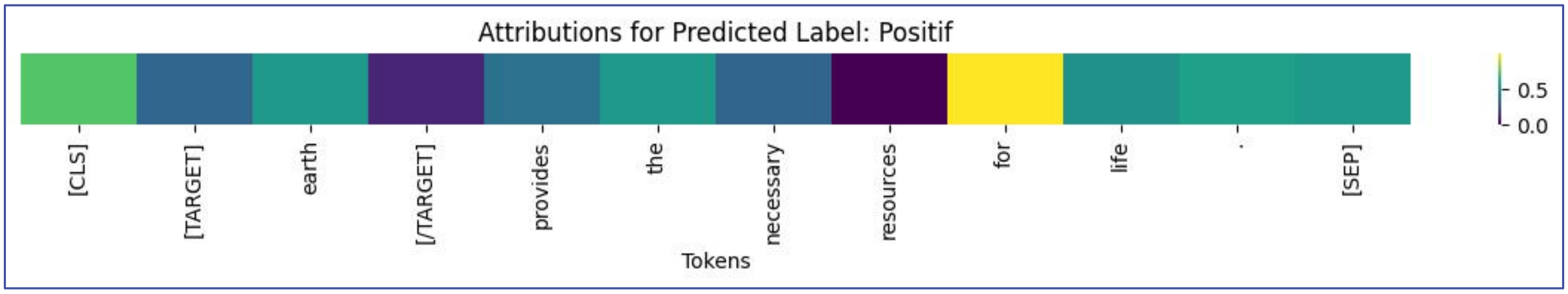

*Figure 53 Attribution Heatmap for Sentence 4.*

Figure 53 shows the attributions for “[TARGET] earth [/TARGET] provides the necessary resources for life”. The heatmap shows that the highest attribution is for the word “for”, while the rest of the words have an attribution of around 0.5. The lowest attribution is for “resources” with an attribution of 0.0.

# 5. DISCUSSION

## 5.1 Lexicon Expansion

The lexicon initially was based on two languages, Ciluba and French. This needed additional languages and four were added for this cause, directly translating from French to English to understand the sentiment from one language to another. It was discovered that the majority of the words had a positive sentiment. Through direct translation from English to the other languages, it was discovered that the words had conjunctions but were not enough requiring additional words to account for the poorly represented negative sentiments and the conjunctions. The addition of the words helped when translations were done across the various languages moving from one language to the next and the reverse using different phrases.

## 5.2 Translation Functions

To test for the translation power of the lexicon, multiple sentences per language were created. This was tested across four official languages of South Africa and two official languages from the DRC. Based on the translations, we discovered that while the lexicon was expanded, the translations were accurate to some extent due to some of the words having a different or similar meaning when translated from one language to another changing the context of the phrase. This could be due to the fact that while direct translation is preferred, there is a trade-off of contextual value from one language to the next, the lack of word to word direct translation and different meaning and use of idioms and homonyms across the languages (Orkphol & Young, 2019). While this was the issue encountered, some direct translations are fully accurate especially when the phrases are short or limited to a few words.

## 5.3 Machine Learning Sentiment Analysis

Sentiment analysis has become a crucial aspect when it comes to grasping the emotion conveyed by the textual data obtained from multicultural languages during the reviews of the products used globally (Muhammadi, Laksana, & Arifa, 2022). The analysis of the machine learning models reveals that Random Forest is the most robust and balanced model for this sentiment classification task across LRLs, likely due to its ensemble nature, which mitigates class imbalance. This is enhanced by the nature of the model itself, based on the multiple decision trees that are less sensitive to noise. The decision trees are able to detect the polarity of the sentiment being positif, negatif or neutral. The model had the capabilities of determining the sentiment polarity but struggled to determine the actual sentiment value. The Gaussian Naive Bayes model is easily interpretable. However, it struggles to learn and deal with the complex, nuanced features of linguistic data, rendering it less effective for this task.

## 5.4 Implementation of a LLM for Sentiment Analysis

LLM architectures are useful in enabling the aspect-based sentiment analysis to distinguish between the sentiments and aid in feature extractions when dealing with the textual data. The use of the models uses the test-based knowledge of the Lexicon for enhanced accuracy and nuance in the linguistic data during the sentiment analysis process. It is notable that the differences in the sentiments for the same word across different cultures and languages is one of the most problematic issues ever when analysing the emotion embedded within texts (Waheeb et al., 2020). BERT processes contextual data bidirectionally taking into account the sentiments of the surrounding words to grasp the sentiment nuances thus leading to more accuracy during the classification. The model has shown its superiority over the four other model by proving the accuracy of 0.99 and 0.98 precision whereas the other models barely had moderate performances. This has shown the effectiveness of the model to capture diverse sentiment patterns within the sampled dataset. While BERT is explained by local interpretable model agnostic explanations, it clarified how the different parts of the input features are of importance towards the final sentiment score. The use of the XAI techniques even aid in clarifying the way in which the sentiment predictions are met, enhancing the user trust and accountability, and enabling the users to make informed decisions based on the analysis insights.

# 6. CONCLUSION

## 6.1 Main Findings

In terms of lexicon expansion, additional words had to be added in order to translate from one language to another, as conjunctions were often missing. When translating sentences, they were accurate to some extent, but often did not fit the contextual meaning of the original sentence. The translation worked best on short, direct sentences. For the machine learning models, random forest was the most balanced, being able to detect the polarity of the sentiment value. For context based analysis using LLM, the BERT model was successful at capturing the diverse sentiment patterns.

## 6.2 Recommendations for Future Work

For any future analysis, a more balanced lexicon in terms of sentiment should be used, to ensure machine learning models can capture the differing sentiments. Language rules for translation functions should be included, such as double negatives for Afrikaans. This will allow for sentences to be translated from one language to another, with a higher accuracy. Any future work should further explore the use of LLMs, to be able to contextually analyse sentiment values. Lastly, future research will incorporate human evaluators to assess the accuracy of translations between low-resource languages.

**Lexicon availability.** The lexicon is available upon request from the Low-Resource Language Processing Lab (LRLPL) coordinator, Dr. Mike Wa Nkongolo, at the University of Pretoria (email: mike.wankongolo@up.ac.za).